\documentclass[10pt,twocolumn,letterpaper]{article}

\usepackage{iccv}
\usepackage{times}
\usepackage{epsfig}
\usepackage{graphicx}
\usepackage{amsmath}
\usepackage{amssymb}
\usepackage{caption}
\usepackage{booktabs}
\usepackage{pifont}
\usepackage{multicol}
\usepackage{multirow}

\usepackage[pagebackref=true,breaklinks=true,letterpaper=true,colorlinks,bookmarks=false]{hyperref}

\newcommand{\PAR}[1]{\vskip4pt \noindent{\bf #1~}}
\newcommand{\bftab}{\fontseries{b}\selectfont}

\iccvfinalcopy 

\def\iccvPaperID{11032} 

\ificcvfinal\fi

\begin{document}
\title{The Animation Transformer: Visual Correspondence via Segment Matching}

\author{
Evan Casey$^1$
\quad
Víctor Pérez$^1$
\quad
Zhuoru Li$^2$ \\
\quad
Harry Teitelman$^1$
\quad
Nick Boyajian$^1$
\quad
Tim Pulver$^1$
\quad
Mike Manh$^1$ 
\quad
William Grisaitis$^1$ \\
\vspace{2mm}
{$^1$Cadmium \quad $^2$Project HAT}
}

\twocolumn[{%
\vspace{-1em}
\maketitle
\vspace{-1em}

\ificcvfinal\thispagestyle{empty}\fi

\makebox[0pt][l]{%
\begin{minipage}{\textwidth}
\centering
    \includegraphics[width=1.0\textwidth]{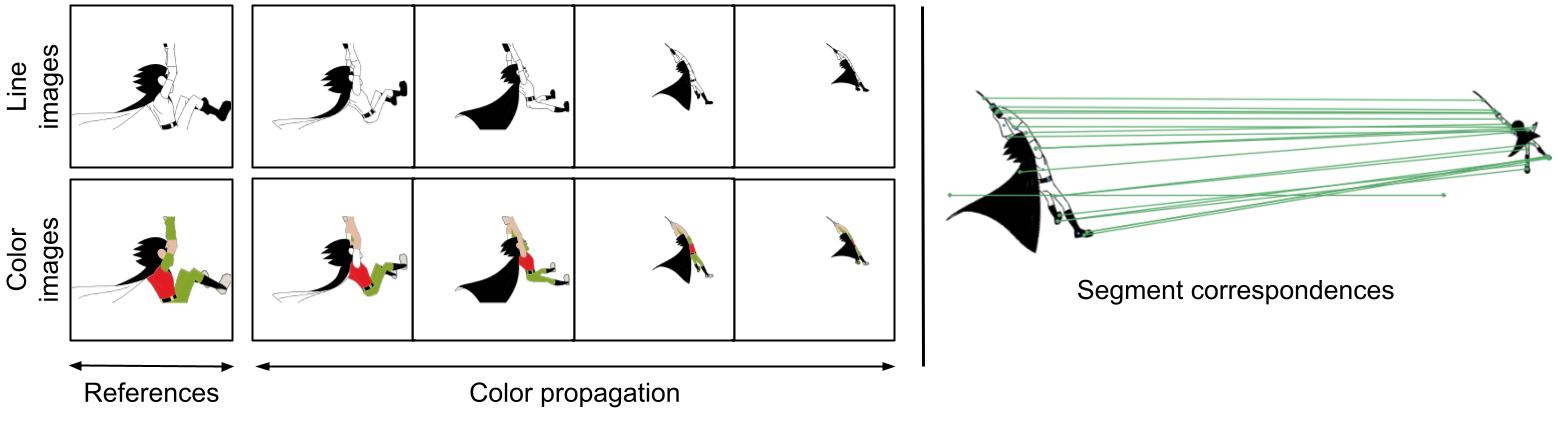}
 \vspace{-.5em}
 \captionof{figure}{(right) Our model learns visual correspondence at the level of line-enclosed segments in the line images. (left) Using the learned segment correspondences, our model performs colorization by propagating colors from a reference image across a sequence of grayscale line images.}
 \label{fig/main}
 \vspace{2em}
\end{minipage}
}

}]

\begin{abstract}
\vspace{-3mm}
Visual correspondence is a fundamental building block on the way to building assistive tools for hand-drawn animation. However, while a large body of work has focused on learning visual correspondences at the pixel-level, few approaches have emerged to learn correspondence at the level of line enclosures (segments) that naturally occur in hand-drawn animation. Exploiting this structure in animation has numerous benefits: it avoids the intractable memory complexity of attending to individual pixels in high resolution images and enables the use of real-world animation datasets that contain correspondence information at the level of per-segment colors. To that end, we propose the Animation Transformer (AnT) which uses a transformer-based architecture to learn the spatial and visual relationships between segments across a sequence of images. AnT enables practical, state-of-art AI-assisted colorization for professional animation workflows and is publicly accessible as a creative tool in Cadmium\footnote{Download Cadmium at \href{https://cadmium.app}{https://cadmium.app}}.

\end{abstract}

\vspace{-0.2in}
\section{Introduction}

Hand-drawn animation has been around for over 100 years and is one of the most popular mediums of digital entertainment today. Though the advent of drawing tablets and digital software have made the process of creating hand-drawn animation substantially easier, it is still a highly manual process that involves drawing and editing each individual frame. Many of these tasks lie in the grey area between repetitively algorithmic processes and artistic choices, opening the door for new assistive tools that augment artists’ workflows.

Existing commercial tools have applied heuristic algorithms in this domain with limited results, usually requiring artists to work in vector format or use complex character rigging that removes the hand-drawn feel of the final product. Deep learning approaches, on the other hand, can act directly on top of raw pixel input but cannot scale easily to HD resolutions and fail to properly exploit the structure of hand-drawn animation drawings -- specifically, the smaller line enclosures (segments) which can be extracted by a flood-fill or morphological algorithm.

In this paper, we focus on the task of learning visual correspondence across sequences of  raster animation line drawings. This is a fundamental building block for building assistive animation tools for tasks such as coloring, in-betweening, and texturing which make up a large portion of the tedious, non-creative work in the animation pipeline. With correspondence information an animator can color or texture a few frames in a sequence and propagate the colors through the rest of the images, saving hours of manual labor. New in-between frames can be generated by morphing neighboring frames with correspondence information, which can reduce the amount of line drawings needed to make smooth looking motion.

Despite demand for a data-driven solution to the correspondence problem, little progress has been made because of the difficult design requirements and lack of available data with correspondence labels. Suitable approaches should (i) operate on raster input and scale to HD (1920×1080) and above resolutions; (ii) produce correspondences on the level of segments; (iii) be able to handle complex real-world animation; (iv) be trainable using colorized images as supervision; (v) be fast enough for interactive applications.



\begin{figure}[h]
\centering
\includegraphics[width=0.95\linewidth]{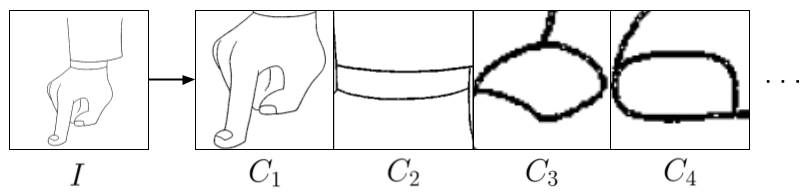}
\vspace{-0.1in}
\caption{ Given an input image $I$ each crop $C_{i}$ is obtained by placing a bounding box around the center of each enclosure of $I$ and resizing it to a common size. }
\vspace{-0.1in}
\label{fig/crops}
\end{figure}

In this paper, we propose the Animation Transformer (AnT) to address these issues. Unlike pixel-based video tracking methods which suffer from the intractability of computing attention over a large number of pixels, AnT operates over the line-enclosed segments (see Figure \ref{fig/crops}) in the line image and uses a Transformer-based architecture to learn the spatial and visual relationships between segments. By operating on this representation AnT avoids the need to directly process HD images in their entirety and is both compute and memory efficient, scaling to 4K images and beyond. We optimize AnT with a forward matching loss and a cycle consistency loss that enables it to be trained on real-world animation datasets without full ground-truth correspondence labels. 

We conduct extensive experiments to show our model’s effectiveness in a variety of settings. When trained on ground-truth correspondence labels generated from 3D rendering software, AnT demonstrates a large improvement over a strong pixel-based baseline even after domain specific improvements are added to the baseline. When AnT is trained solely on colorized images from a real-world animation dataset, its performance approaches that of the model trained on ground-truth correspondence labels -- showing that AnT is not bounded by the availability of large datasets with correspondence labels. While AnT has broad applicability in animation, we highlight its potential as a creative tool through showcasing results on guided colorization from reference images.

\section{Related work}

\PAR{Correspondence Matching:}
Our paper builds off of a growing body of research that learns correspondence by matching features extracted from a deep neural network across images. A common approach is to extract high-level activation maps from the image and match corresponding regions in feature space. This framework has been applied to video tracking \cite{vondrick2018tracking, lai2019self} and exemplar-based colorization \cite{meyer2018color, zhang2019deep} in the photo-realistic domain as well as exemplar-based colorization in the line image animation domain \cite{shi2020deep,zhang2021lineart}. However, representations learned in this way are inherently limited by the memory complexity of computing dense pixel attention maps. Even with multi-scale techniques \cite{li2020correspondence} or local attention \cite{lai2019self}, it is computationally infeasible to use these techniques for HD and above resolutions. In contrast, our approach computes attention over the line-enclosed segments in the images, which makes the attention operation bounded by the number of segments rather than pixels in the input image.

Research has also explored using the feature matching framework in combination with different ways of representing image regions, such as patches \cite{jabri2020spacetime,bertasius2021spacetime} and local descriptors \cite{superglue, luo2019contextdesc}. Of particular interest to our approach is the line of research that learns multi-view correspondence across sketch images with local descriptors \cite{navarro2021sketchzooms,deng2020sketchdesc}. However, our domain necessitates we learn correspondence at the level of segments so that we can train on real-world animation datasets with segment-level color labels and use the learned correspondences as an assistive tool for coloring.

\PAR{Segment-based methods:} Segments offer a natural way to decompose line images into a useful representation for learning tasks such as correspondence. Relevant to our approach is the work of Zhu et al. \cite{zhu2016toontrack}  which formulates segment-level correspondence matching across a sequence of images as a network flow graph problem and solves for the global optimum using the k-shortest path algorithm with Shape Context \cite{shape_context} features. Other work in this direction adopt a similar graph matching approach and apply spectral matching \cite{maejima2019graph} and quadratic programming \cite{sato2014reference} on top of non-learned segment features. Recent work from Dang et al. \cite{dang2020correspondence} proposes using a U-Net to extract local features and optimize for correspondence matches with a triplet loss that minimizes the distances between matching segments and penalizes low distances between non-corresponding segments. Similar to these approaches, AnT uses global feature aggregation across segments to learn correspondences. However, we are the first to explore using a  to aggregate segment features and do not require ground truth correspondences or hard example mining as input data.

\PAR{Transformers:} Transformers have been shown to be highly effective at learning a wide range of tasks in domains such as language
modeling \cite{attention-is-all-you-need}, image recognition \cite{vit}, object detection \cite{detr}, and protein folding \cite{senior2020improved}. Transformers introduced self-attention layers, which, similarly to Non-Local Neural
Networks \cite{wang2018nonlocal}, scan through each element of a sequence and update it by aggregating information from the whole sequence. Recent applications of Transformers to computer vision use image patches \cite{vit, detr, bertasius2021spacetime} to break up the image into a tractable sequence length that avoids the quadratic complexity of computing attention over every pixel. Sarlin et al. \cite{superglue} propose using a Transformer-based architecture to match sets of local feature descriptors where the matching assignments are estimated by solving a differentiable optimal transport problem. We design our Transformer architecture in a similar fashion, but use a different matching and loss formulation to handle the fact that one-to-none and one-to-many correspondences can occur in our domain.
\PAR{Cycle consistency:} Cycle-consistency has been applied as a learning objective for 3D shape matching, image alignment, depth estimation, and image-to-image-translation \cite{zhu2017cyclegan}. In the context of temporal domains, it can be a rich source of learning signal because the visual world is continuous and smoothly-varying. Recent work has shown that cycle-consistency is useful for learning visual tracking in the photo-realistic domain \cite{lai2019self, jabri2020spacetime} by learning to propagate labels in a forward-backward fashion. Our work applies this idea in the context of segment labels which allow us to train on datasets without ground-truth correspondence labels.
\PAR{Sketch-oriented Deep Learning:} Our work is also tangentially related to the broader area of sketch-oriented deep learning. Research has investigated methods for a variety of tasks, such as single-image colorization from hints \cite{zhang2018twostage,yonetsuji2017paintschainer,ci2018user}, sketch clean-up \cite{simo2018realtime,simo2016learning}, sketch generation \cite{ha2017neural,ge2021creative}, sketch shadowing \cite{zheng2020learning}, and synthesis of vector graphics \cite{reddy2021im2vec}.
\PAR{Assistive Animation Tools:} Finally, we take inspiration from a variety of creative tools that aim to augment the animation pipeline. LazyBrush by Sýkora et al. \cite{sykora2009lazybrush} paints hand-made cartoon drawings from imprecise color strokes. EBSynth by Jamriška et al. \cite{jamriska2019ebsynth} uses patch-based synthesis to paint over photo-realistic video from exemplar images with texture coherence, contrast and high frequency details. BetweenIT by Whited et al. \cite{brian2010betweenit} use stroke interpolation from keyframes for smooth in-betweening of vectorized animation. Zhang et al. \cite{zhang2021lineart} propose a system for colorizing in-between frames from line frames and colorized keyframes using a deep neural network. This work shares a similar goal to AnT but operates on the level of pixels instead of segments.

\section{Method}

\PAR{Motivation:}
Our goal is to estimate visual correspondence across a sequence of animation frames at the level of the line-enclosed segments in the line images. By using this naturally occurring structure (see Figure \ref{fig/crops}), we learn the spatial and positional relationships between segments; for example, a hand will have segments for each finger which are all connected to a segment for the palm. As the character moves throughout the sequence, we can expect the structure to hold; if we see several finger shaped segments we know we will see a round palm segment or small fingernail segments nearby (see Figure 3). However, due to occlusion and motion, a segment may completely go out of frame or be split into smaller sub-parts in the next frame that both correspond to the same segment in the earlier frame (see Figure 3). Thus, we formulate AnT as a segment matching problem where segments can match to 0, 1, or multiple segments in the other frames.


\PAR{Data:} The architecture for AnT is motivated by the structure of data it operates on as well as the availability of two types of labels: correspondence labels that assign each segment a consistent, \textit{unique} ID throughout the sequence and color labels that assign each segment a consistent, but possibly \textit{non-unique} color. Correspondence labels offer the cleanest, most direct form of supervision for our task; but they come at the expense of not existing in the real-world -- in order to obtain these we use 3D rendering software to generate the realistic looking line images with unique segment IDs (for more details see Section \ref{dataset}). On the other hand, colorized animation is plentiful in the real-world but offers a weaker form of supervision for our task; multiple segments often share the same color across the sequence, so color labels only tell the model that a segment in one frame corresponds to something in the set of segments in the other frame that share the same color. Our architecture is able to operate on and learn effectively from both forms of supervision.

\PAR{Formulation:} Consider two line images $A \in \mathbb{R}^{H\times{W}\times{1}}$ and $B \in \mathbb{R}^{H\times{W}\times{1}}$ which have $M$ and $N$ segments and are indexed by $\mathcal{A} := \{1,..., M\}$ and $\mathcal{B} := \{1,..., N\}$, respectively.  We extract segments from the line images using a trapped-ball filling algorithm where each line enclosed region is a separate segment. We divide the image into a set of smaller cropped images using the bounding box coordinates of each segment and then resize each cropped image to a smaller resolution $H_{c},{W_c}$. Each segment has positional information $\mathbf{p}_i = (x_i,y_i,h_i,w_i)$ in the form of its bounding box coordinates and visual information $\mathbf{d}_i \in \mathbb{R}^{H_{c}\times{W_c}\times{2}}$ in the form of the concatenated line image and binary segmentation mask crops. We refer to these features $\mathbf{x}_i$ jointly as the local segment features.


\begin{figure*}[h]
\centering
\includegraphics[width=\linewidth]{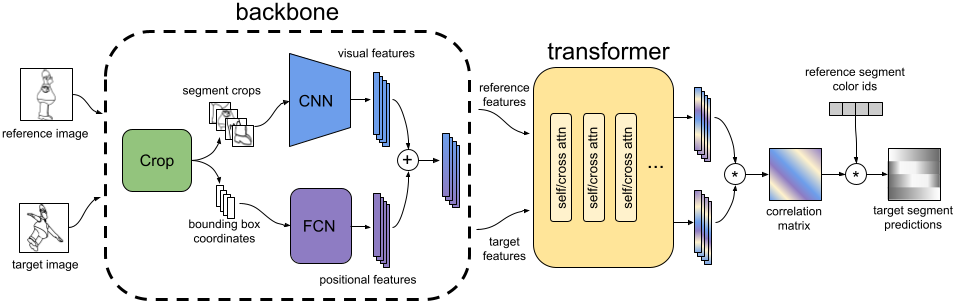}
\vspace{-0.1in}
\caption{\textbf{AnT Architecture.} Given reference and target line images, the backbone module extracts visual and positional information for each segment. The per-segment features are passed through a multiplex transformer architecture that aggregates information across segments and frames, yielding a similarity matrix between the reference and target segments. The final color predictions are computed via
a linear combination of the color labels in the reference frame.}
\label{fig/arch}
\vspace{-0.1in}
\end{figure*}

\subsection{AnT Architecture}

As shown in Figure \ref{fig/arch}, our model consists of three main modules: the CNN backbone network to extract visual features for each segment, the bounding box encoder to extract positional embeddings for each segment, and a multiplex transformer which learns the global structure across segments and frames and predicts the final match matrix. 

\begin{figure}[h]
\centering
\includegraphics[width=\linewidth]{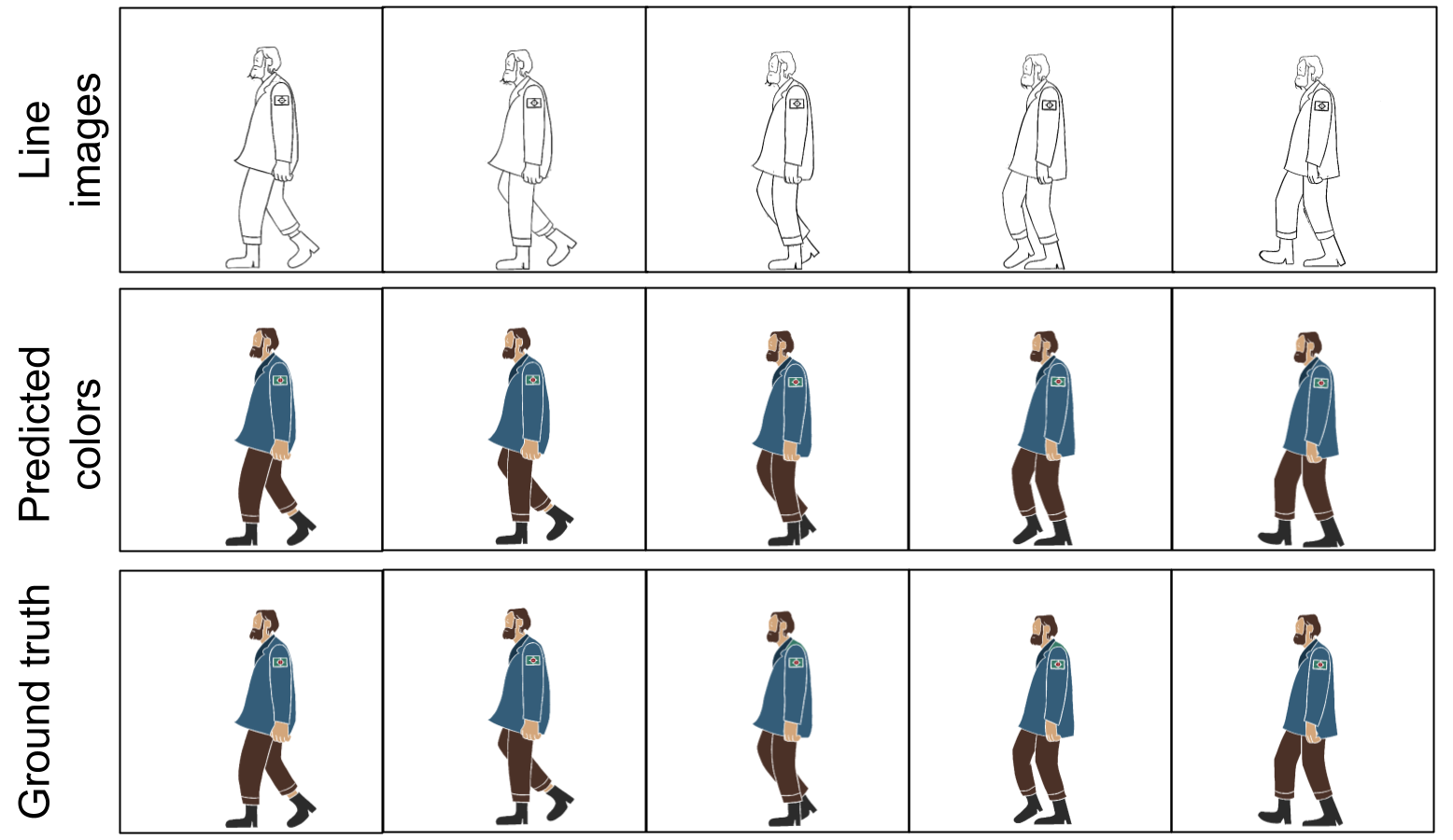}
\vspace{-0.1in}
\caption{An example of occlusion in our evaluation data. We show results of our model's performance on this sequence in the middle row.}
\label{fig/occlusion}
\vspace{-0.1in}
\end{figure}

\begin{figure}[h]
\centering
\includegraphics[width=0.9\linewidth]{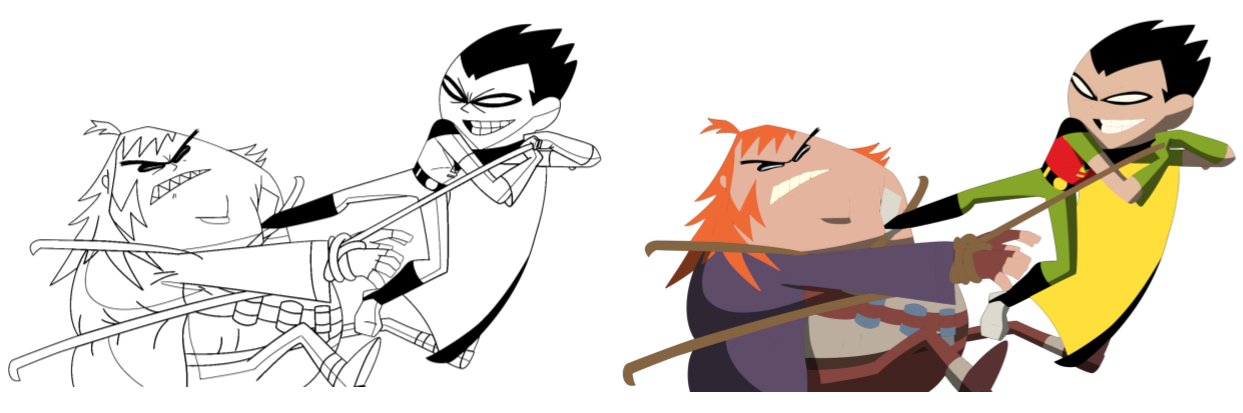}
\vspace{-0.1in}
\caption{An example of grouping in animation: if you zoom in on the image on the left, you will see many smaller segments generated from the shadow pass that pertain to the same semantic group.}
\label{fig/grouping}
\vspace{-0.2in}
\end{figure}

The multiplex transformer architecture is inspired by \cite{superglue} and we encourage readers to refer to the original SuperGlue paper for additional details. While the positional and visual features are an important foundation for estimating segment correspondences, there are often visual ambiguities that arise which cannot be solved by looking at local features alone. For example, in Figure \ref{fig/occlusion} we see examples of cases that would make matching on local features alone impossible: an occlusion or deformation can disfigure an individual segment or there may be multiple segments such as eyes that are indistinguishable from one another if viewed in isolation. Additionally, animation line drawings often contain groups of neighboring segments that pertain to the same semantic part but are split into multiple segments because the artist has drawn an object in the foreground whose contour lines intersect with that an object behind it (see Figure \ref{fig/grouping}).

These challenges motivate the need for an architecture that can aggregate global feature information across segments within the individual images as well as integrate segment information across images. We describe this in more detail in the following sections.


\PAR{CNN Backbone:}
Starting from the cropped images $\mathbb{R}^{H_{c}\times{W_c}\times{2}}$, a conventional CNN backbone generates high-level activation maps for each segment crop. A $1\times{1}$ convolution squashes the spatial dimensions of the high-level activation maps yielding $D$ dimensional feature vectors. In our experiments we use $D = 256$ and $H_c, W_c = 32$.

\PAR{Positional Encoder:}
We combine the visual features from the CNN backbone with positional information from the bounding box coordinates to get the final local features $\mathbf{x}_i$ for each segment. We embed the bounding box coordinates into a $D$ dimensional vector with a multilayer perceptron (MLP) and add these into the visual features:
\begin{equation}
    \label{eq:keypoint-encoder}
        \mathbf{x_i} = \text{CNN}_{\text{enc}}(\mathbf{d}_i ) + \text{MLP}_{\text{enc}}\left(\mathbf{p}_i\right).
    \end{equation}
Unlike \cite{superglue}, we train both the CNN backbone and positional encoder end-to-end with the multiplex transformer.

\PAR{Multiplex Transformer:} As in \cite{superglue}, we adopt a multiplex transformer architecture, which has two modes of information aggregation: it connects segments to all the other segments within the same image (self-attention) and connects segments to all the segments in the other image (cross-attention). In self-attention, features are aggregated at the level of segments within each individual image yielding features $\mathbf{z}_{\mathcal{A}}^\ell,  \mathbf{z}_{\mathcal{B}}^\ell$ for input images $\mathcal{A}, \mathcal{B}$, respectively. Cross-attention operates over the output of the last self-attention step but aggregates information across images, yielding a new set of features  $\mathbf{z}_{\mathcal{A}}^{\ell+1},  \mathbf{z}_{\mathcal{B}}^{\ell+1}$.

\begin{figure}[h]
\centering
\setlength\fboxsep{0pt}
\setlength\fboxrule{0.2pt}
\includegraphics[width=0.9\linewidth]{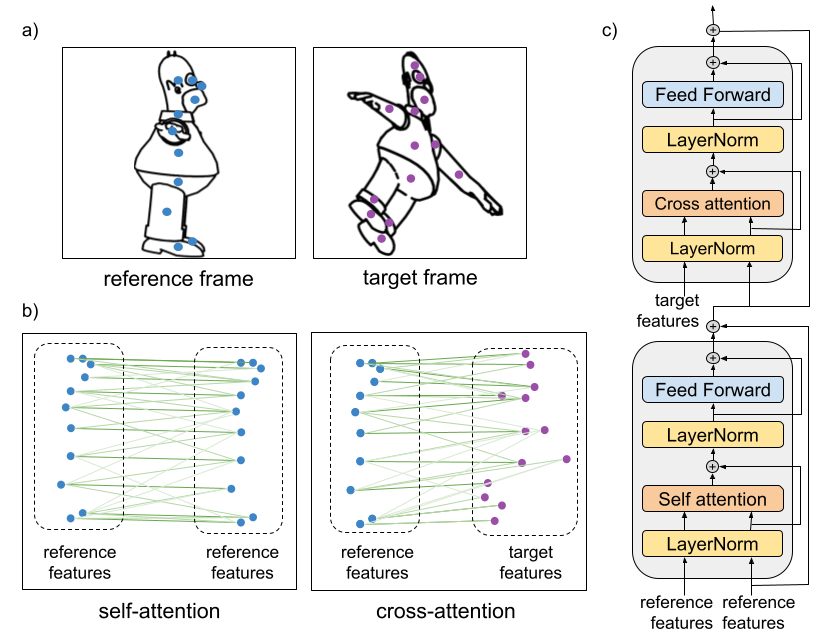}
\vspace{-0.1in}
\caption{The blue and purple circles in \textit{a)}, represent the center of the segments in each line image. The \textit{self-} and \textit{cross-attention} blocks in \textit{b)} show how attention can be computed between segment features from the same image or across images. In our architecture, we have used an interleaved approach of \textit{self-} and \textit{cross-attention} combined with skip connections between each transformer block as it is depicted in \textit{c)}.} 
\label{fig/multiplex}.
\vspace{-0.2in}
\end{figure}

In query, key, value notation our attention function can be described as a variant of the classic formulation:
\begin{equation}
   \mathrm{Attention}(\mathbf{Q}, \mathbf{K}, \mathbf{V}) = \mathrm{softmax}(\frac{\mathbf{Q}\mathbf{K}^T}{\sqrt{D}})\mathbf{V}
\end{equation}
where in the cross-attention layers, the keys and values originate from the aggregated features $\mathbf{z}_j$ of a \textit{target} image and the queries originate from the aggregates features originating from a \textit{reference} image $\mathbf{z}_i$. In the self-attention layers, the queries, keys, and values all originate from the same source features $\mathbf{z}_i$. It is important to note that \textit{reference} and \textit{target} are relative terms -- \textit{target} just denotes the other image with respect to the \textit{reference} image. In our architecture, both directions of cross-attention are happening simultaneously. From the perspective of image $\mathcal{A}$, image $\mathcal{B}$ is the \textit{target} and from the perspective of image $\mathcal{B}$, $\mathcal{A}$ is the target. An overview of the self and cross-attention blocks is illustrated in Figure \ref{fig/multiplex}.

Similar to the original transformer implementation, the multiplex transformer is made up of stacked transformer blocks that each consists of a multi-headed attention layer followed by a point-wise fully connected layer. We alternate between self and cross-attention in the transformer blocks and add residual connections between each block. The final matching features are computed by the output of the last transformer block and a final linear projection layer, yielding final features $\mathbf{z}_\mathcal{A}^L \in \mathbb{R}^{M\times{D}}$ and $\mathbf{z}_\mathcal{B}^L \in \mathbb{R}^{N\times{D}}$.

\subsection{Matching}

AnT learns a similarity matrix between the aggregated reference and target features from the multiplex transformer and then predicts the target label with a weighted sum of all the labels in the reference frame.  We compute the predicted target labels $\mathbf{\hat{c}}_j \in \mathbb{R}^N$ as a linear combination of the labels $\mathbf{c}_i \in \mathbb{R}^M$ in the reference frame:
\begin{equation}
  \mathbf{\hat{c}}_j = \sum_{i=1}^{M} \mathcal{S}_{ij} \mathbf{c}_i
  \label{eq:fwd_nonlocal}
\end{equation}
where $\mathcal{S}_{ij}$ is a similarity matrix between the target and reference frame such that the rows sum to one. As in \cite{deep-exemplar-video-colorization}, we use inner product similarity normalized by softmax:
\begin{equation}
    \mathcal{S}_{ij} = \frac{\exp\left(\mathbf{f}_i^T \mathbf{f}_j\right)}{\sum_{i=1}^{M} \exp\left(\mathbf{f}_i^T \mathbf{f}_j\right)}
    \label{eqn:sim}
\end{equation}
where $\mathbf{f}_i \in \mathbb{R}^D$ is the feature vector corresponding to the segment at index $i$ in  $\mathbf{z}_\mathcal{A}^L$ and $\mathbf{f}_j \in \mathbb{R}^D$ is the feature vector corresponding to the segment at index $j$ in  $\mathbf{z}_\mathcal{B}^L$.

\subsection{Loss}

\begin{figure}[h]
\centering
\includegraphics[width=0.9\linewidth]{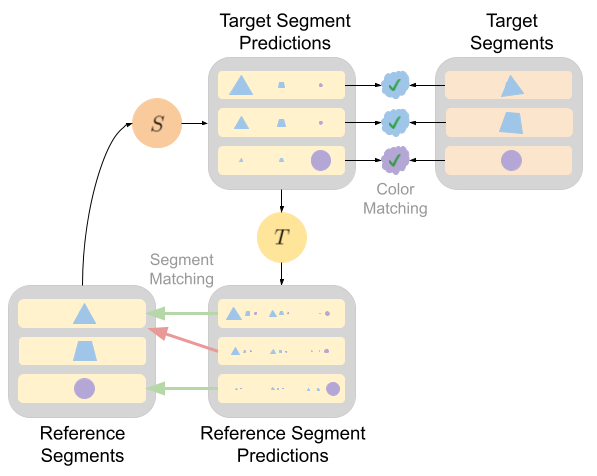}
\vspace{-0.1in}
\caption{ The cycle consistency loss allows the model to utilize real world animation data without ground truth correspondences. An example scenario is shown in which the second target segment is incorrectly matched in the forward propagation but the model is not penalized by the color matching loss because both the predicted segment color label and the ground truth color label have the same color. To solve this, we propagate unique segment IDs through the forward pass and then back again to the reference image segments, enabling our cycle consistency loss to penalize the model according to whether the propagated IDs match their original values.}
\label{fig/cycle}
\vspace{-0.1in}
\end{figure}

In order to be able to learn from both correspondence and color labels, AnT employs two loss functions that can be used independently or averaged together depending on the label source.

\PAR{Forward match loss:}
To encourage the model to directly use the correspondence or color labels, we use categorical cross-entropy loss between the predicted target labels $\mathbf{\hat{c}}_j$ and the $\mathbf{c}_j$ ground truth target labels from the dataset. In cases where we have correspondence labels, both the target labels $\mathbf{c}_j$ and the reference labels $\mathbf{c}_i$ used as input to the weighted average calculation are unique and thus the model directly minimizes incorrect correspondences. However, in the case of color labels, $\mathbf{c}_i$ and $\mathbf{c}_j$ are non-unique and the model only minimizes incorrect color assignments. This leads to the model learning to shortcut and find matches that yield the correct color assignments but are incorrect correspondences (see Figure \ref{fig/cycle}).

\PAR{Cycle consistency loss:} 
In order to solve the previously mentioned issues, we employ a cycle consistency loss that prevents the model from learning to shortcut in cases where we have non-unique color labels. Instead of using the reference labels from the dataset, we initialize a random vector of unique segment IDs $\mathbf{r}_i \in \mathbb{R}^M$ and use these in place of $\mathbf{c}_i$ for the weighted label aggregation:
\begin{equation}
  \mathbf{\hat{r}}_j = \sum_{i=1}^{M} \mathcal{S}_{ij} \mathbf{r}_i
  \label{eq:fwd_nonlocal}
\end{equation}
We then propagate the predicted target labels $\mathbf{\hat{r}}_j$ in the backward direction:
\begin{equation}
  \mathbf{\hat{r}}_i = \sum_{j=1}^{N} \mathcal{T}_{ij} \mathbf{\hat{r}}_j
  \label{eq:fwd_nonlocal}
\end{equation}
where $\mathcal{T}$ is the backward correlation matrix computed by:
\begin{equation}
    \mathcal{T}_{ij} = \frac{\exp\left(\mathbf{f}_j^T \mathbf{f}_i\right)}{\sum_{j=1}^{N} \exp\left(\mathbf{f}_j^T \mathbf{f}_i\right)}
    \label{eqn:fwd_sim}
\end{equation}
As with the forward match loss, we use categorical cross-entropy loss between the randomly initialized segment IDs $\mathbf{r}_i$ and the predicted segment IDs $\mathbf{\hat{r}}_i$ propagated over the entire cycle. Our final loss term with both losses is:
 \begin{equation}
    \mathcal{L} = \sum_{j=1}^{N} \mathcal{L}_{fwd}\left(\mathbf{\hat{c}}_j, \mathbf{c_j}\right) +  \alpha \sum_{i=1}^{Q} \mathcal{L}_{cyc}\left(\mathbf{\hat{r}}_i, \mathbf{r}_i\right)
   \label{eqn:objective}
\end{equation}
where $\mathcal{L}_{fwd}$ is the forward matching loss, $\mathcal{L}_{cyc}$ is the cycle consistency loss, and $\alpha$ is a hyper-parameter that weights the cycle consistency loss. In our experiments we use both losses with $\alpha = 0.25$.




\section{Experiments}

\subsection{Dataset details}
\label{dataset}

\PAR{Synthetic Dataset:}
To train AnT with ground truth segment correspondence labels, we generate a synthetic dataset in Cinema4D using freely available 3D models. We render realistic looking line images using a toon shader and generate the segment correspondence labels by assigning unique IDs to individual meshes. The characters are rigged with different movements, deformations, and rotations to simulate actual animation. We use 11 3D character models from TurboSquid and generate 1000 frames at 1500x1500 pixel resolution for each character, yielding 11000 frames in total. During training, we apply random frame skipping and other augmentation techniques such as cropping, jittering, and shearing. The characters range in complexity from some characters with as few as 10 segments to others with as high as 50. We create our evaluation set by randomly selecting sequences for a total of 1100 frames (10\% of the dataset), uniformly split across each character.

\PAR{Real Dataset:} As a medium, hand-drawn animation is much more diverse and expressive than its 3D counterpart. Since animators are not confined to the limits of a 3D program, hand-drawn animation encompasses a much broader set of animation styles and character designs. For any visual correspondence model to work in the wild on a variety of animation styles, it cannot only be trained on synthetic data from 3D programs. To solve this, we collect a dataset of high resolution hand-drawn animation from 17 different real-world animation productions, totalling 3578 frames. The animation style of each production varies greatly, although the style is closer to U.S. and European animation. The dataset is extremely diverse, with hundreds of different characters. Importantly, the real dataset does not have unique correspondence labels; we use the segment colors in the colorized images to extract labels. In contrast with the synthetic dataset, this yields \textit{non-unique} numeric segment labels. We create our evaluation set by randomly selecting sequences 25 variable-length sequences uniformly across each production. For 5 sequences in the evaluation set, no training data from the originating production exists in the training set at all.

\subsection{Implementation details}

\PAR{Training details:} We train AnT using the AdamW optimizer with and a learning rate of 5e-4, weight decay of 1e-4, gradient clipping at global norm 1. We use a learning rate warmup of 1K steps, and train for 100k iterations with no learning rate decay. AnT is trained with an effective batch size of 64 using gradient accumulation over 4 batches of 16 image pairs each. The transformer has input and attention dropout of 0.1, which we found helpful for regularization. Unless otherwise specified, we train AnT with $L=9$ layers of alternating multi-head self- and cross-attention with 4 heads each and $D=256$ dimensional local features.

\PAR{Time and memory complexity:} A single forward pass of AnT takes on average 76ms (13 FPS) on a Nvidia Tesla V100 GPU. Using $M$ and $N$ to denote the number of reference and target segments, each cross attention layer AnT has to make $\mathcal{O}(MN)$ comparisons and each self attention layer AnT has to make $\mathcal{O}(M^2 + N^2)$ comparisons. By comparison, a forward pass of DEVC takes on average 147ms (6 FPS). Memory-wise, DEVC has to make $\mathcal{O}((HW)^2)$ comparisons, where $H$,$W$ are the spatial dimensions of the CNN features. We were limited to using a batch size of 3 for DEVC, whereas we could use a batch size of 64 for AnT, yielding much faster training. Our leak-proof filling method implemented in OpenGL takes on average 1.4s on the same hardware, yielding a total inference speed of 2.16s for AnT or 2.87s for DEVC. 

\subsection{Comparisons}

\PAR{Baselines:} We compare the performance of AnT to both the vanilla implementation of Deep Exemplar Video Colorization (DEVC) \cite{deep-exemplar-video-colorization} as well as variants of DEVC with domain specific modifications. DEVC is a state-of-the-art video colorization network that operates on the pixel-level and matches features with a deep neural network. To use DEVC in our tasks, we use only the correspondence subnet and use the categorical cross-entropy loss on the colorized warped image. We then generate a predicted segment label for each segment by a non-learned post-processing step: we take the maximally occurring color in each of the segment locations on the warped image.

Since DEVC is a pixel-based approach, we create two variants with domain specific enhancements that take advantage of the problem structure. Since small segments are the hardest to predict, we weight the loss of each pixel in the warped image output inversely proportional to the size of the segment corresponding to that pixel location. This helps prevent the network from unknowingly focusing on large segment areas while ignoring smaller ones. We refer to this network as DEVC (Weighted Loss). 

We also observed that high resolution is important for performance. We introduce the local attention mechanism used in \cite{lai2019self} in place of global attention to enable training at higher resolutions. This model is referred to as DEVC (Local Attention). We train DEVC and DEVC (Weighted Loss) at 512x512 pixel resolution with batch size of 2 until convergence. We train DEVC (Local Attention) at 640x640 pixel resolution also with a batch size of 2 until convergence.

\begin{figure}[h]
\centering
\includegraphics[width=0.9\linewidth]{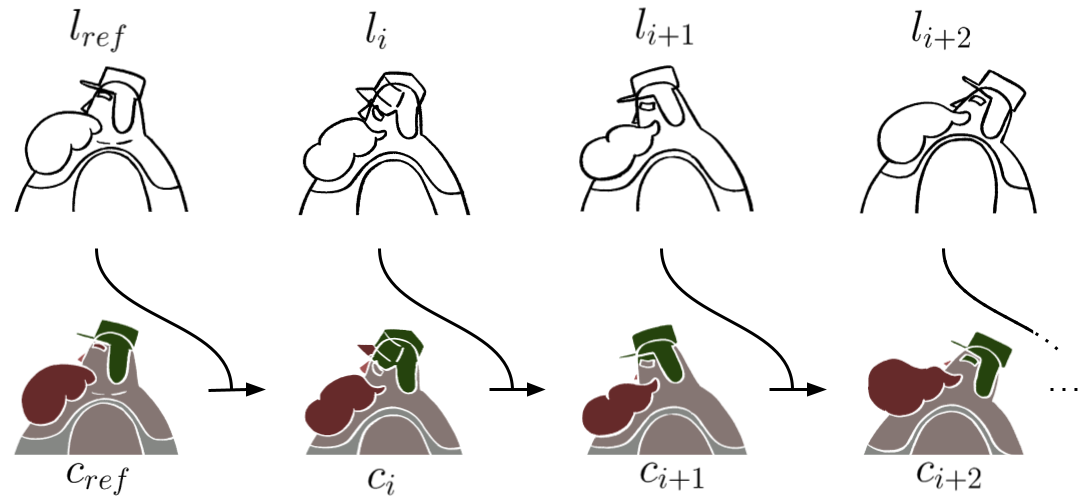}
\vspace{-0.1in}
\caption{ Starting from a reference color and line images i.e. $c_{ref}$ and $l_{ref}$, we recursively propagate the colors of each generated image $c_{i}$ to colorize every incoming line image $l_{i+1}$. }
\label{fig/recursive}
\vspace{-0.2in}
\end{figure}

\PAR{Metrics:} To measure correspondence across sequences, we recursively propagate segment labels over 10 frames as illustrated in Figure \ref{fig/recursive}, using a single ground truth reference frame to seed the colors for the rest of the predictions. We use per-segment label accuracy and mean Intersection-Over-Union averaged over the label classes as our evaluation metrics.

\begin{figure*}[h]
\centering
\includegraphics[width=\linewidth]{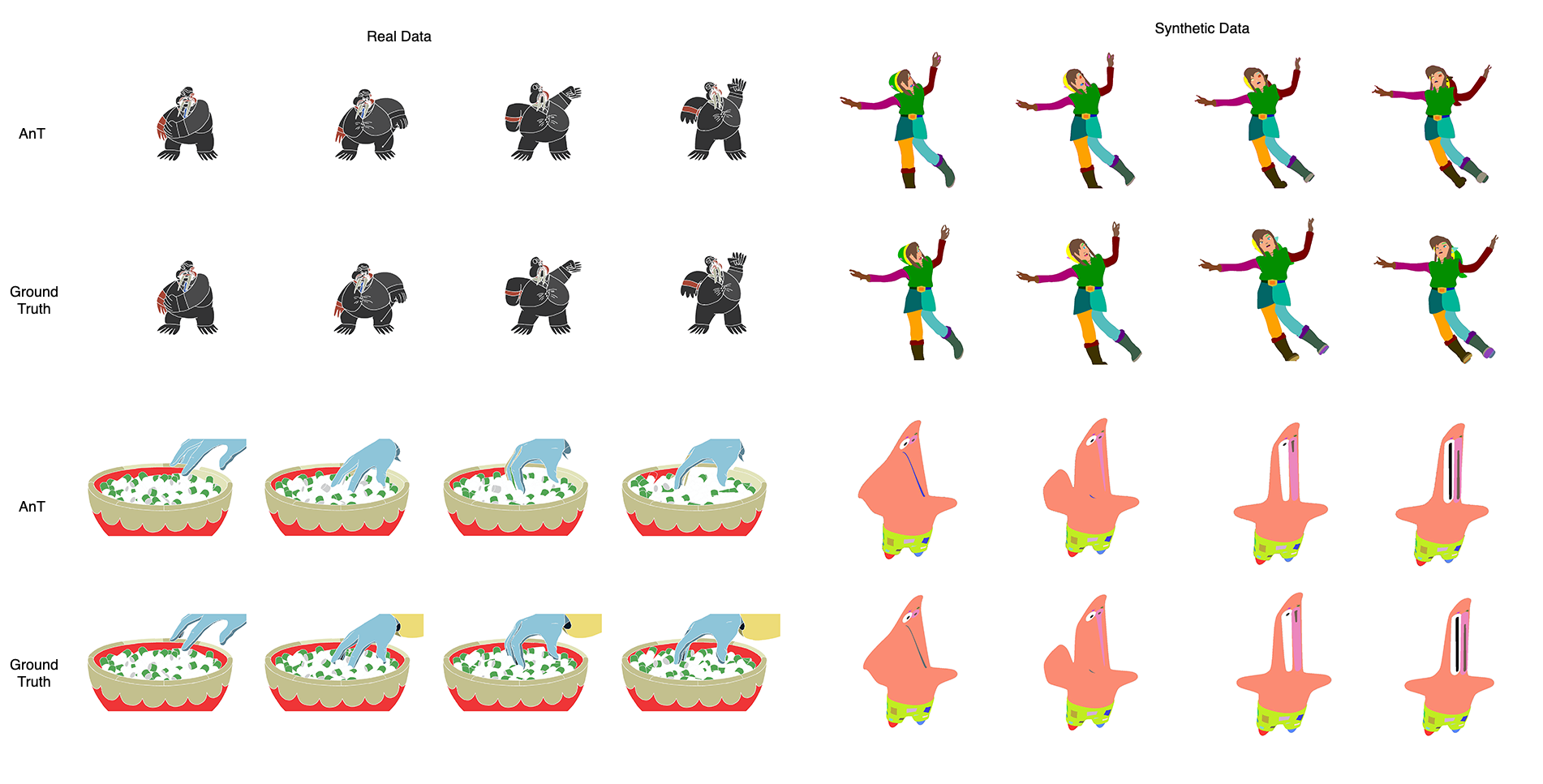}
\vspace{-0.3in}
\caption{AnT is effective at colorizing complex scenes with occlusion, small segments, and complex deformations. In the bottom left example, AnT fails to colorize the yellow sleeve because it was not present in the reference line image. We encourage readers to look at the Appendix for additional results.}
\label{fig/wide}
\vspace{-0.1in}
\end{figure*}

\PAR{Results:} We show qualitative results in Figure \ref{fig/wide} and results of comparing AnT to DEVC in Table \ref{tab:comp} on both the synthetic and real datasets. The synthetic column is evaluated on the ground truth segment correspondence labels, while the real dataset is evaluated on the non-unique color labels.

\begin{table}[h]
  \centering
  \resizebox{\columnwidth}{!}{%
  \begin{tabular}{r  c c c c}
    \toprule
    & \multicolumn{2}{c}{\bf Synthetic} & \multicolumn{2}{c}{\bf Real} \\
  & Accuracy & Mean IoU & Accuracy & Mean IoU \\
  \midrule
    DEVC & 66.19 & 43.17 & 42.86 & 29.37 \\
    DEVC (Weighted Loss) & 79.92 & 55.98 & 61.86 & 38.05 \\
    DEVC (Local Attention) & 84.11 & 62.60 & 57.34 & 32.98 \\
    \bftab{AnT (Ours)} & \bftab{92.17} &  \bftab{72.90} & \bftab{79.38} & \bftab{45.38}  \\
    \bottomrule
  \end{tabular}
  }
  \caption{\label{tab:comp}
    {\bf Evaluation on Correspondence (Synthetic) and Colorization (Real).} AnT strictly outperforms all the baselines, even after segment-specific modifications are added. The real dataset contains chunkier motion that moves outside the field of view in DEVC (Local Attention)
    }
    \vspace{-0.2in}
\end{table}

\subsection{Model Ablation Study}
We pull apart several key components of AnT to show how performance changes when these components are removed (see Table \ref{tab:ablation}). The transformer is highly correlated with performance, which shows that the global feature aggregation helps learn effective representations. Similarly, spatial information is also necessary for AnT to reason about the segment structures effectively. When cycle-consistency is removed in the model trained on the real dataset, the model avoids learning generalizable correspondences -- it "cheats" by matching non-corresponding segments with the same color.

\begin{table}[h]
  \centering

  \resizebox{\columnwidth}{!}{%
  \begin{tabular}{r  c c c c}
    \toprule
    & \multicolumn{2}{c}{\bf Synthetic} & \multicolumn{2}{c}{\bf Real} \\
  & Accuracy & Mean IoU & Accuracy & Mean IoU \\
  \midrule
    No transformer & 78.56 & 67.82 & 65.91 & 39.53 \\
    No positional embedding & 81.88 & 67.23 & 68.23 & 40.20  \\
    No cycle consistency & 91.49 & 71.01 & 68.48 & 41.10 \\
    Smaller (3 layers) & 88.03 & 69.90 & 76.09 & 44.02 \\  
    Full (9 layers) & \bftab{92.17} &  \bftab{72.90} & \bftab{79.38} & \bftab{45.38} \\
    \bottomrule
  \end{tabular}
  }
    \caption{\label{tab:ablation}
    {\bf Model Ablation study.} Comparison of different model variants in AnT.
  }
  \vspace{-0.2in}
\end{table}

\subsection{Training Data Ablation Study}
In order to assess AnT's ability to learn without ground-truth correspondence labels, we perform an ablation study (see Table \ref{tab:training}) across 3 different training sets: synthetic, real, and mixed (which is simply the sum of synthetic and real). As in the earlier section, the synthetic and real columns denote the evaluation set with the same metrics as before.
Notably, we see that when AnT is trained on the real dataset, its performance on the synthetic correspondence dataset approaches that of when it has access to correspondences at training time. The real dataset is much more challenging and diverse, leading to a more robust model that can predict correspondences on the less challenging synthetic dataset. The inverse is not true; when the model trained on only synthetic data is evaluated on real, a bigger performance differential exists. We hypothesize this is because the synthetic dataset lacks diversity and challenge. 

\begin{table}[h]
  \centering

  \resizebox{\columnwidth}{!}{%
  \begin{tabular}{r  c c c c}
    \toprule
    & \multicolumn{2}{c}{\bf Synthetic} & \multicolumn{2}{c}{\bf Real} \\
  & Accuracy & Mean IoU & Accuracy & Mean IoU \\
  \midrule
    Synthetic & 92.17 & 72.90 & 72.55 & 39.93 \\
    Real & 89.46 & 70.20 & 79.38 & 45.38 \\
    Mixed & \bftab{94.25} & \bftab{77.27} & \bftab{79.84} & \bftab{51.64} \\
    \bottomrule
  \end{tabular}
  }
  \caption{\label{tab:training}
    {\bf Training Data Ablation Study.} While the best results on synthetic come from the mixed training set, when AnT is only trained on the real dataset its performance approaches that of the model trained on correspondence labels.
  }
  \vspace{-0.2in}
\end{table}

\section{Conclusion}

In this paper, we have shown that segment is an effective structure for learning visual correspondence on hand-drawn images. Our results show our method's ability to leverage real-world animation datasets that are crucial for learning accurate correspondences on a wide variety of animation styles.

We hope this work encourages more research into practical, data-driven creative tools for animation. Although we focused on flat-filled animation in this work, our method can be extended to other tasks such as propagating shadows and texture or predicting optical flow.

\section{Acknowledgements}

We'd like to thank Andrew Drozdov, Xavier Snelgrove, James MacGlashan, Eric Jang and everyone at ML Collective for helpful conversations that helped inform the design of AnT. We would also like to thank Paul-Edouard Sarlin for the SuperGlue code release and answering questions about hyperparameter settings and model scaling. We also thank Masha Shugrina for answering many questions about the CreativeFlow dataset. Finally, we are grateful to the engineering team at Spell for their work on training infrastructure that this project used. 

\section*{Author Contributions}
\noindent
\textbf{Evan Casey} led the research and technical direction and wrote the majority of this manuscript.

\noindent
\textbf{Evan Casey and Víctor Pérez} implemented the models, training infrastructure, and conducted all the experiments used in the paper.

\noindent
\textbf{Víctor Pérez} helped come up with the transformer-based architecture, wrote early versions of the manuscript, and created most of the visualizations and figures used in the paper.

\noindent
\textbf{Zhuoru Li} contributed to many of the ideas in AnT from his parallel research on applying transformers to colorization. He advised on some critical pieces of the research and wrote parts of the manuscript.
   
\noindent
\textbf{Harry Teitelman} managed the non-synthetic (real) dataset curation and labeling pipeline.
   
\noindent
\textbf{Harry Teitelman and Nick Boyajian} developed the synthetic data generation pipeline.

\noindent
\textbf{Tim Pulver, Harry Teitelman, and Nick Boyajian} built the Cadmium desktop app.

\noindent
\textbf{Mike Manh} developed the OpenGL segmentation algorithm.

\noindent
\textbf{William Grisaitis} contributed to early versions of the training infrastructure and was helpful in early research discussions.


{\small
\bibliographystyle{ieee_fullname}
\bibliography{egbib}

\begin{thebibliography}{10}\itemsep=-1pt

\bibitem{shape_context}
Serge Belongie, Jitendra Malik, and Jan Puzicha.
\newblock Shape context: A new descriptor for shape matching and object
  recognition.
\newblock In T. Leen, T. Dietterich, and V. Tresp, editors, {\em Advances in
  Neural Information Processing Systems}, volume~13. MIT Press, 2001.

\bibitem{bertasius2021spacetime}
Gedas Bertasius, Heng Wang, and Lorenzo Torresani.
\newblock Is space-time attention all you need for video understanding?, 2021.

\bibitem{detr}
Nicolas Carion, Francisco Massa, Gabriel Synnaeve, Nicolas Usunier, Alexander
  Kirillov, and Sergey Zagoruyko.
\newblock End-to-end object detection with transformers, 2020.

\bibitem{ci2018user}
Yuanzheng Ci, Xinzhu Ma, Z. Wang, Haojie Li, and Zhongxuan Luo.
\newblock User-guided deep anime line art colorization with conditional
  adversarial networks.
\newblock 2018.

\bibitem{dang2020correspondence}
Trung D.~Q. Dang, Thien Do, Anh Nguyen, Van Pham, Quoc Nguyen, Bach Hoang, and
  Giao Nguyen.
\newblock Correspondence neural network for line art colorization.
\newblock In {\em ACM SIGGRAPH 2020 Posters}, SIGGRAPH '20, New York, NY, USA,
  2020. Association for Computing Machinery.

\bibitem{vit}
Alexey Dosovitskiy, Lucas Beyer, Alexander Kolesnikov, Dirk Weissenborn,
  Xiaohua Zhai, Thomas Unterthiner, Mostafa Dehghani, Matthias Minderer, Georg
  Heigold, Sylvain Gelly, Jakob Uszkoreit, and Neil Houlsby.
\newblock An image is worth 16x16 words: Transformers for image recognition at
  scale, 2020.

\bibitem{ge2021creative}
Songwei Ge, Vedanuj Goswami, C.~Lawrence Zitnick, and Devi Parikh.
\newblock Creative sketch generation, 2021.

\bibitem{ha2017neural}
David Ha and Douglas Eck.
\newblock A neural representation of sketch drawings, 2017.

\bibitem{jabri2020spacetime}
Allan Jabri, Andrew Owens, and Alexei~A. Efros.
\newblock Space-time correspondence as a contrastive random walk, 2020.

\bibitem{jamriska2019ebsynth}
Ond\v{r}ej Jamri\v{s}ka, \v{S}\'{a}rka Sochorov\'{a}, Ond\v{r}ej Texler, Michal
  Luk\'{a}\v{c}, Jakub Fi\v{s}er, Jingwan Lu, Eli Shechtman, and Daniel
  S\'{y}kora.
\newblock Stylizing video by example.
\newblock {\em ACM Transactions on Graphics}, 38(4), 2019.

\bibitem{lai2019self}
Z. Lai and W. Xie.
\newblock Self-supervised learning for video correspondence flow.
\newblock In {\em BMVC}, 2019.

\bibitem{li2020correspondence}
Shuda Li, Kai Han, Theo~W. Costain, Henry Howard-Jenkins, and Victor
  Prisacariu.
\newblock Correspondence networks with adaptive neighbourhood consensus, 2020.

\bibitem{luo2019contextdesc}
Zixin Luo, Tianwei Shen, Lei Zhou, Jiahui Zhang, Yao Yao, Shiwei Li, Tian Fang,
  and Long Quan.
\newblock Contextdesc: Local descriptor augmentation with cross-modality
  context, 2019.

\bibitem{maejima2019graph}
Akinobu Maejima, Hiroyuki Kubo, Takuya Funatomi, Tatsuo Yotsukura, Satoshi
  Nakamura, and Yasuhiro Mukaigawa.
\newblock Graph matching based anime colorization with multiple references.
\newblock In {\em ACM SIGGRAPH 2019 Posters}, SIGGRAPH '19, New York, NY, USA,
  2019. Association for Computing Machinery.

\bibitem{meyer2018color}
Simone Meyer, , Victor Cornill\`ere, Abdelaziz Djelouah, Christopher Schroers,
  and Markus Gross.
\newblock Deep video color propagation.
\newblock In {\em Proceedings of the British Machine Vision Conference {BMVC}},
  2018.

\bibitem{navarro2021sketchzooms}
Pablo Navarro, J.~Ignacio Orlando, Claudio Delrieux, and Emmanuel Iarussi.
\newblock Sketchzooms: Deep multi-view descriptors for matching line drawings.
\newblock {\em Computer Graphics Forum}, 40(1):410--423, 2021.

\bibitem{reddy2021im2vec}
Pradyumna Reddy, Michael Gharbi, Michal Lukac, and Niloy~J. Mitra.
\newblock Im2vec: Synthesizing vector graphics without vector supervision,
  2021.

\bibitem{superglue}
Paul-Edouard Sarlin, Daniel DeTone, Tomasz Malisiewicz, and Andrew Rabinovich.
\newblock Superglue: Learning feature matching with graph neural networks,
  2020.

\bibitem{sato2014reference}
Kazuhiro Sato, Yusuke Matsui, Toshihiko Yamasaki, and Kiyoharu Aizawa.
\newblock Reference-based manga colorization by graph correspondence using
  quadratic programming.
\newblock In {\em SIGGRAPH Asia 2014 Technical Briefs}, SIGGRAPH ASIA '14, New
  York, NY, USA, 2014. Association for Computing Machinery.

\bibitem{senior2020improved}
Andrew~W Senior, Richard Evans, John Jumper, James Kirkpatrick, Laurent Sifre,
  Tim Green, Chongli Qin, Augustin {\v{Z}}{\'\i}dek, Alexander~WR Nelson, Alex
  Bridgland, et~al.
\newblock Improved protein structure prediction using potentials from deep
  learning.
\newblock {\em Nature}, 577(7792):706--710, 2020.

\bibitem{shi2020deep}
Min Shi, Jia-Qi Zhang, Shu-Yu Chen, Lin Gao, Yu-Kun Lai, and Fang-Lue Zhang.
\newblock Deep line art video colorization with a few references.
\newblock {\em arXiv preprint arXiv:2003.10685}, 2020.

\bibitem{simo2018realtime}
Edgar Simo-Serra, Satoshi Iizuka, and Hiroshi Ishikawa.
\newblock Real-time data-driven interactive rough sketch inking.
\newblock {\em ACM Transactions on Graphics (TOG)}, 37(4):98, 2018.

\bibitem{simo2016learning}
Edgar Simo-Serra, Satoshi Iizuka, Kazuma Sasaki, and Hiroshi Ishikawa.
\newblock Learning to simplify: fully convolutional networks for rough sketch
  cleanup.
\newblock {\em ACM Transactions on Graphics (TOG)}, 35(4):121, 2016.

\bibitem{sykora2009lazybrush}
D. S{\'y}kora, J. Dingliana, and S. Collins.
\newblock Lazybrush: Flexible painting tool for hand‐drawn cartoons.
\newblock {\em Computer Graphics Forum}, 28, 2009.

\bibitem{attention-is-all-you-need}
Ashish Vaswani, Noam Shazeer, Niki Parmar, Jakob Uszkoreit, Llion Jones,
  Aidan~N. Gomez, Lukasz Kaiser, and Illia Polosukhin.
\newblock Attention is all you need, 2017.

\bibitem{vondrick2018tracking}
Carl Vondrick, Abhinav Shrivastava, Alireza Fathi, Sergio Guadarrama, and Kevin
  Murphy.
\newblock Tracking emerges by colorizing videos.
\newblock In {\em Proceedings of the European Conference on Computer Vision
  (ECCV)}, September 2018.

\bibitem{wang2018nonlocal}
Xiaolong Wang, Ross Girshick, Abhinav Gupta, and Kaiming He.
\newblock Non-local neural networks, 2018.

\bibitem{brian2010betweenit}
Brian Whited, Gioacchino Noris, Maryann Simmons, Robert~W. Sumner, Markus
  Gross, and Jarek Rossignac.
\newblock Betweenit: An interactive tool for tight inbetweening.
\newblock {\em Computer Graphics Forum}, 29(2):605--614, 2010.

\bibitem{yonetsuji2017paintschainer}
Taizan Yonetsuji.
\newblock Paintschainer, 2017.
\newblock https://paintschainer.preferred.tech/.

\bibitem{deng2020sketchdesc}
D. {Yu}, L. {Li}, Y. {Zheng}, M. {Lau}, Y. {Song}, C. {Tai}, and H. {Fu}.
\newblock Sketchdesc: Learning local sketch descriptors for multi-view
  correspondence.
\newblock {\em IEEE Transactions on Circuits and Systems for Video Technology},
  pages 1--1, 2020.

\bibitem{zhang2019deep}
Bo Zhang, Mingming He, Jing Liao, Pedro~V Sander, Lu Yuan, Amine Bermak, and
  Dong Chen.
\newblock Deep exemplar-based video colorization.
\newblock In {\em Proceedings of the IEEE Conference on Computer Vision and
  Pattern Recognition}, pages 8052--8061, 2019.

\bibitem{deep-exemplar-video-colorization}
Bo Zhang, Mingming He, Jing Liao, Pedro~V. Sander, Lu Yuan, Amine Bermak, and
  Dong Chen.
\newblock Deep exemplar-based video colorization, 2019.

\bibitem{zhang2018twostage}
Lvmin Zhang, Chengze Li, Tien-Tsin Wong, Yi Ji, and Chunping Liu.
\newblock Two-stage sketch colorization.
\newblock {\em ACM Trans. Graph.}, 37(6), Dec. 2018.

\bibitem{zhang2021lineart}
Qian Zhang, Bo Wang, Wei Wen, Hai Li, and Junhui Liu.
\newblock Line art correlation matching feature transfer network for automatic
  animation colorization.
\newblock In {\em Proceedings of the IEEE/CVF Winter Conference on Applications
  of Computer Vision (WACV)}, pages 3872--3881, January 2021.

\bibitem{zheng2020learning}
Qingyuan Zheng, Zhuoru Li, and Adam Bargteil.
\newblock Learning to shadow hand-drawn sketches.
\newblock In {\em Proceedings of the IEEE Conference on Computer Vision and
  Pattern Recognition (CVPR)}, 2020.

\bibitem{zhu2016toontrack}
Haichao Zhu, Xueting Liu, Tien-Tsin Wong, and Pheng-Ann Heng.
\newblock Globally optimal toon tracking.
\newblock {\em ACM Transactions on Graphics (SIGGRAPH 2016 issue)},
  35(4):75:1--75:10, July 2016.

\bibitem{zhu2017cyclegan}
Jun-Yan Zhu, Taesung Park, Phillip Isola, and Alexei~A Efros.
\newblock Unpaired image-to-image translation using cycle-consistent
  adversarial networks.
\newblock In {\em Computer Vision (ICCV), 2017 IEEE International Conference
  on}, 2017.

\end{thebibliography}
}

\cleardoublepage

\appendix

\section*{Appendix}
In the following pages, we present additional background information, experimental details, qualitative examples of AnT in action, user study results, as well as visualizations and analysis of the learned attention patterns.

\section{Industry standards in animation}
To understand the motivation behind AnT, it is important to consider how hand-drawn animation is produced at studios today. The vast majority of animation is produced at HD (1080 x 1920) or beyond resolution on digital drawing tablets or scanned in from pencil drawings. Once it is converted to a uniform line drawing (also known as a clean line), artists colorize by clicking on each individual line enclosure (segment) with a uniform color and flood filling it with a color. This painstakingly laborious process can take on the order of minutes per frame to do manually for complex animation.

The traditional style of flood filling each line enclosure has been around since the dawn of animation and continues to be the de facto standard because it allows the artist to quickly color many images in a short amount of time. For an assistive colorization tool to be effective in this domain, it is crucial that it integrates easily with this workflow and thus produce predictions at the level of segments. By doing so, this also enables the artist to manually intervene and correct any mistakes with the existing flood fill tool they are accustomed to.

\section{Pitfalls of pixel-based approaches}

One approach is to combine the output of a pixel-based model with flood fill segmentation information and choose the maximally occurring color in each segment. We explored this approach and highlight several issues that can occur.

\begin{figure}[h]
\centering
\includegraphics[width=0.9\linewidth]{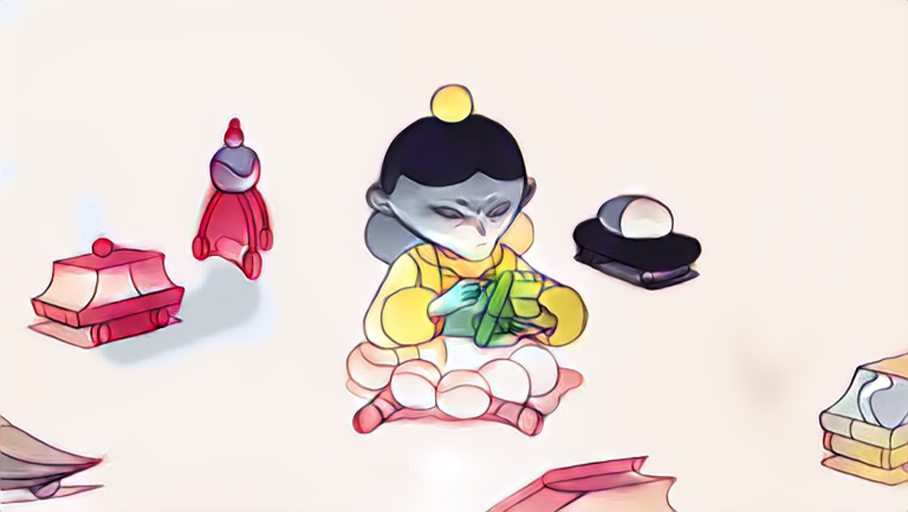}
\caption{\bf{Color mixing in PaintsChainer.}  }
\label{fig/paints_chainer}
\end{figure}

In Figure \ref{fig/paints_chainer} we use the popular open-source colorization model PaintsChainer and provide a color hint for every segment. The model is trained with an MSE loss in RGB space, so it learns to predict colors that are close to the user-provided color palette. When multiple colors are used, it quickly starts mixing the reference colors and diverging from the user-specified color palette.

\begin{figure}[h]
\centering
\includegraphics[width=0.9\linewidth]{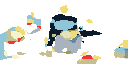}
\caption{\bf{Raw output of the correspondence subnetwork of DEVC.} }
\label{fig/devc}
\end{figure}

To overcome the color mixing issue we can train a pixel model with categorical cross entropy loss by discretizing the input color palette into a compact label space. We use this approach with the correspondence network in Deep Exemplar Video Colorization (see Figure \ref{fig/devc}). The resulting output stays true to the provided color palette, but the model loses important details due to input downsampling and max-pooling in the CNN backbone (both of which are necessary to compute pixel attention on a 16GB GPU). We use DEVC in our benchmarks, but convert the pixel output to segment labels for evaluation.

\section{Choice of evaluation metrics}

Given that our task is to output predictions at the level of segments, how do we measure performance? Existing metrics for pixel-based tracking and colorization tasks are not suitable: a practical metric would roughly approximate how many corrections an animator would need to make to correct any inaccurate predictions. Since the artists make corrections at the level of segments, this begs the need for segment-level evaluation metrics. Thus, we define two evaluation metrics that are specifically suited for the task: \textbf{Accuracy} and \textbf{Mean IoU}. We describe each of these in detail and discuss their connection with other evaluation metrics in colorization and tracking.

\PAR{Accuracy} is defined as the percentage of correct segment-level label predictions averaged over all segments in each of the target sequences. In colorization, this is somewhat analogous to MSE in RGB color space -- we want to predict the right color label and penalize incorrect colors. However, unlike in photo-realistic colorization we are predicting from a discrete set of labels. 

\PAR{Mean IoU} is defined as the mean Intersection-over-Union for each segment averaged over all segments in the target sequence. In the video segmentation context, our Mean IoU metric is analogous to Region Similarity $\mathcal{J}$. However, instead of measuring the similarity between pixel regions we are measuring in the level of segments.

\section{User study}
\begin{figure*}[h]
\centering
\includegraphics[width=1\linewidth]{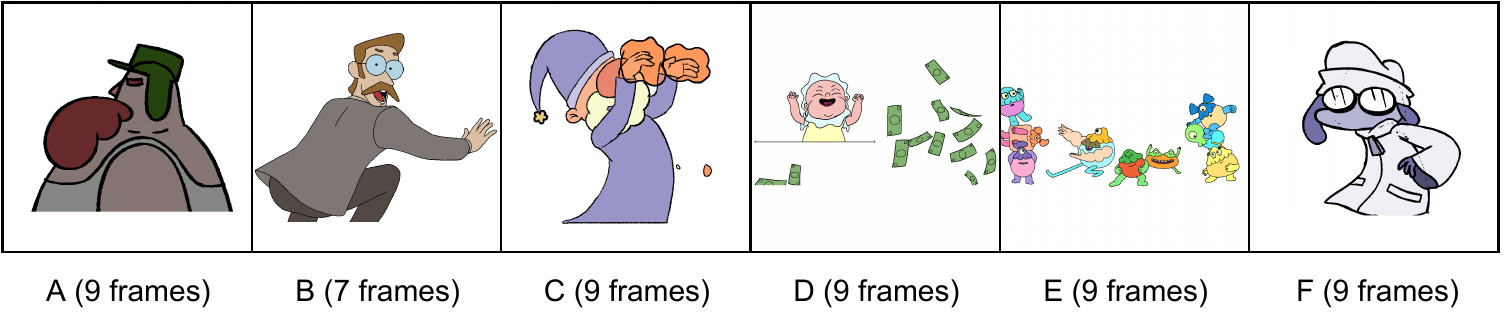}
\caption{ \bf{Samples from sequences used in the user study.} }
\label{fig/userstudycase}
\end{figure*}

To evaluate our approach we conducted a user study. We ask professional artists to colorize sequences (see  Figure \ref{fig/userstudycase}) from the real dataset without and with the assistance from AnT. In the test with Ant, we colorized the sequences with AnT then asked users to check and correct incorrect parts in the results. All tests were done in professional software, and we record users’ interactions and work time. The summary result is shown in Table \ref{tab:userstudy}. We can see that AnT significantly increases the work efficiency. 
\begin{table*}[h]
  \centering
  \resizebox{2\columnwidth}{!}{%
\begin{tabular}{ccccccccccc}
\toprule
\multirow{2}{*}{\bf{Case}} & \multicolumn{4}{c}{\bf{Human}}           & \multicolumn{4}{c}{\bf{AnT+Human}}       & \multirow{2}{*}{\bf{\begin{tabular}[c]{@{}c@{}}Interactions\\ (AnT+Human / Human)\end{tabular}}} & \multirow{2}{*}{\bf{\begin{tabular}[c]{@{}c@{}}Time\\ (AnT+Human / Human)\end{tabular}}} \\
                               & Mouse click & Key down & Interactions & Time (s) & Mouse click & Key down & Interactions & Time (s) &                                                                                                 &                                                                                            \\
\midrule
A                              & 180         & 629      & 809       & 174.10  & 31          & 16       & 47        & 56.16   & 5.81\%                                                                                          & 32.26\%                                                                                    \\
B                              & 402         & 1013     & 1415      & 429.53  & 93          & 206      & 299       & 119.57  & 21.13\%                                                                                         & 27.84\%                                                                                    \\
C                              & 365         & 1497     & 1862      & 369.09  & 131         & 700      & 831       & 140.85  & 44.63\%                                                                                         & 38.16\%                                                                                    \\
D                              & 605         & 897      & 1502      & 550.23  & 138         & 551      & 689       & 169.93  & 45.87\%                                                                                         & 30.88\%                                                                                    \\
E                              & 2151        & 5058     & 7209      & 1826.52 & 90          & 167      & 257       & 169.70  & 3.56\%                                                                                          & 9.29\%                                                                                     \\
F                              & 280         & 849      & 1129      & 237.22  & 79          & 270      & 349       & 91.41   & 30.91\%                                                                                         & 38.53\%                                                                                   
\\
\bottomrule
\end{tabular}
  }
    \caption{\label{tab:userstudy}
    {\bf User study result.} Comparison of user effort on colorization task without/with assistance from AnT. "Mouse click" interactions mainly include switching, moving, and zooming the canvas, filling, and picking colors. "Key down" interactions mainly include toggling tools, file operations, undo/redo.
  }
\end{table*}

\section{Qualitative results}

\PAR{Comparison with other methods}
In this section, we show results of our proposed approach (AnT) to: DEVC, Lazy Brush, EBSynth, Style2paints. LazyBrush fails to handle large movements but fills segments with a uniform color, making it suitable for animation workflows. EBSynth similarly degrades with large movements but is not segment-aware so it blends pixels together. Style2paints is not suitable for animation colorization task.

\begin{figure*}
\centering
\resizebox{0.9\linewidth}{!}{%
\begin{tabular}{c|c|cccc}
  
  & Reference & \multicolumn{4}{c}{Colorized Sequence} \\
 
  \hline
  
  \raisebox{3\normalbaselineskip}[0pt][0pt]{\rotatebox[origin=c]{90}{AnT}} &
  \raisebox{0\normalbaselineskip}[0pt][0pt]{\includegraphics[width=0.2\linewidth]{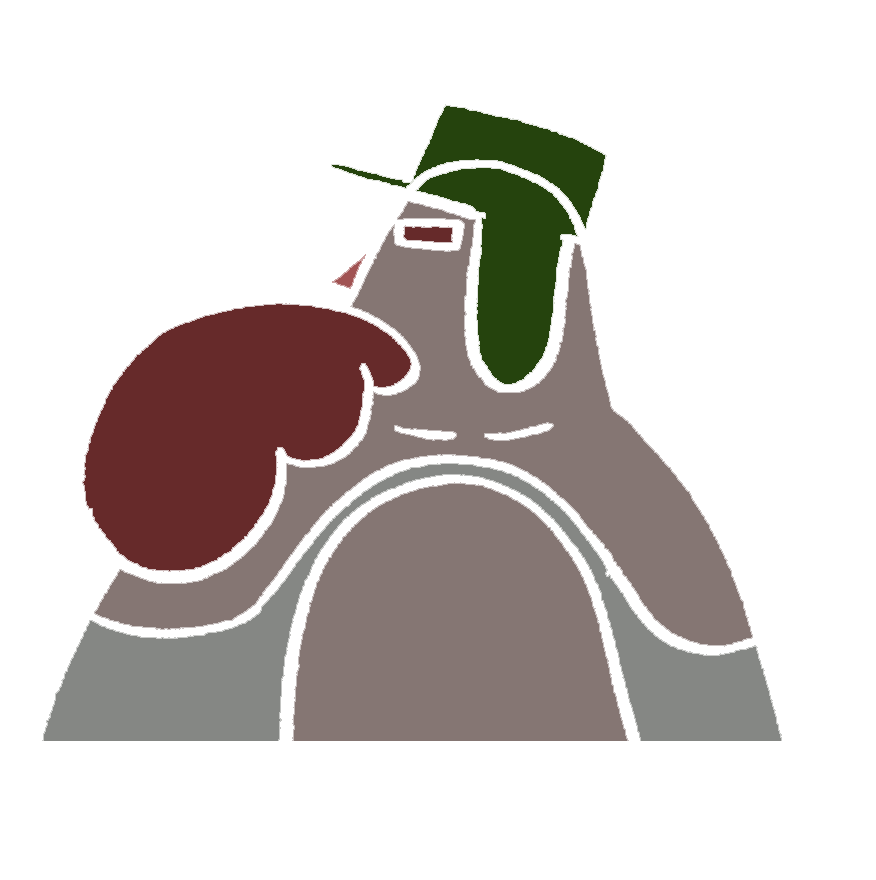}} &
  \includegraphics[width=0.2\linewidth]{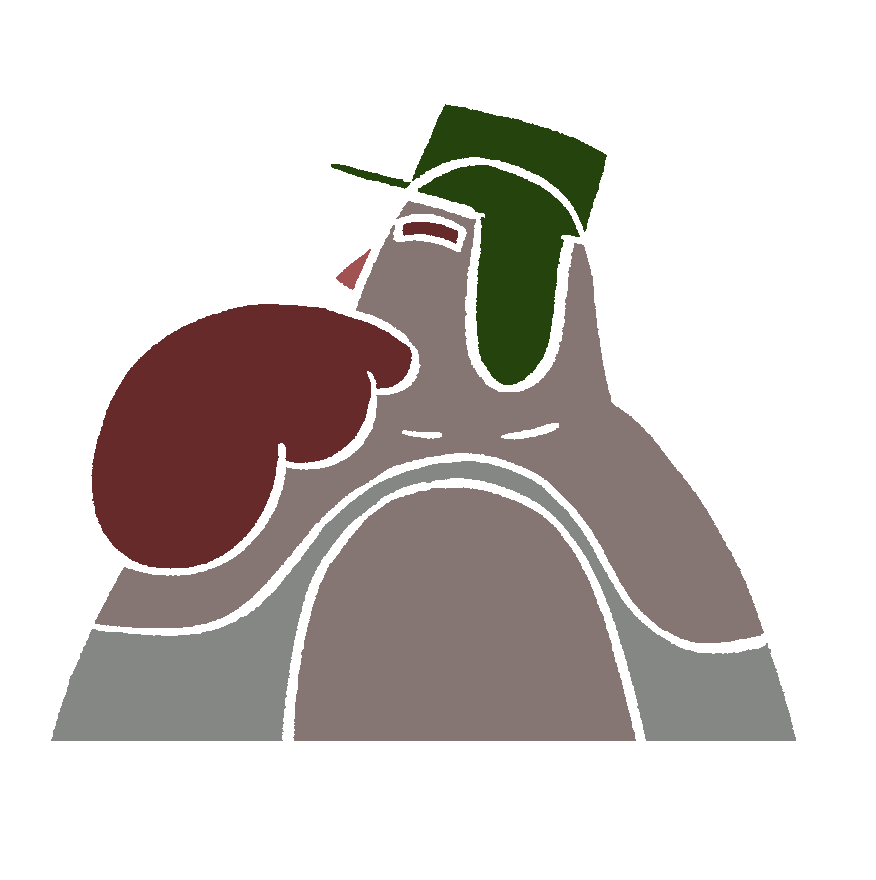} &
  \includegraphics[width=0.2\linewidth]{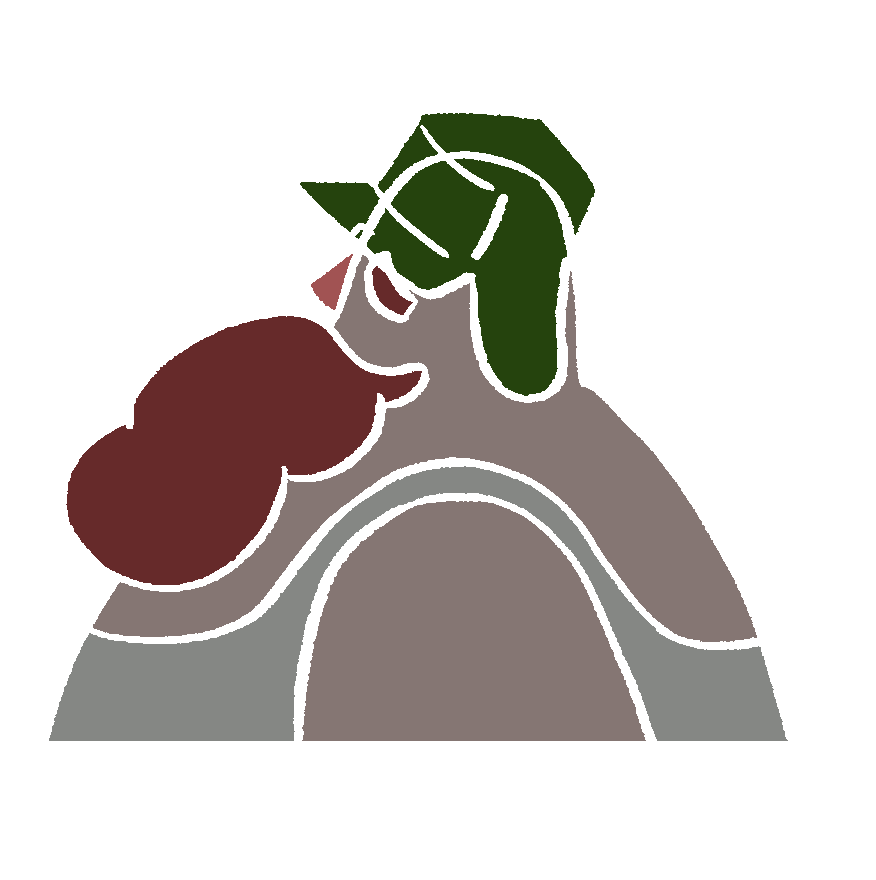} &
  \includegraphics[width=0.2\linewidth]{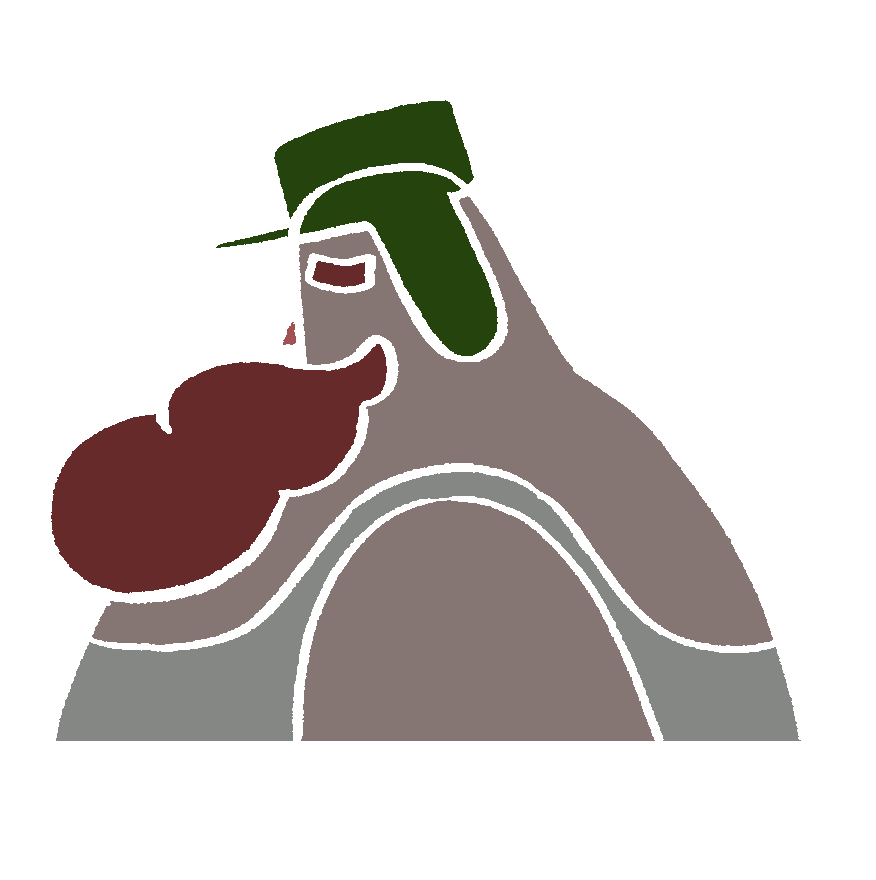} &
  \includegraphics[width=0.2\linewidth]{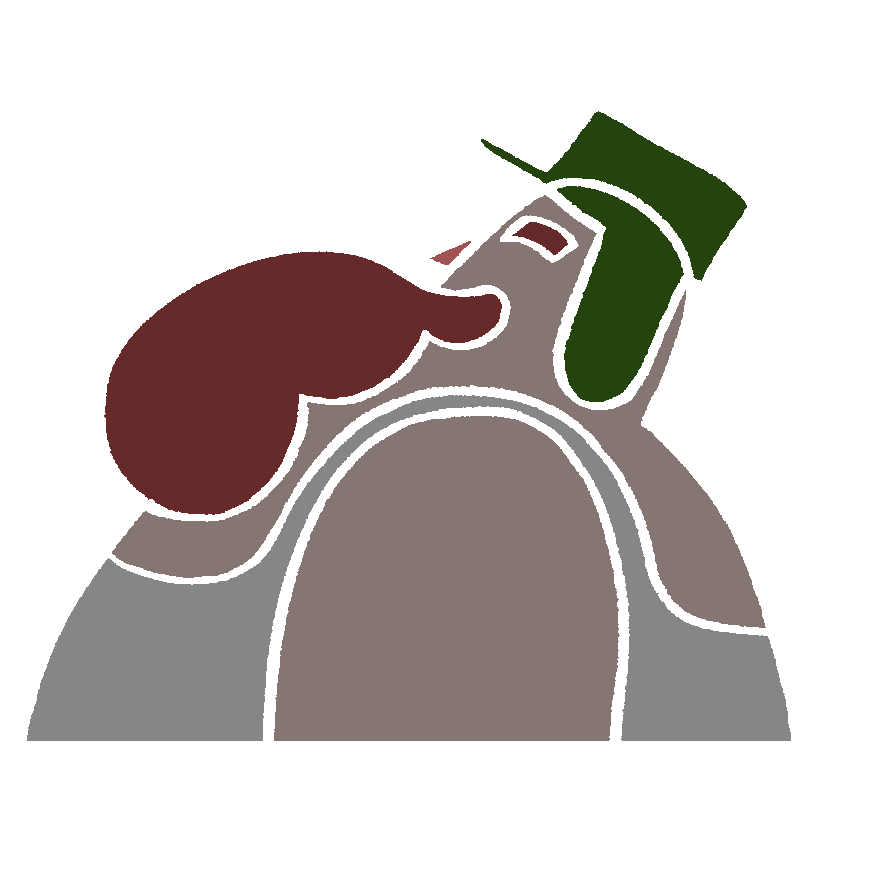} \\
  
  \raisebox{3\normalbaselineskip}[0pt][0pt]{\rotatebox[origin=c]{90}{EBSynth}} &
  \raisebox{0\normalbaselineskip}[0pt][0pt]{\includegraphics[width=0.2\linewidth]{figures/SH240/eb/e0}} &
  \includegraphics[width=0.2\linewidth]{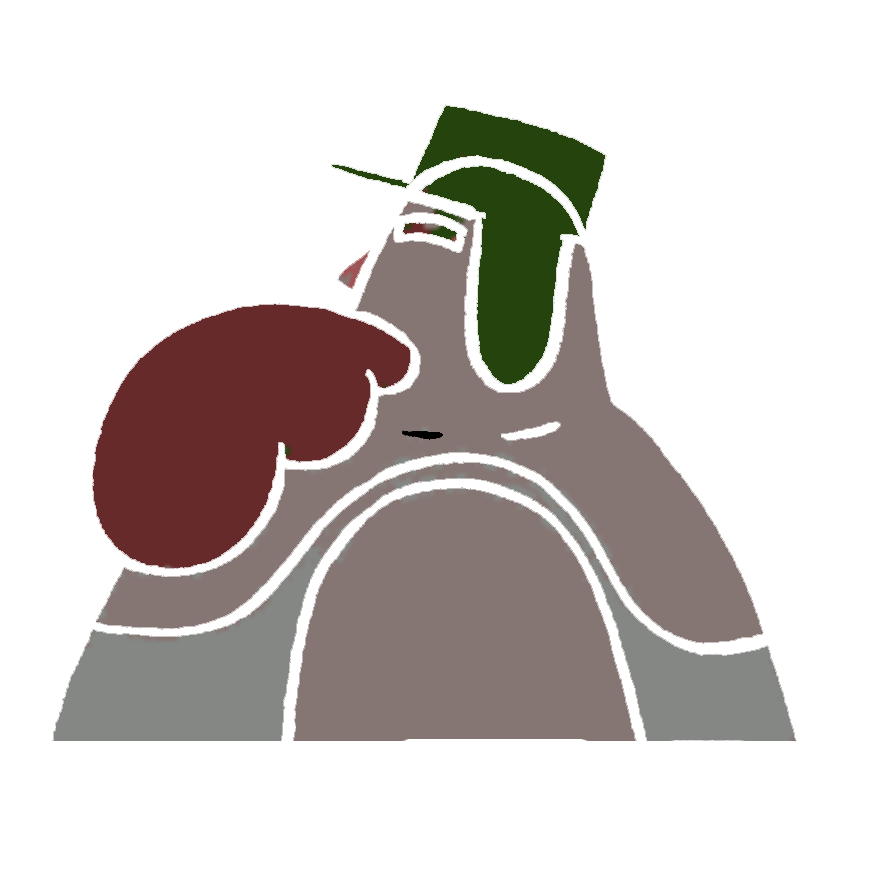} &
  \includegraphics[width=0.2\linewidth]{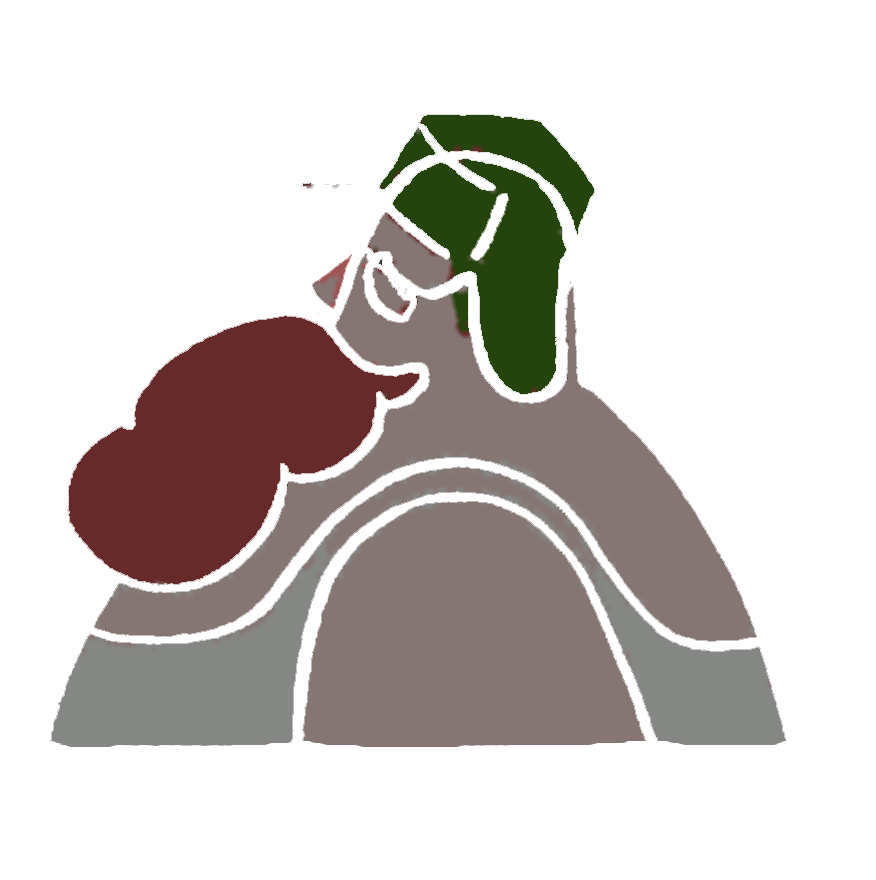} &
  \includegraphics[width=0.2\linewidth]{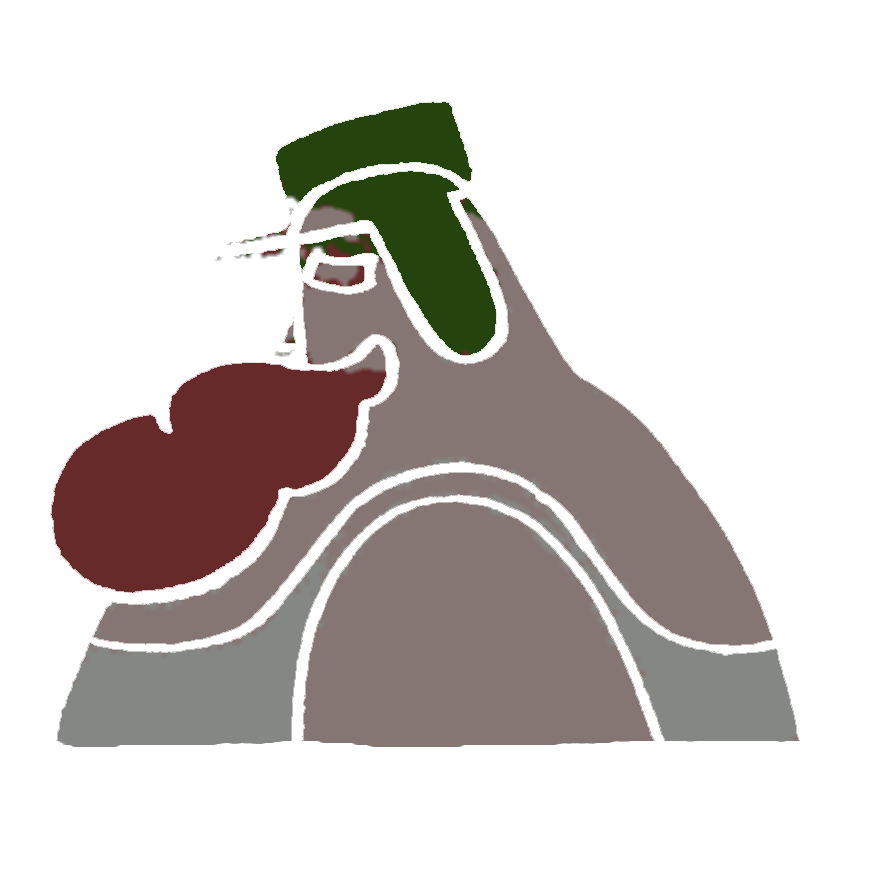} &
  \includegraphics[width=0.2\linewidth]{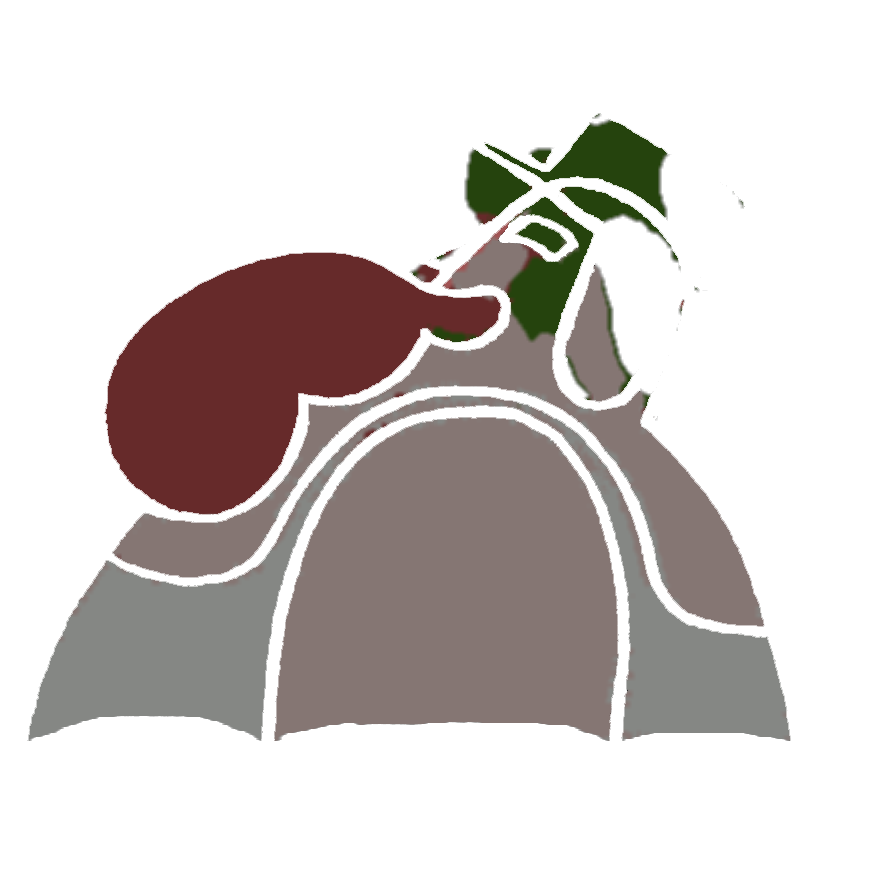} \\
  
  \raisebox{3\normalbaselineskip}[0pt][0pt]{\rotatebox[origin=c]{90}{LazyBrush}} &
  \raisebox{0\normalbaselineskip}[0pt][0pt]{\includegraphics[width=0.2\linewidth]{figures/SH240/eb/e0}} &
  \includegraphics[width=0.2\linewidth]{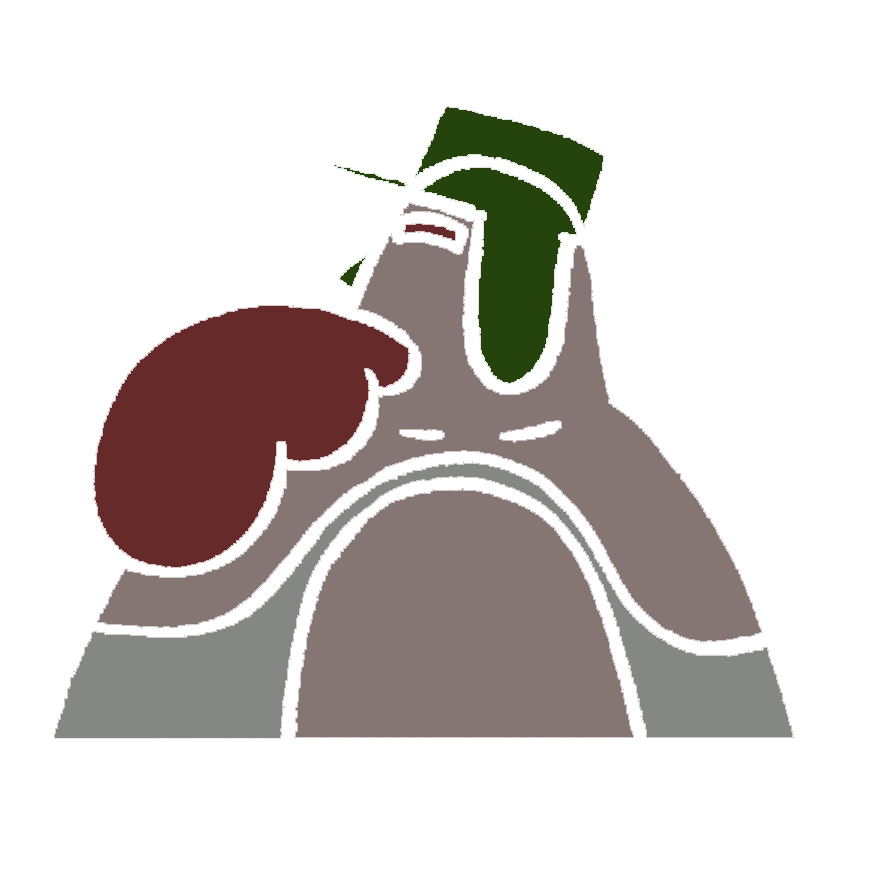} &
  \includegraphics[width=0.2\linewidth]{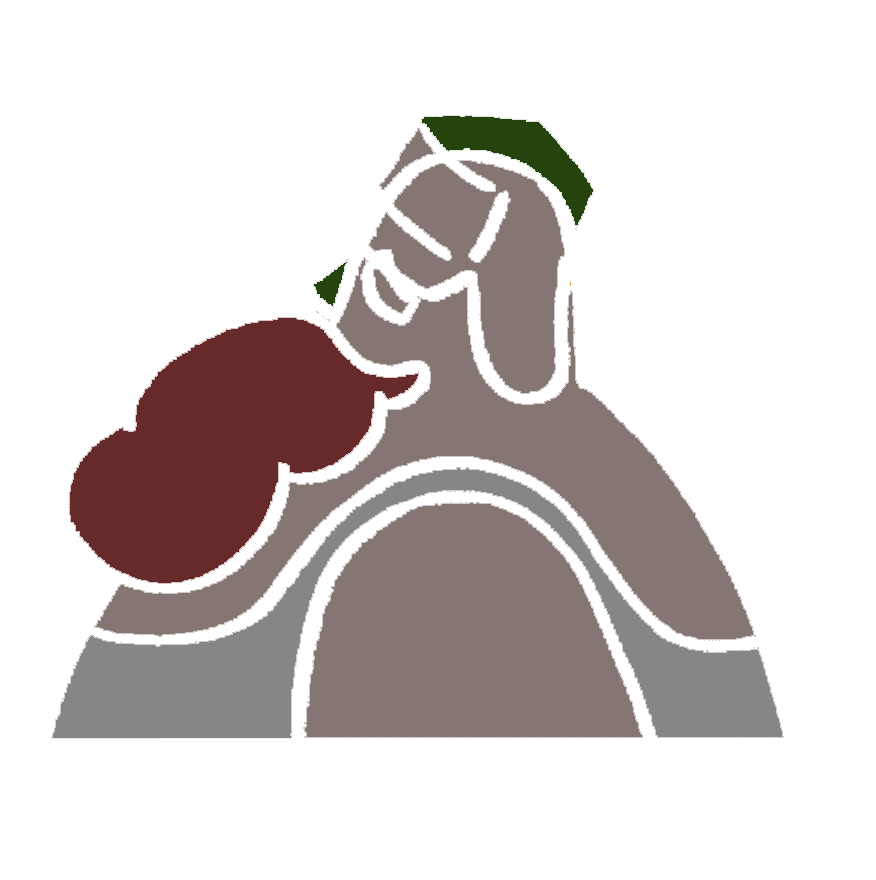} &
  \includegraphics[width=0.2\linewidth]{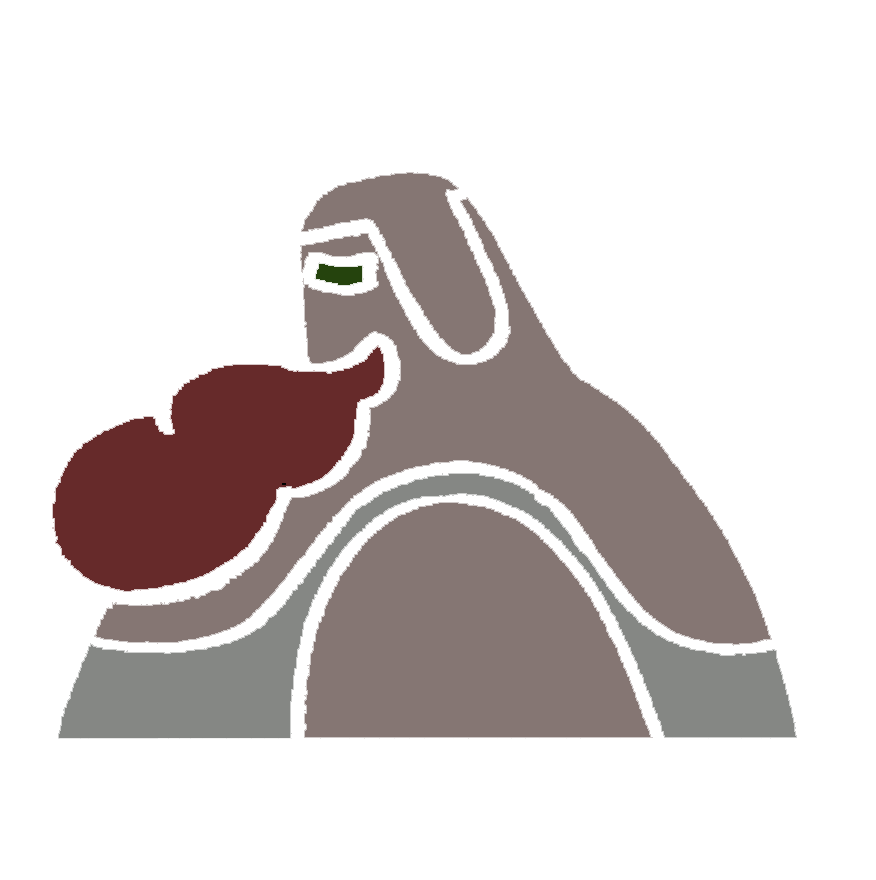} &
  \includegraphics[width=0.2\linewidth]{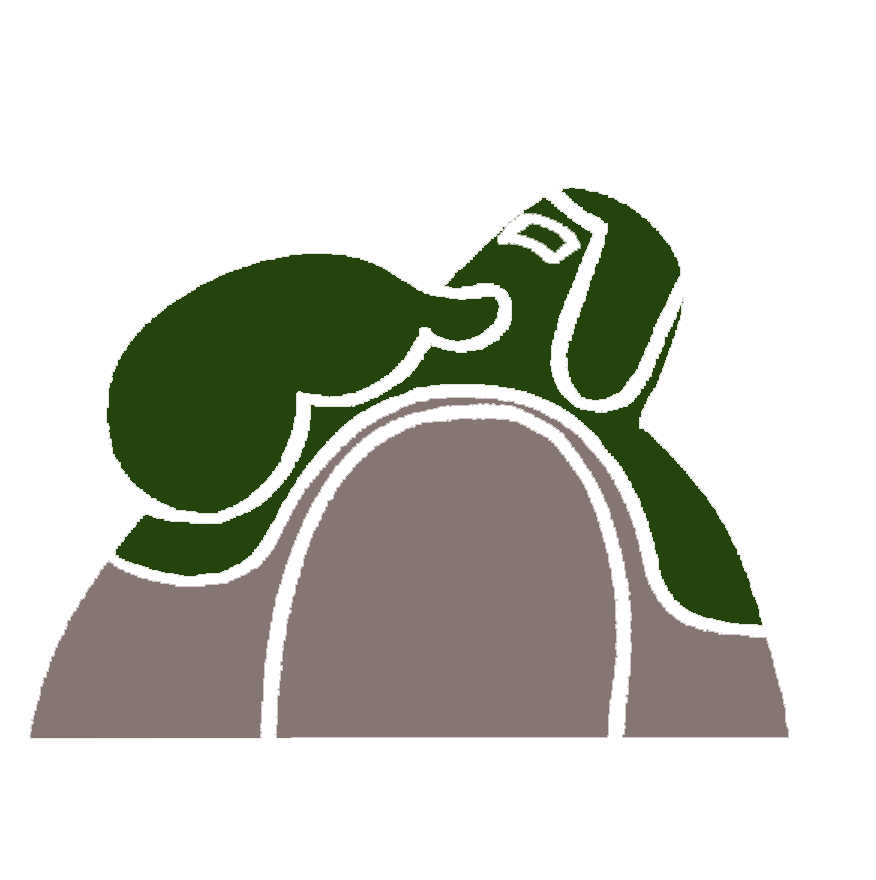} \\

  \raisebox{3\normalbaselineskip}[0pt][0pt]{\rotatebox[origin=c]{90}{Style2paints (exemplar)}} &
  \raisebox{0\normalbaselineskip}[0pt][0pt]{\includegraphics[width=0.2\linewidth]{figures/SH240/eb/e0}} &
  \includegraphics[width=0.2\linewidth]{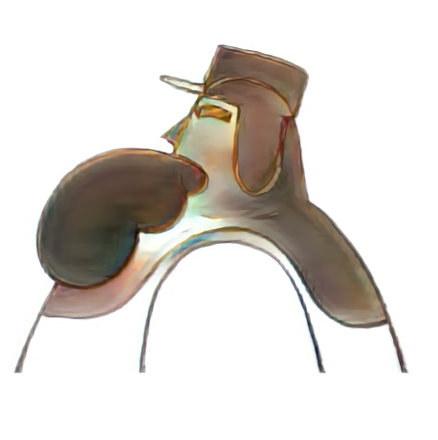} &
  \includegraphics[width=0.2\linewidth]{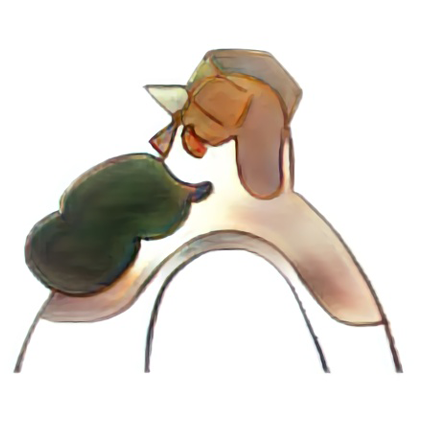} &
  \includegraphics[width=0.2\linewidth]{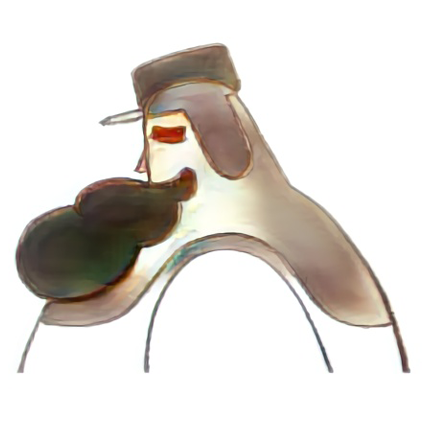} &
  \includegraphics[width=0.2\linewidth]{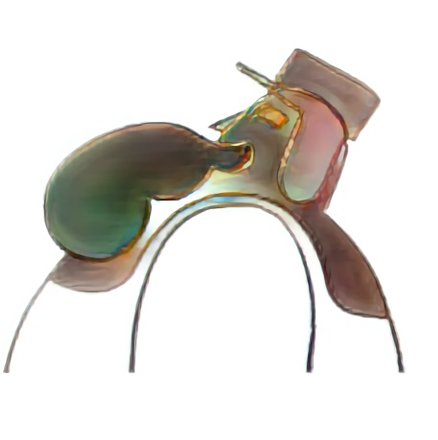} \\
  
  \raisebox{3\normalbaselineskip}[0pt][0pt]{\rotatebox[origin=c]{90}{Style2paints (hint)}} &
  \raisebox{0\normalbaselineskip}[0pt][0pt]{\includegraphics[width=0.2\linewidth]{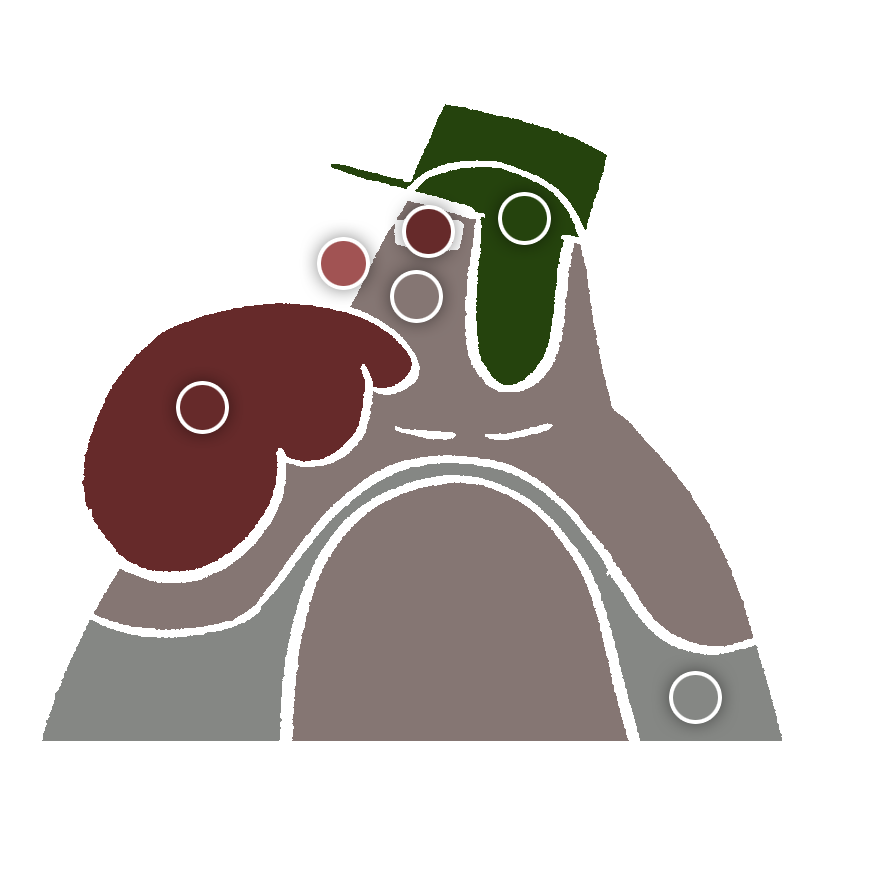}} &
  \includegraphics[width=0.2\linewidth]{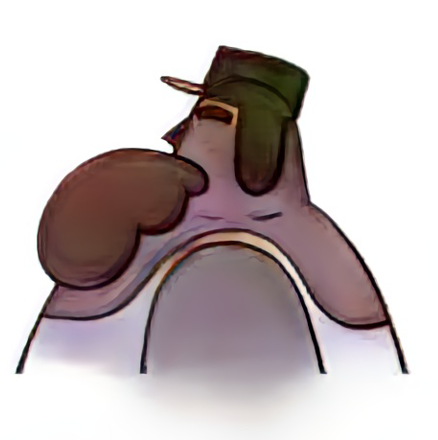} &
  \includegraphics[width=0.2\linewidth]{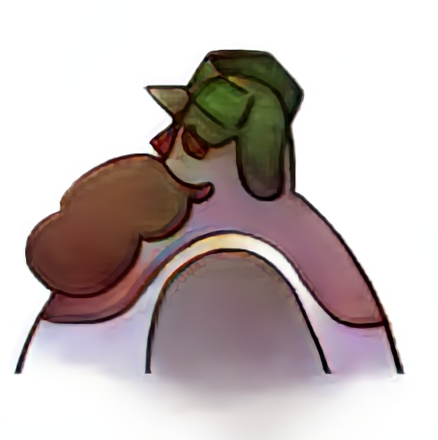} &
  \includegraphics[width=0.2\linewidth]{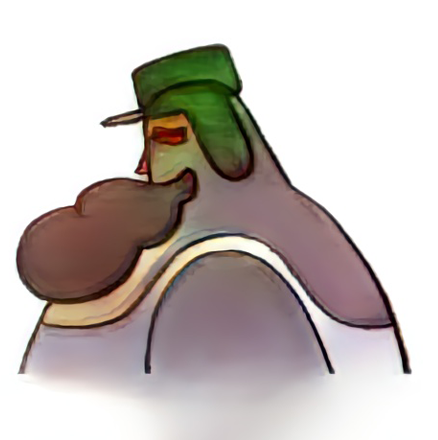} &
  \includegraphics[width=0.2\linewidth]{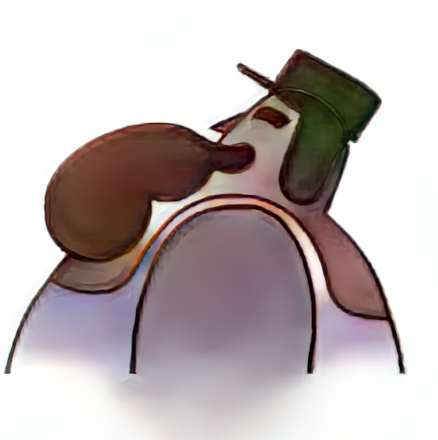} \\
  
  \hline
\end{tabular}
}
\caption{\bf{Comparison with other methods.} }
\label{fig/compqual}
\end{figure*}

\begin{figure*}
\centering
\resizebox{0.85\linewidth}{!}{%
\begin{tabular}{c|c|cccc}
  
  & Reference & \multicolumn{4}{c}{Colorized Sequence} \\
 
  \hline
  
  \raisebox{3\normalbaselineskip}[0pt][0pt]{\rotatebox[origin=c]{90}{DEVC}} &
  \raisebox{0\normalbaselineskip}[0pt][0pt]{\includegraphics[width=0.2\linewidth]{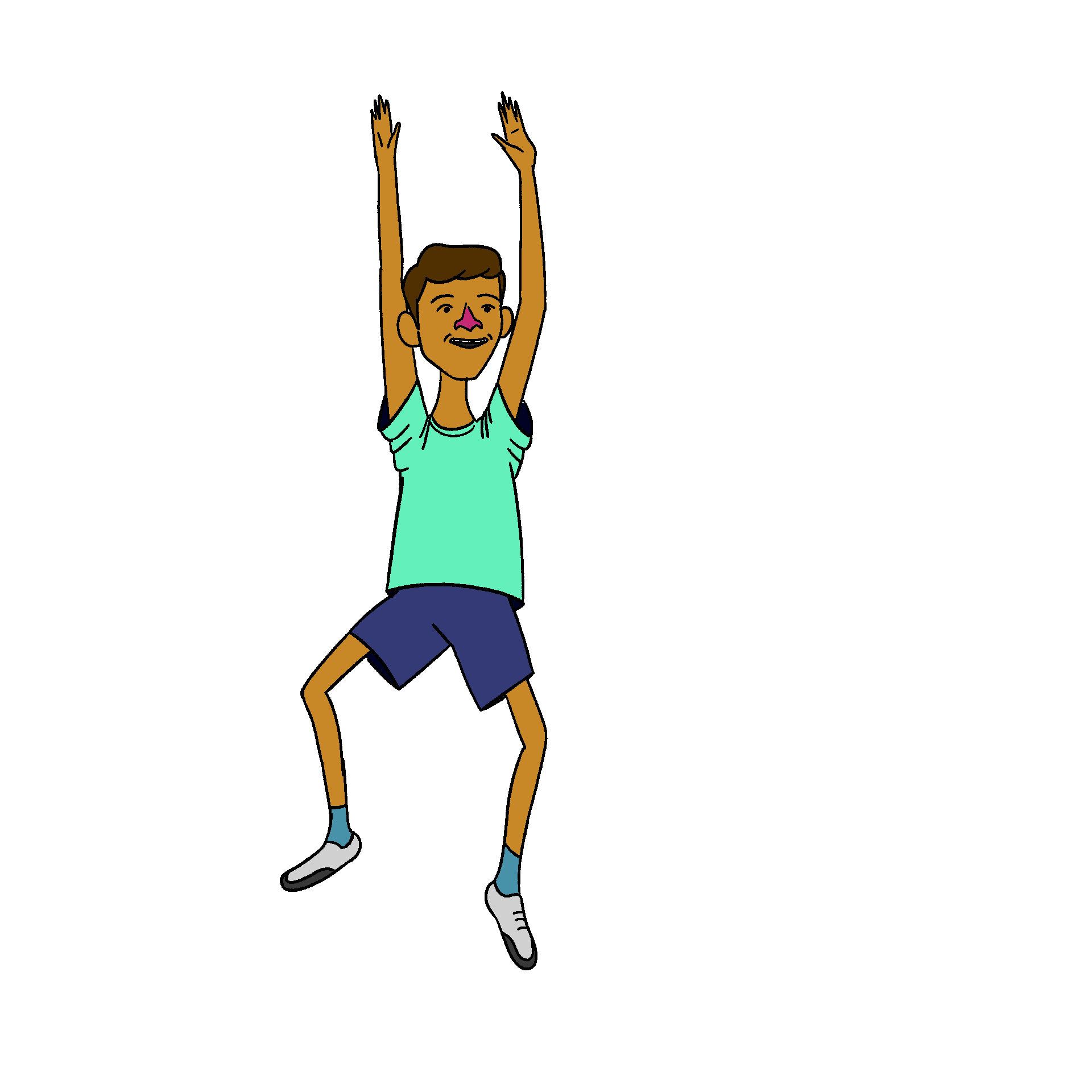}} &
  \includegraphics[width=0.2\linewidth]{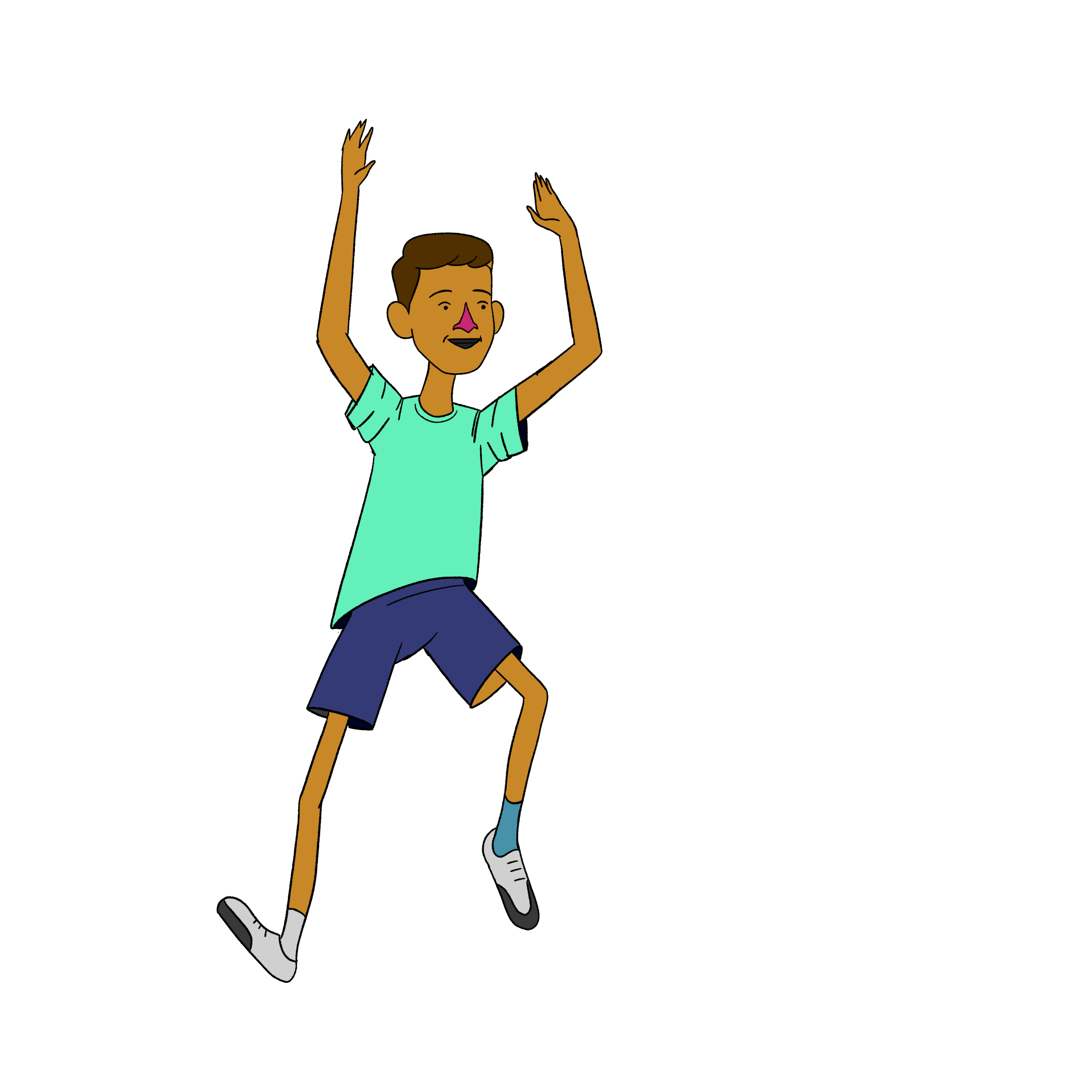} &
  \includegraphics[width=0.2\linewidth]{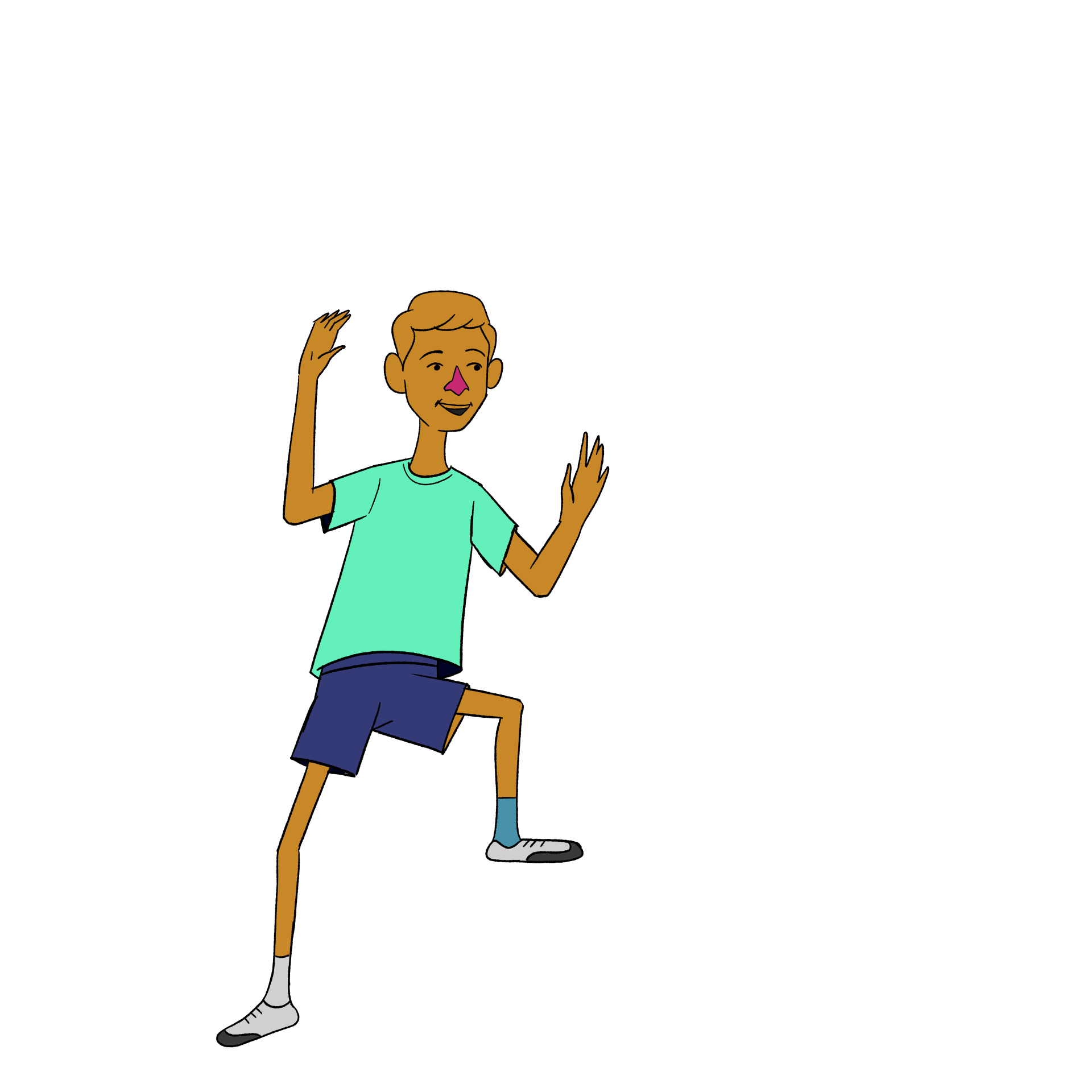} &
  \includegraphics[width=0.2\linewidth]{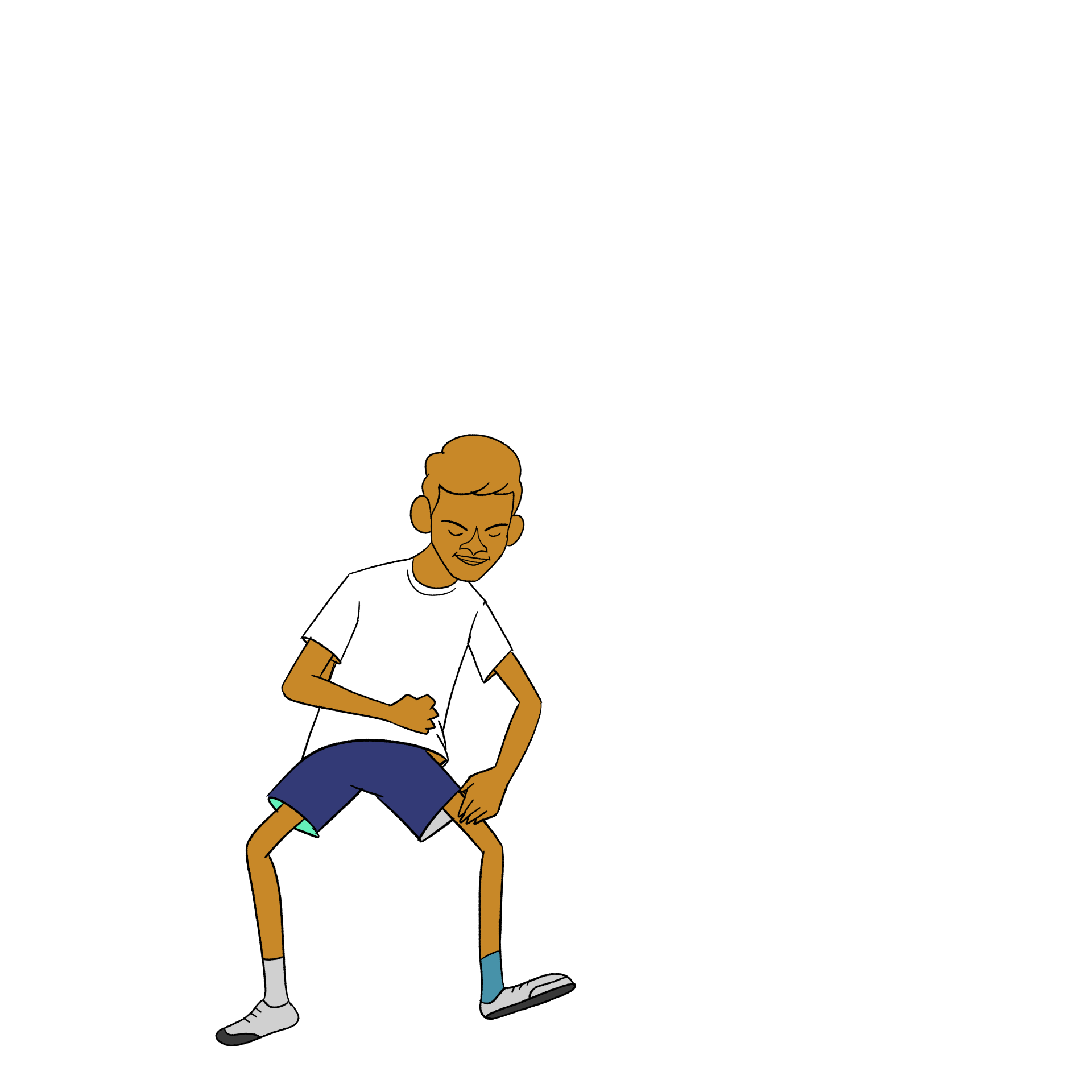} &
  \includegraphics[width=0.2\linewidth]{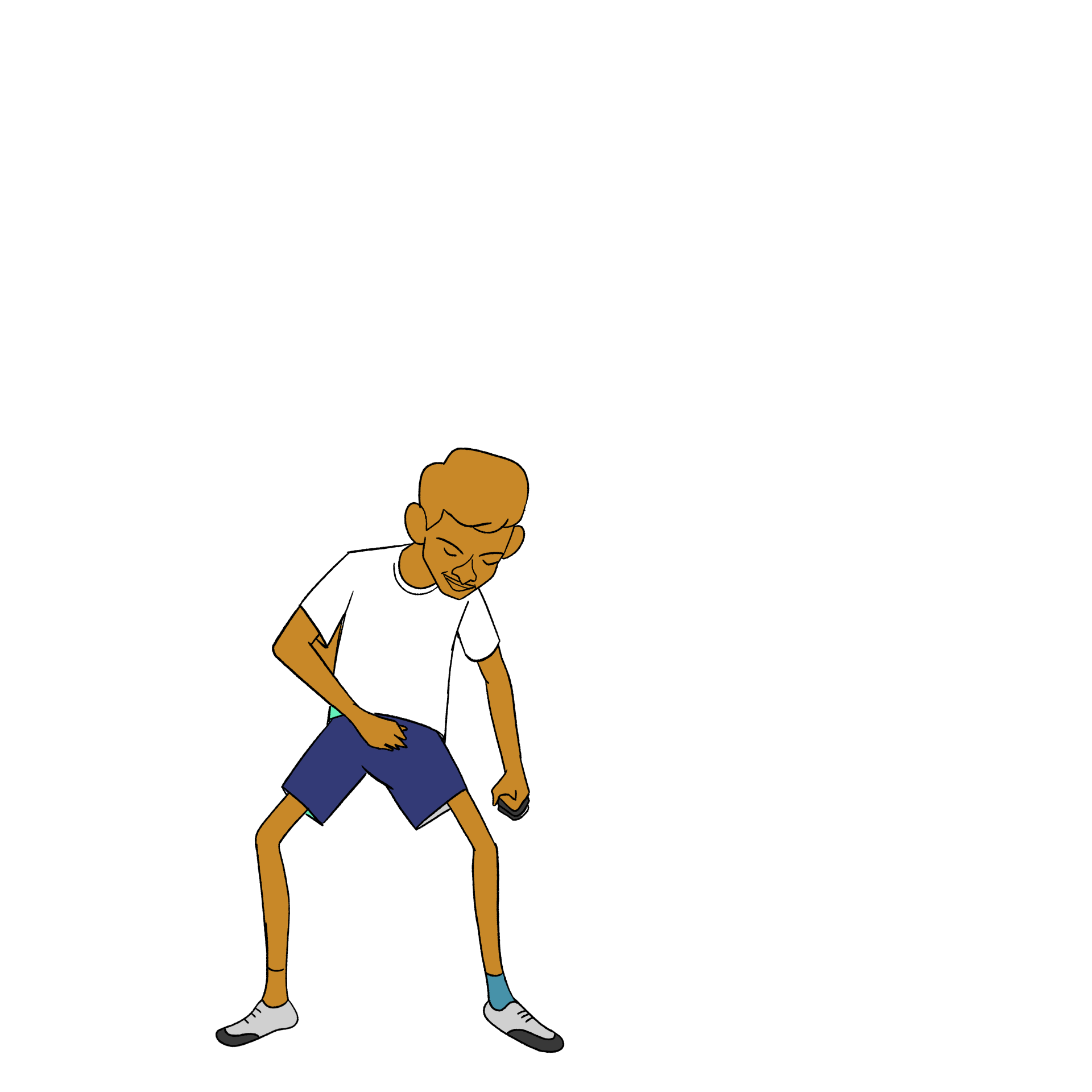} \\
  
  \raisebox{3\normalbaselineskip}[0pt][0pt]{\rotatebox[origin=c]{90}{AnT}} &
  \raisebox{0\normalbaselineskip}[0pt][0pt]{\includegraphics[width=0.2\linewidth]{figures/sh010-30/ref}} &
  \includegraphics[width=0.2\linewidth]{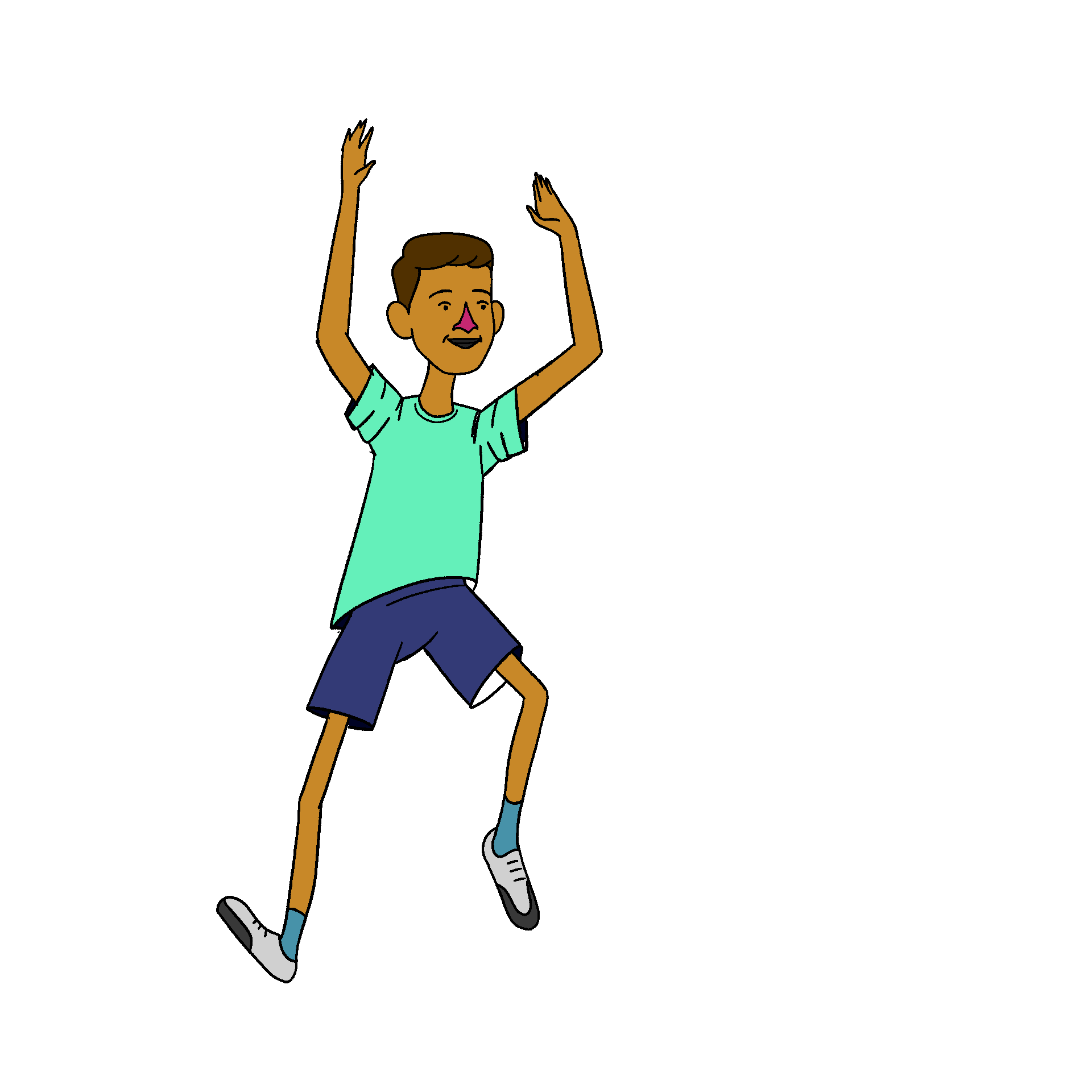} &
  \includegraphics[width=0.2\linewidth]{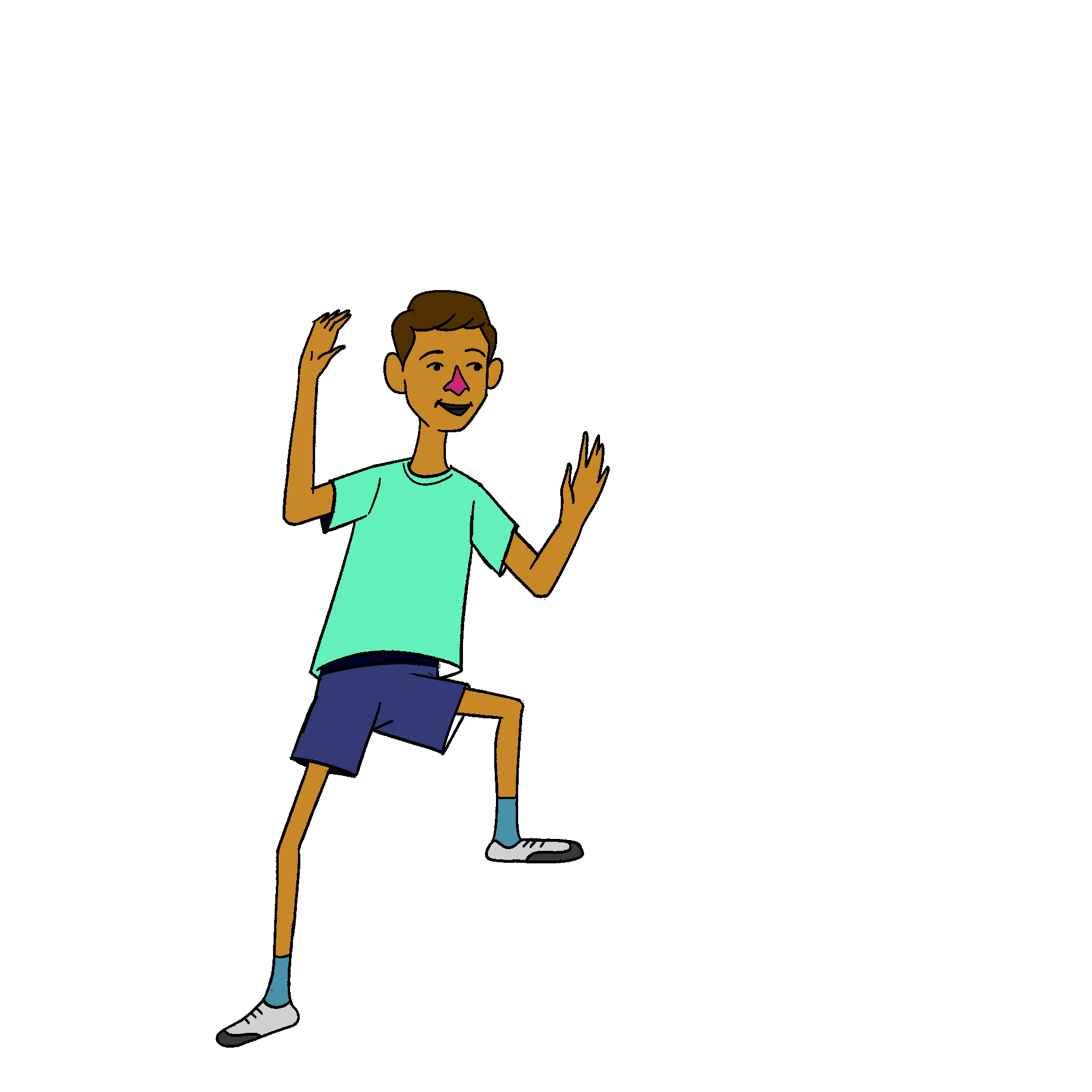} &
  \includegraphics[width=0.2\linewidth]{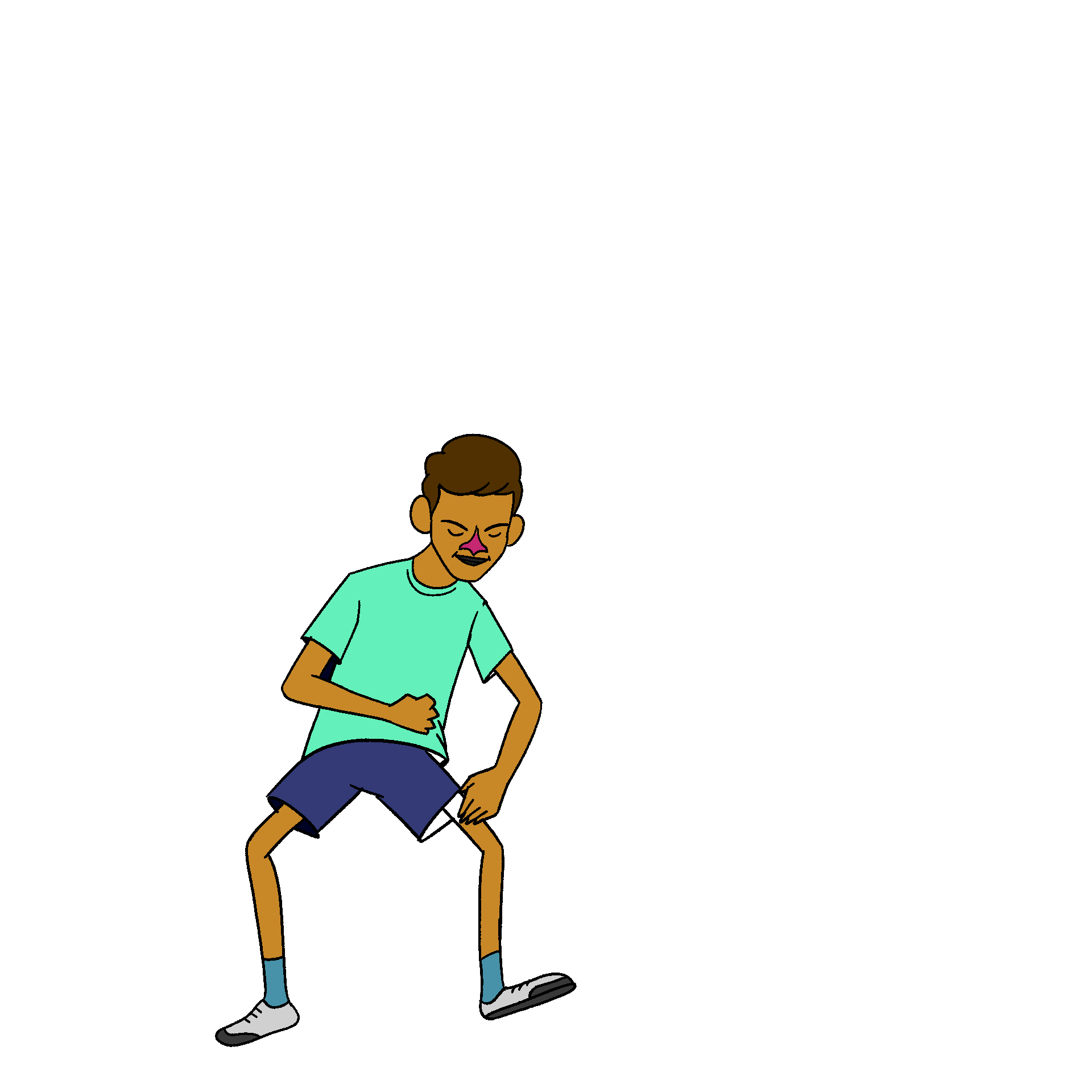} &
  \includegraphics[width=0.2\linewidth]{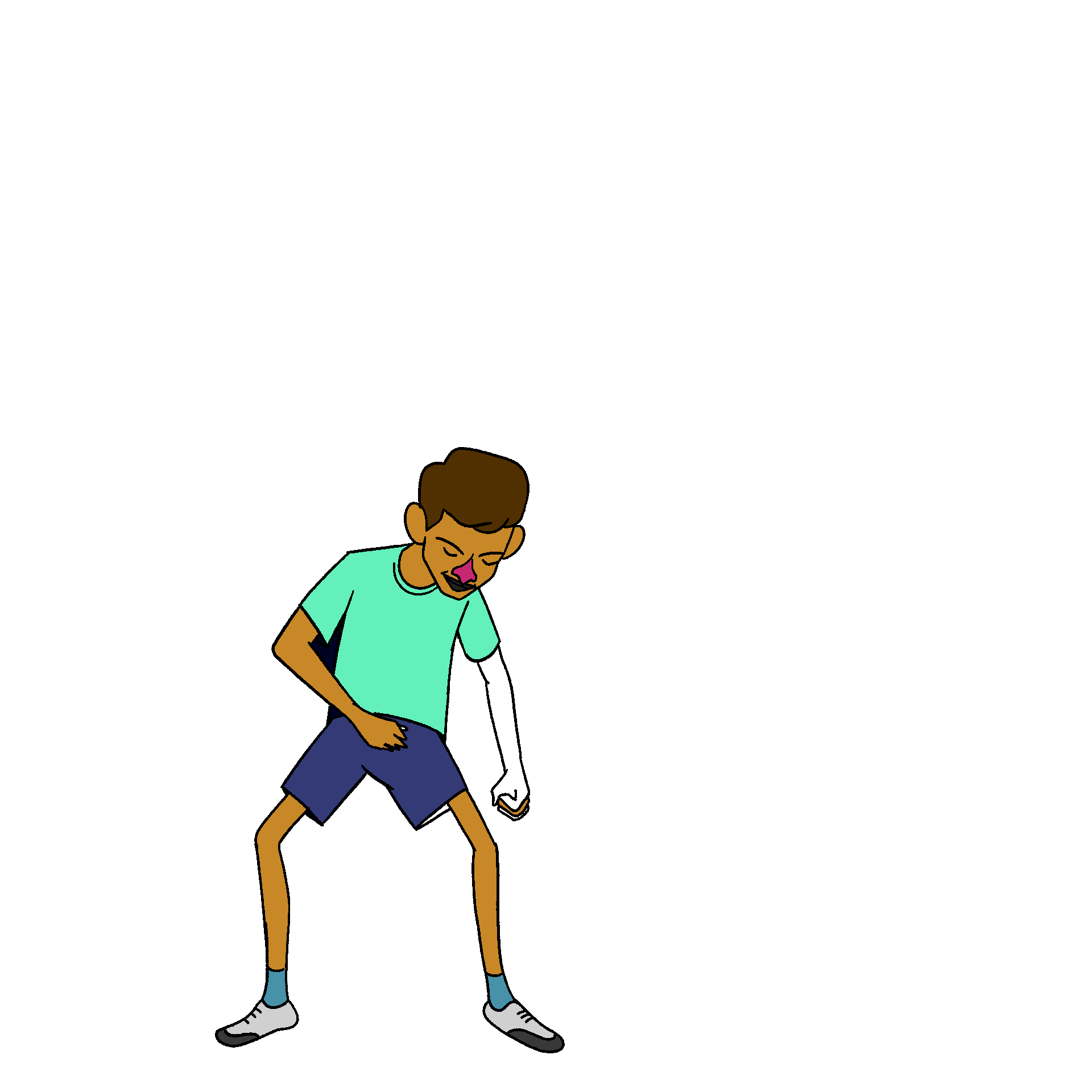} \\

  \hline
  
  \raisebox{3\normalbaselineskip}[0pt][0pt]{\rotatebox[origin=c]{90}{DEVC}} &
  \raisebox{0\normalbaselineskip}[0pt][0pt]{\includegraphics[width=0.2\linewidth]{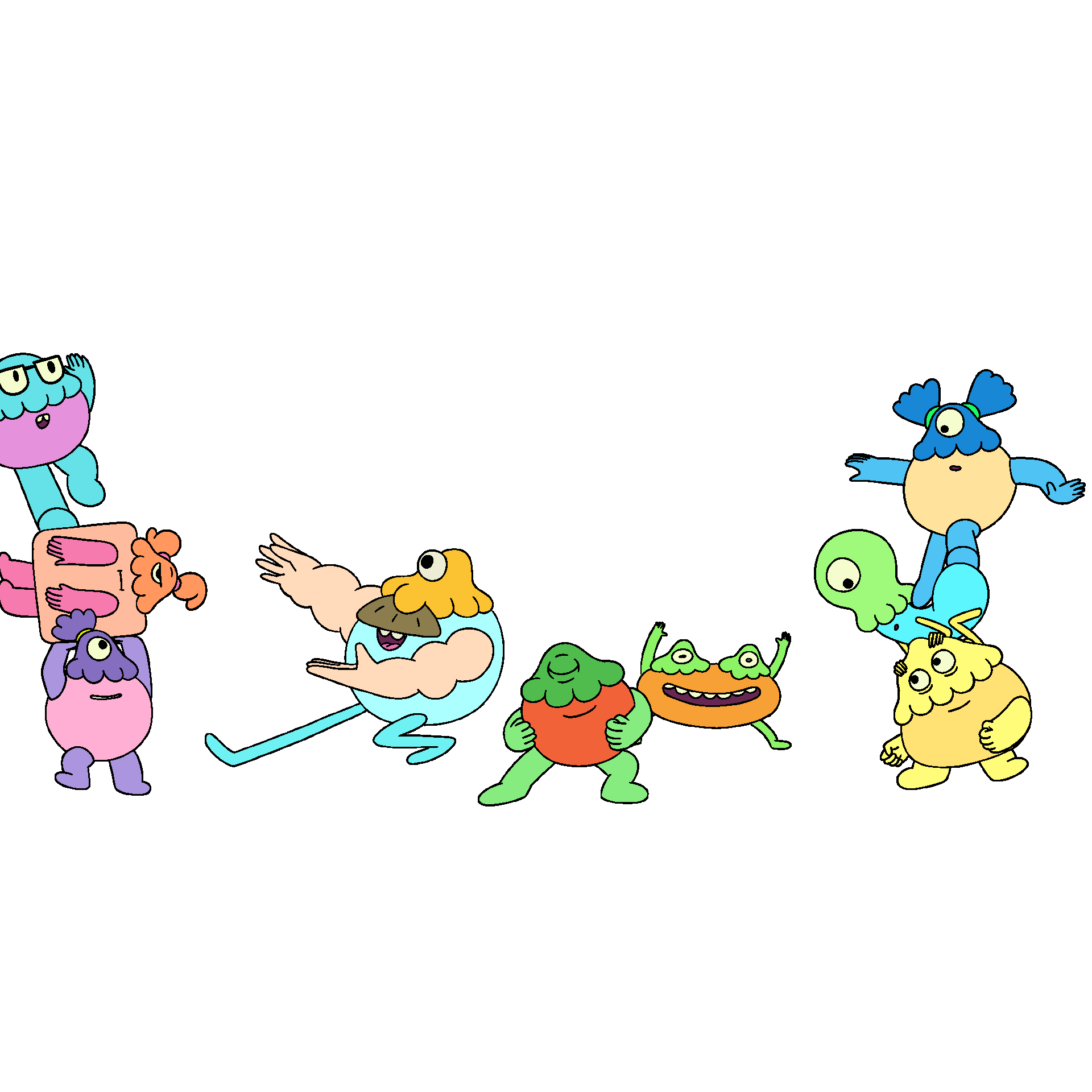}} &
  \includegraphics[width=0.2\linewidth]{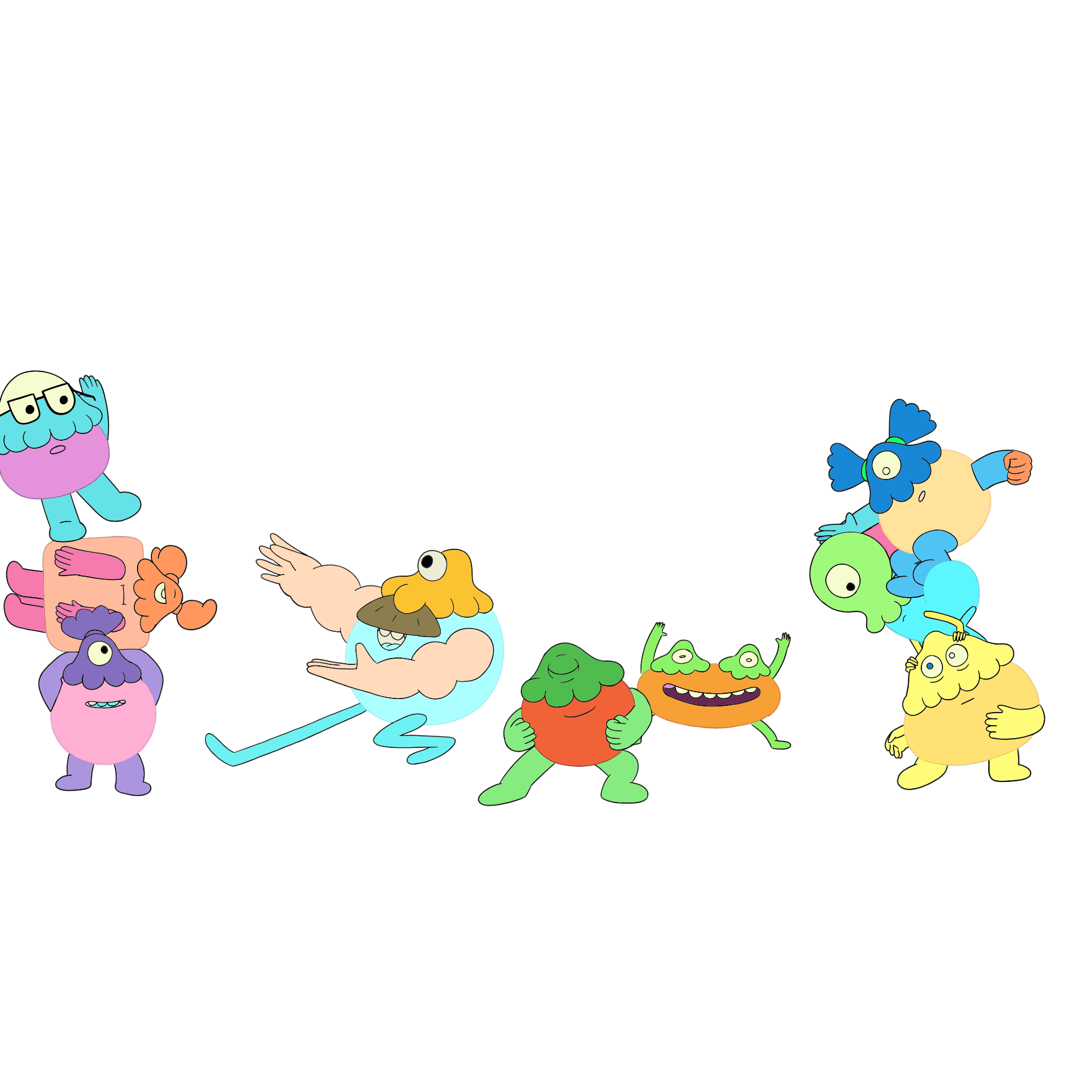} &
  \includegraphics[width=0.2\linewidth]{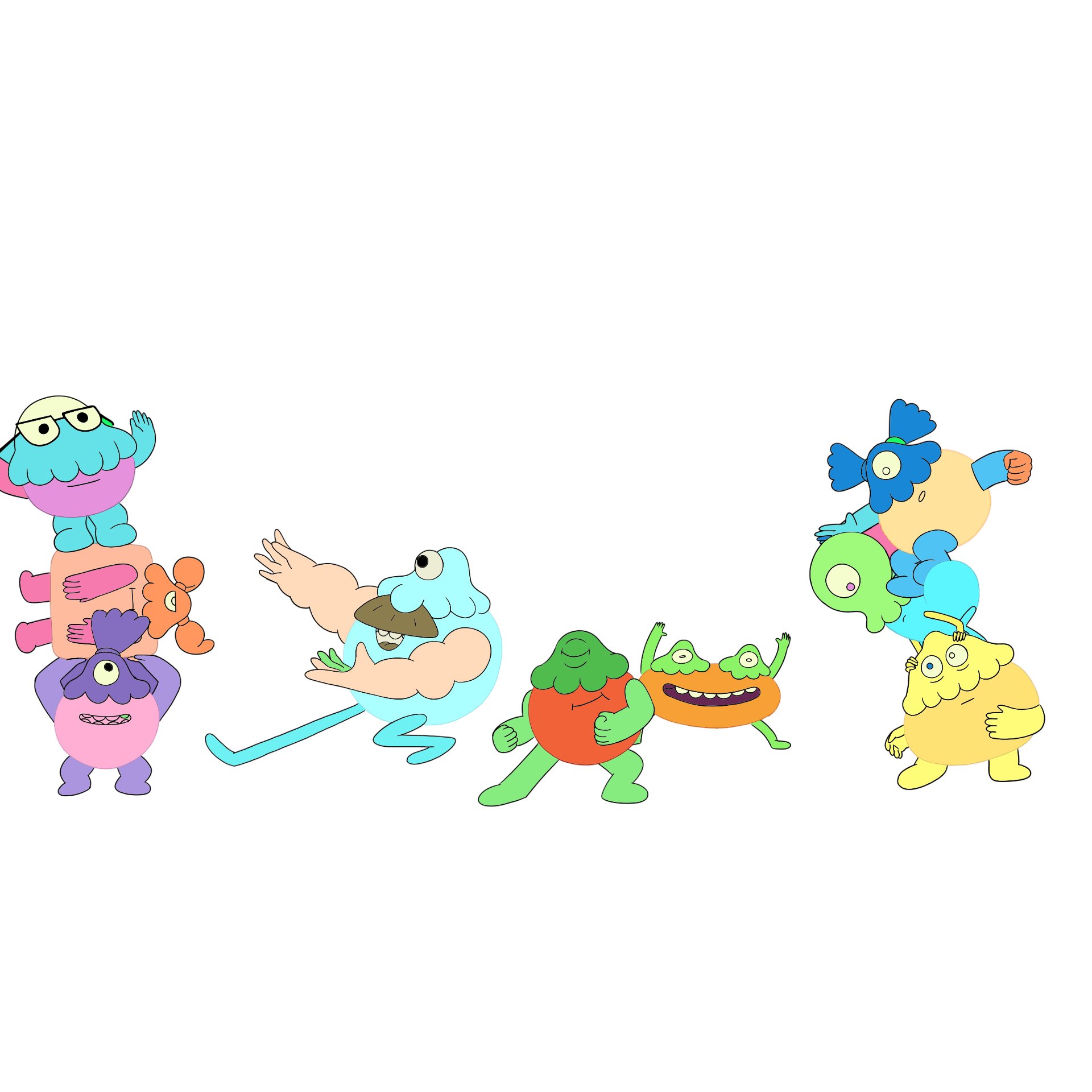} &
  \includegraphics[width=0.2\linewidth]{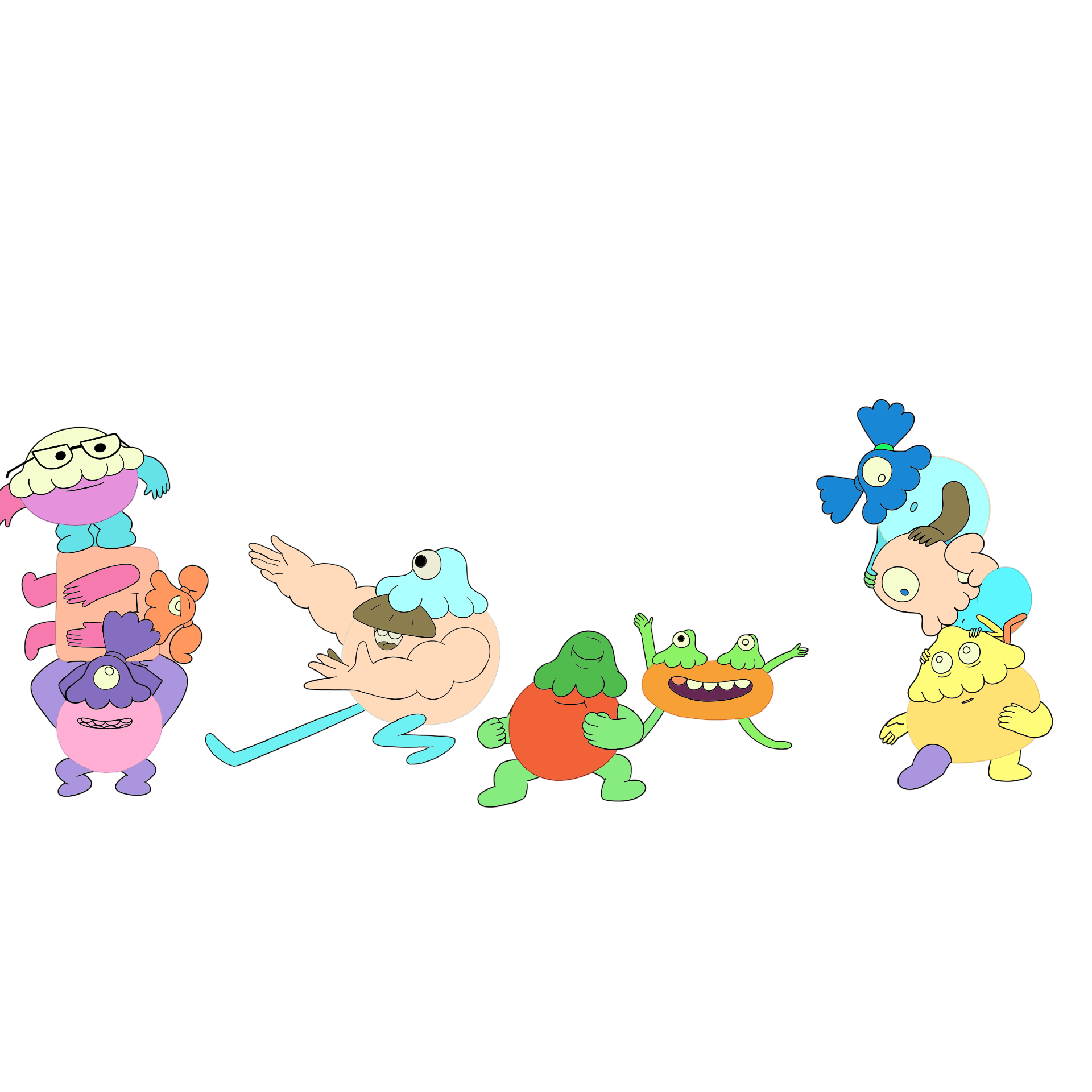} &
  \includegraphics[width=0.2\linewidth]{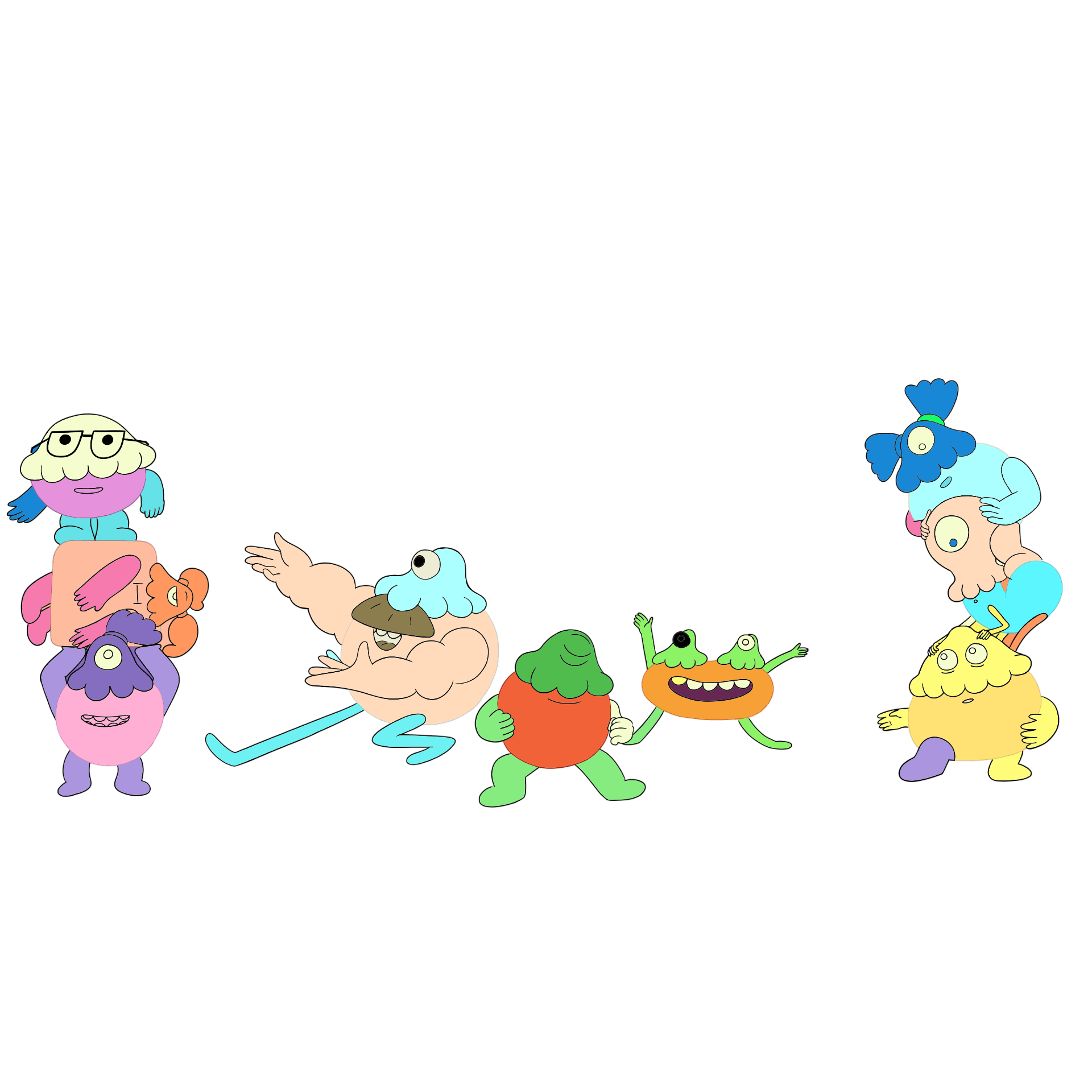} \\
  
  \raisebox{3\normalbaselineskip}[0pt][0pt]{\rotatebox[origin=c]{90}{AnT}} &
  \raisebox{0\normalbaselineskip}[0pt][0pt]{\includegraphics[width=0.2\linewidth]{figures/sh11-30/ref}} &
  \includegraphics[width=0.2\linewidth]{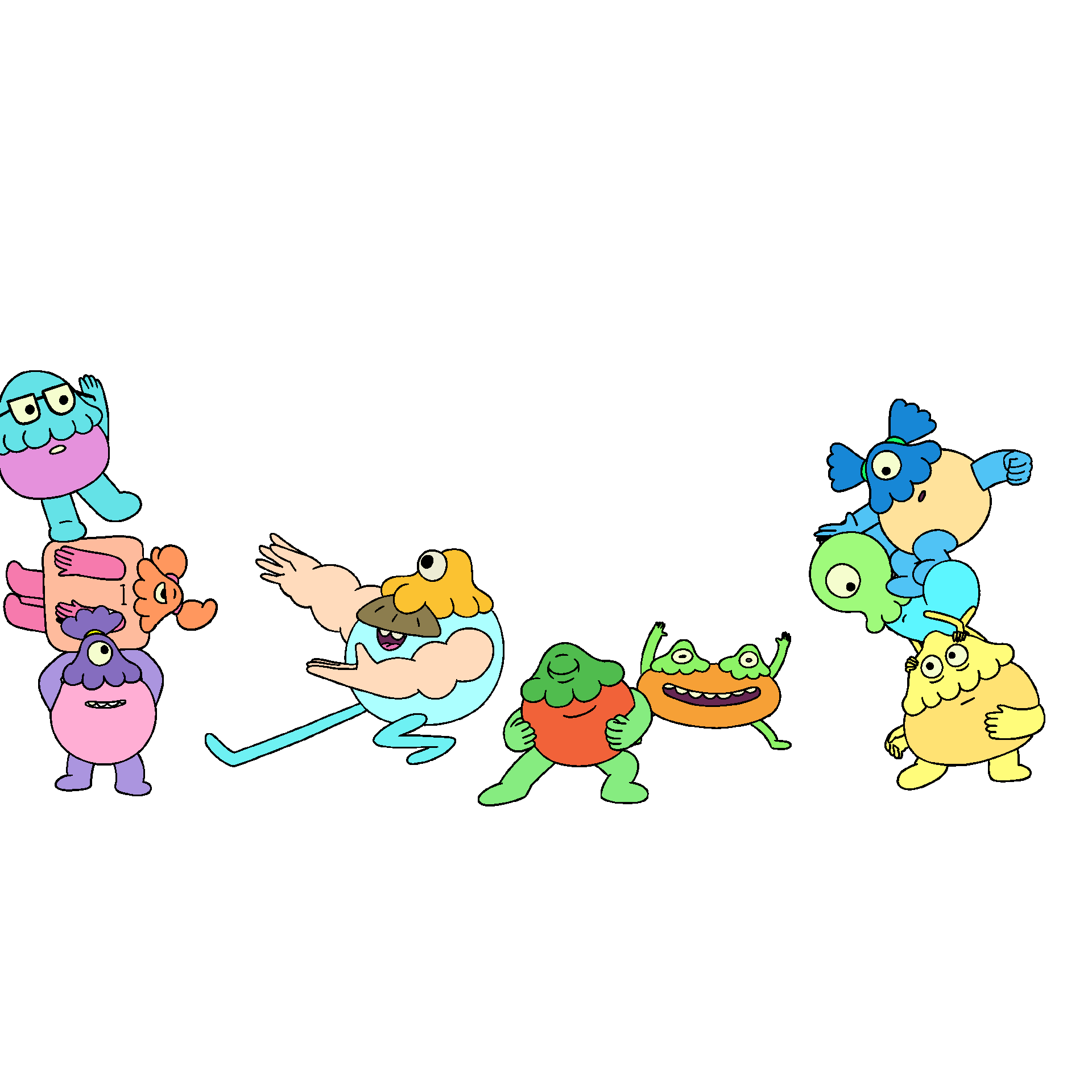} &
  \includegraphics[width=0.2\linewidth]{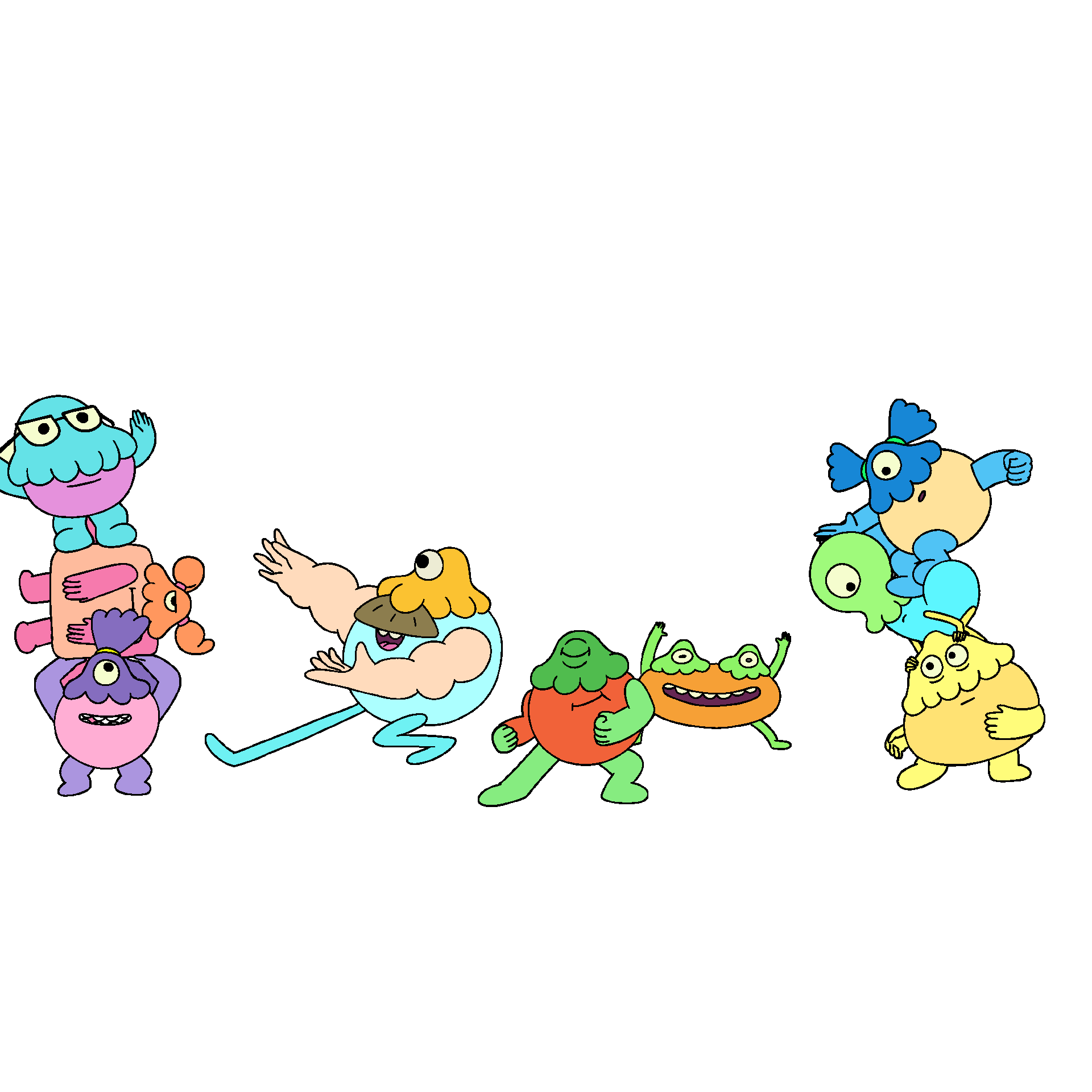} &
  \includegraphics[width=0.2\linewidth]{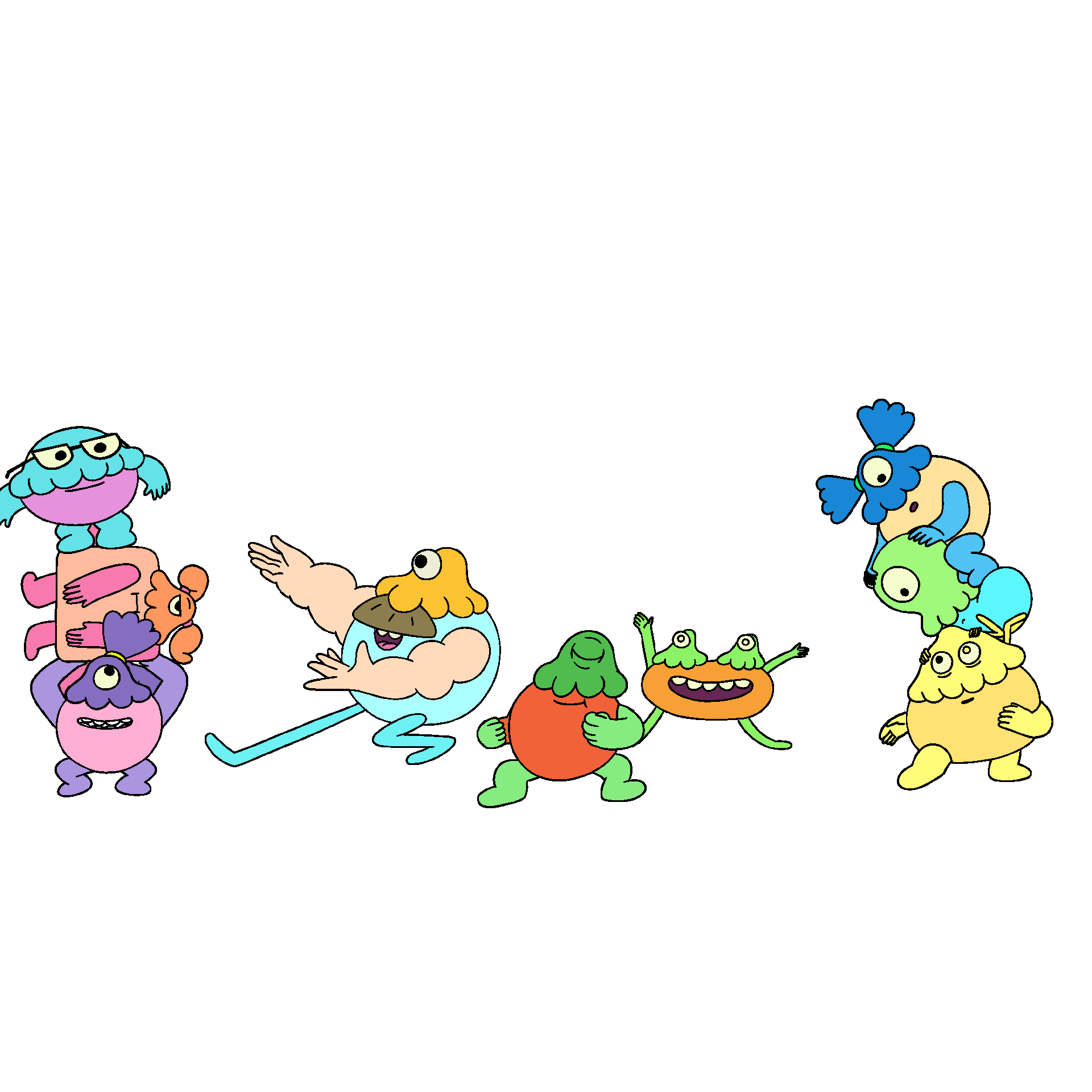} &
  \includegraphics[width=0.2\linewidth]{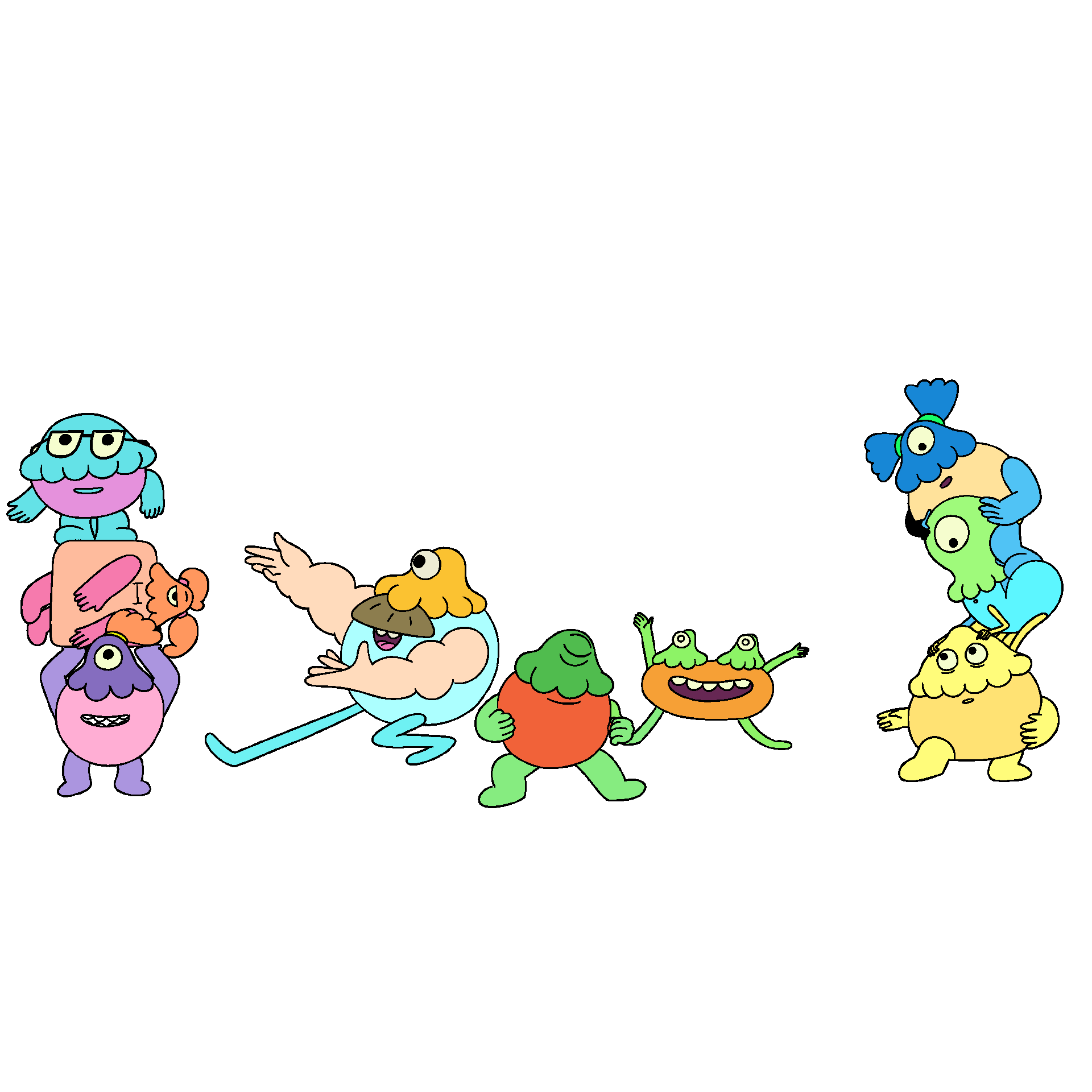} \\
  
  \hline
  
  \raisebox{3\normalbaselineskip}[0pt][0pt]{\rotatebox[origin=c]{90}{DEVC}} &
  \raisebox{0\normalbaselineskip}[0pt][0pt]{\includegraphics[width=0.2\linewidth]{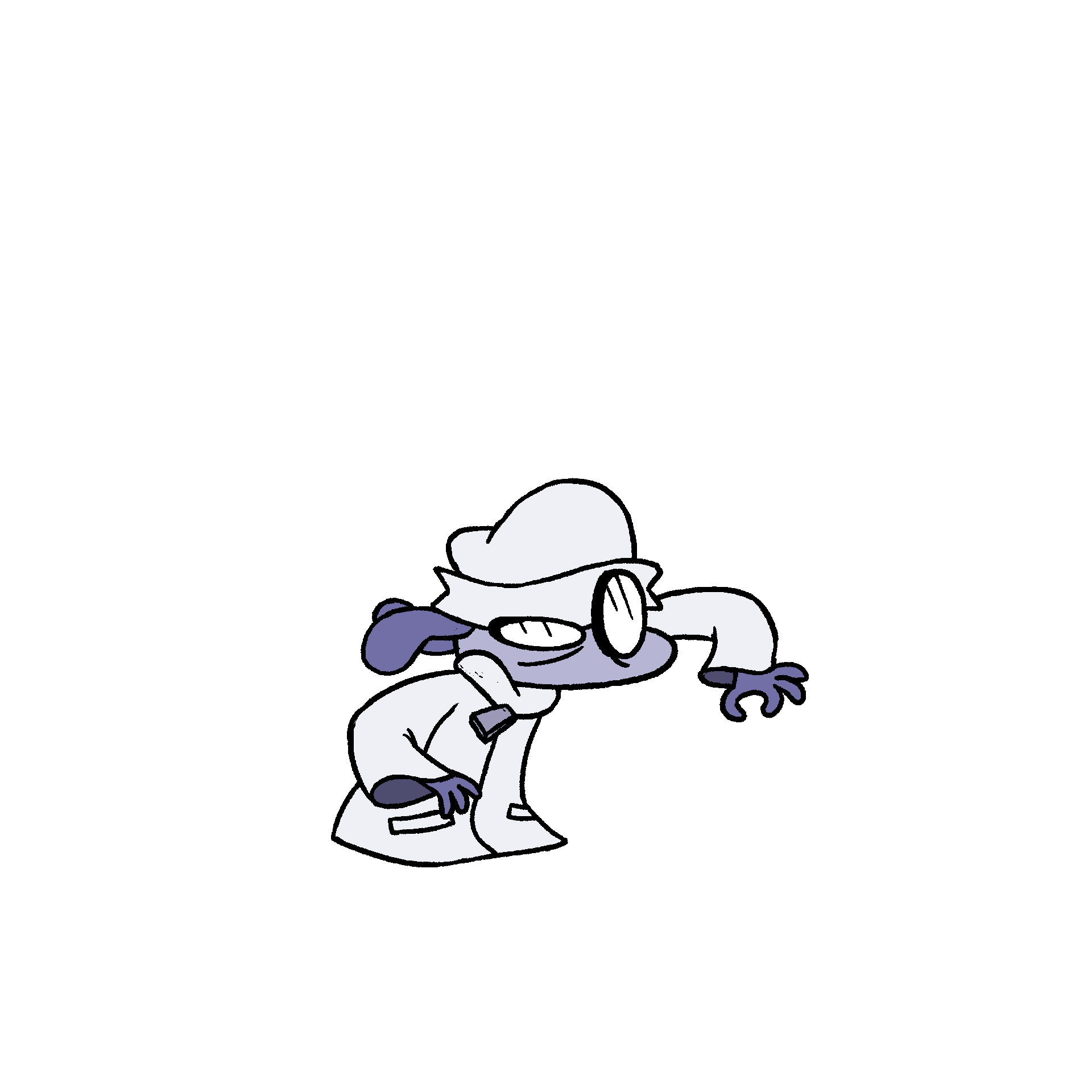}} &
  \includegraphics[width=0.2\linewidth]{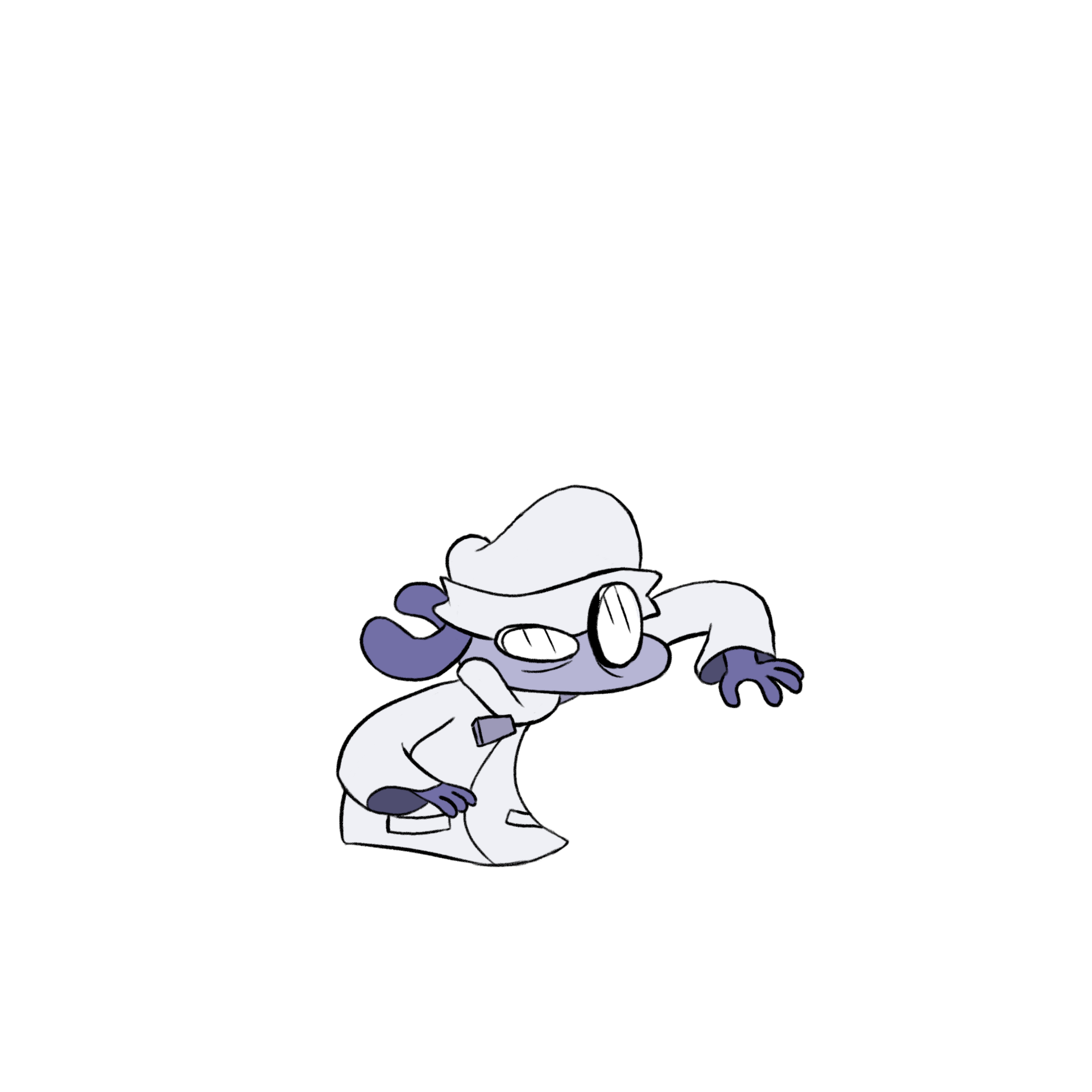} &
  \includegraphics[width=0.2\linewidth]{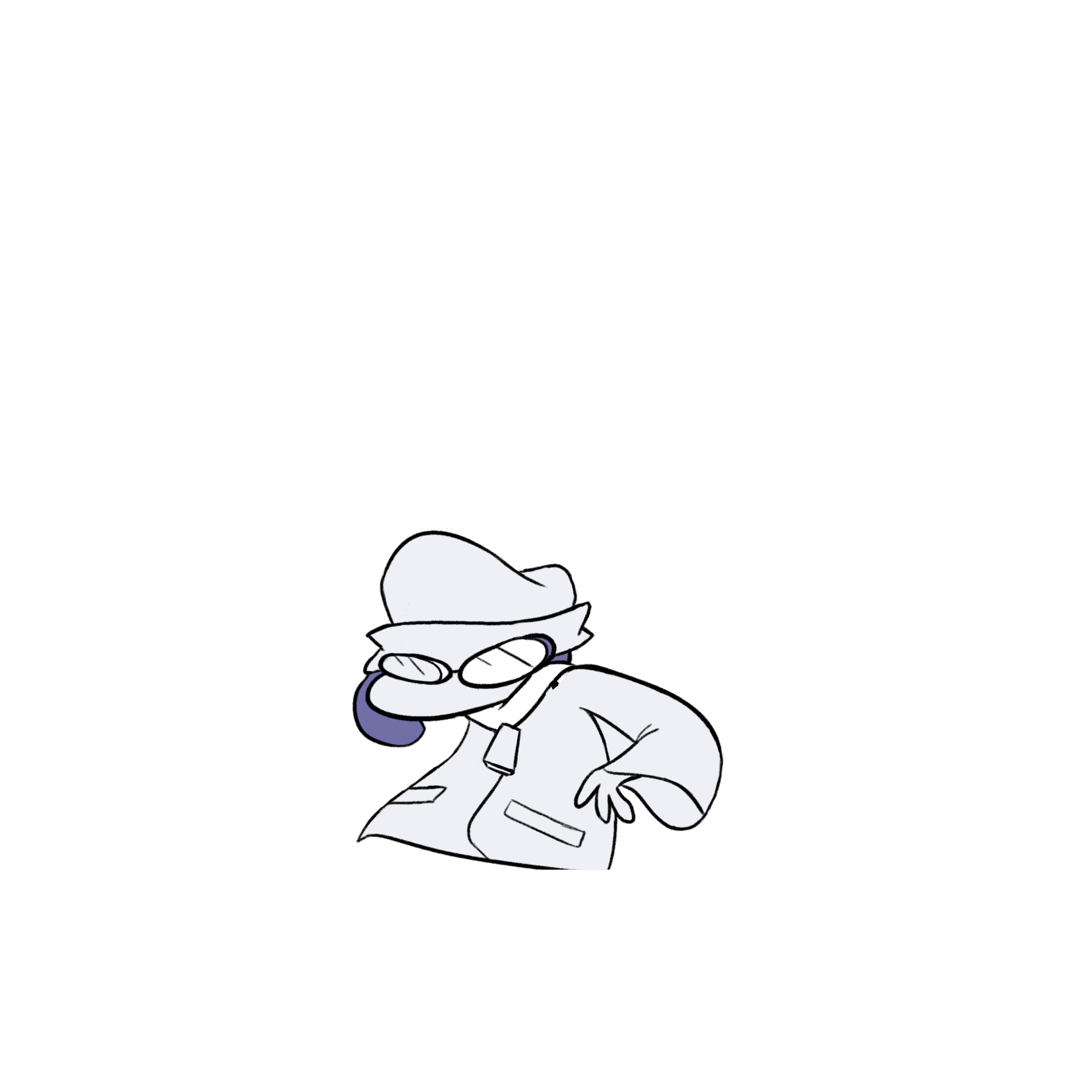} &
  \includegraphics[width=0.2\linewidth]{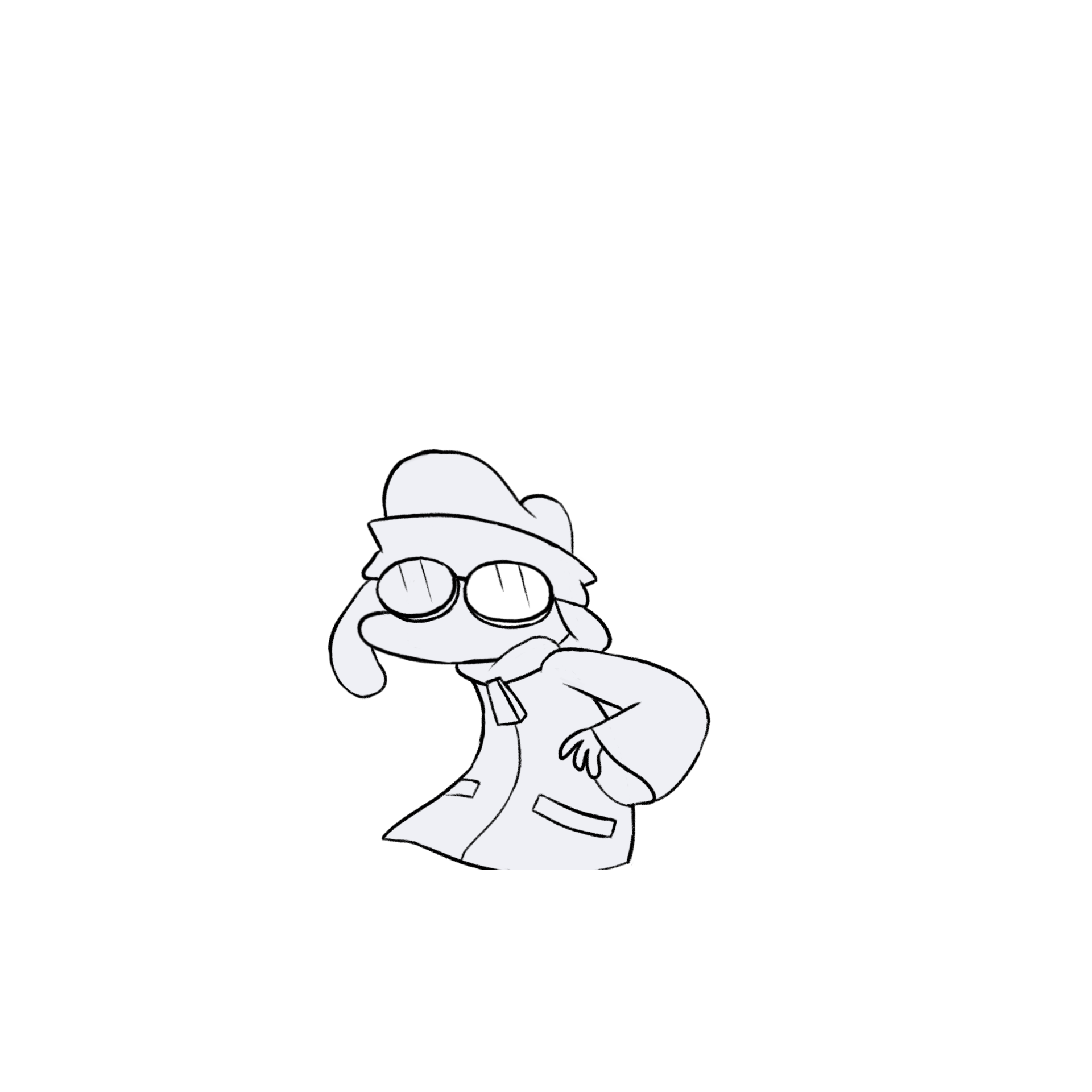} &
  \includegraphics[width=0.2\linewidth]{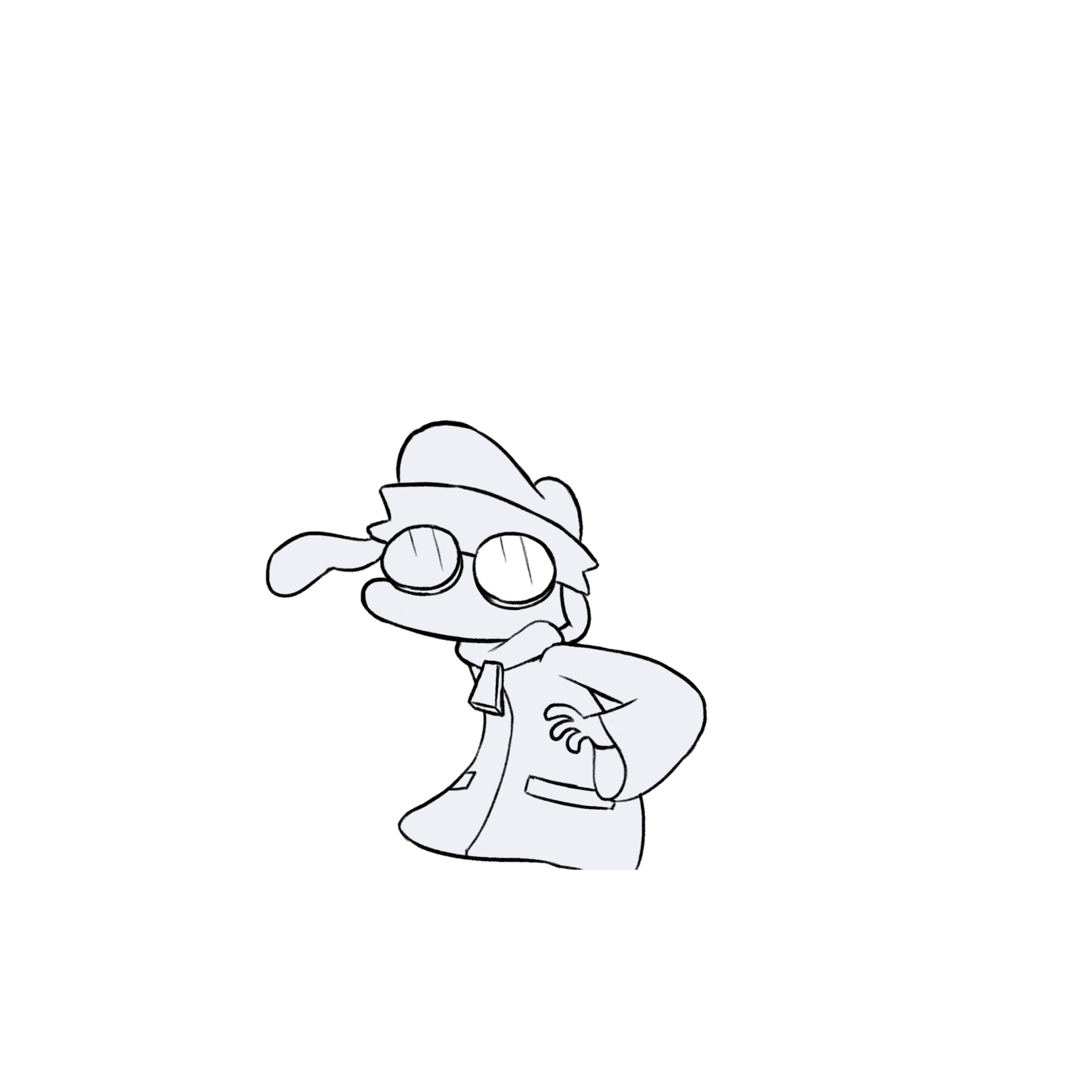} \\
  
  \raisebox{3\normalbaselineskip}[0pt][0pt]{\rotatebox[origin=c]{90}{AnT}} &
  \raisebox{0\normalbaselineskip}[0pt][0pt]{\includegraphics[width=0.2\linewidth]{figures/sh052-10/ref}} &
  \includegraphics[width=0.2\linewidth]{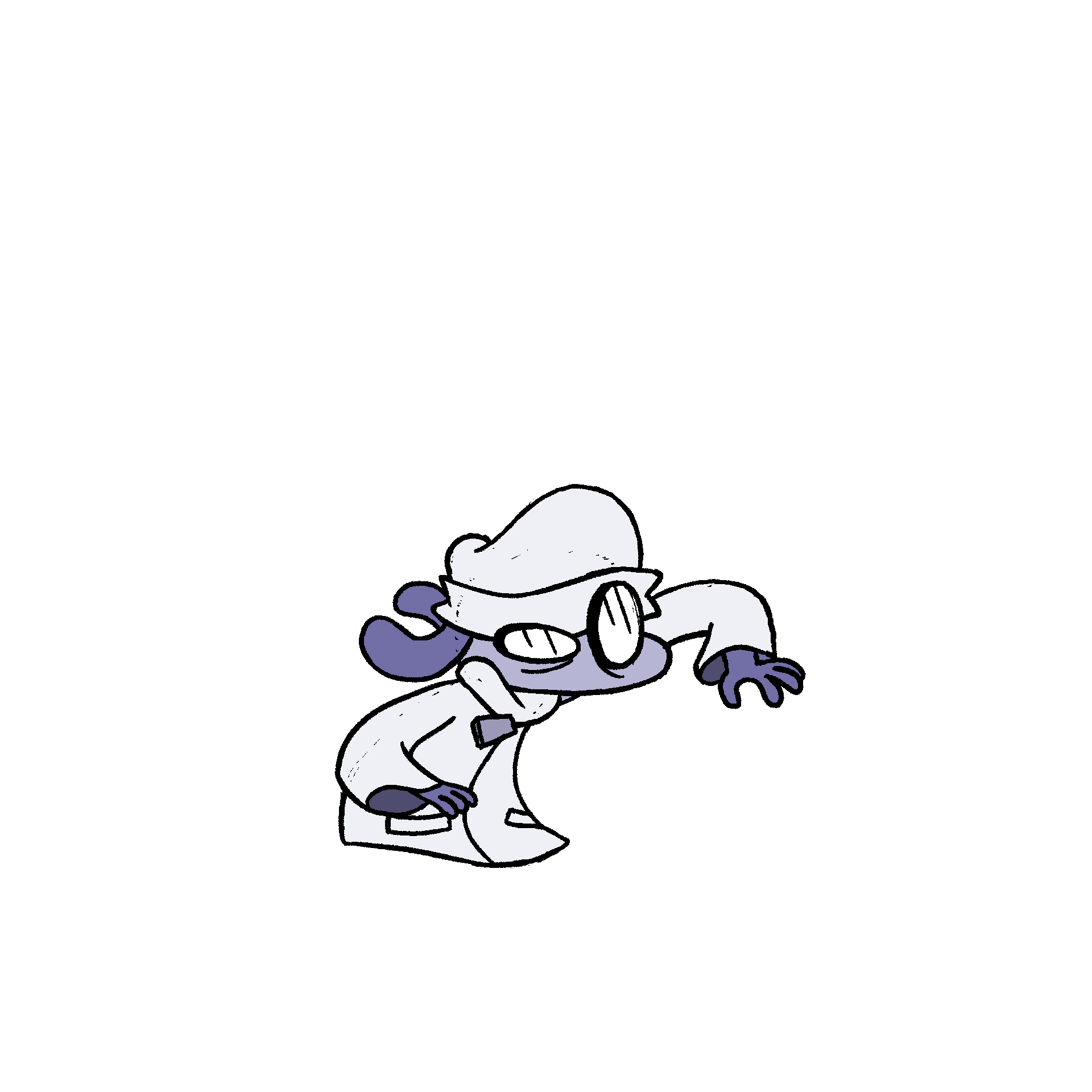} &
  \includegraphics[width=0.2\linewidth]{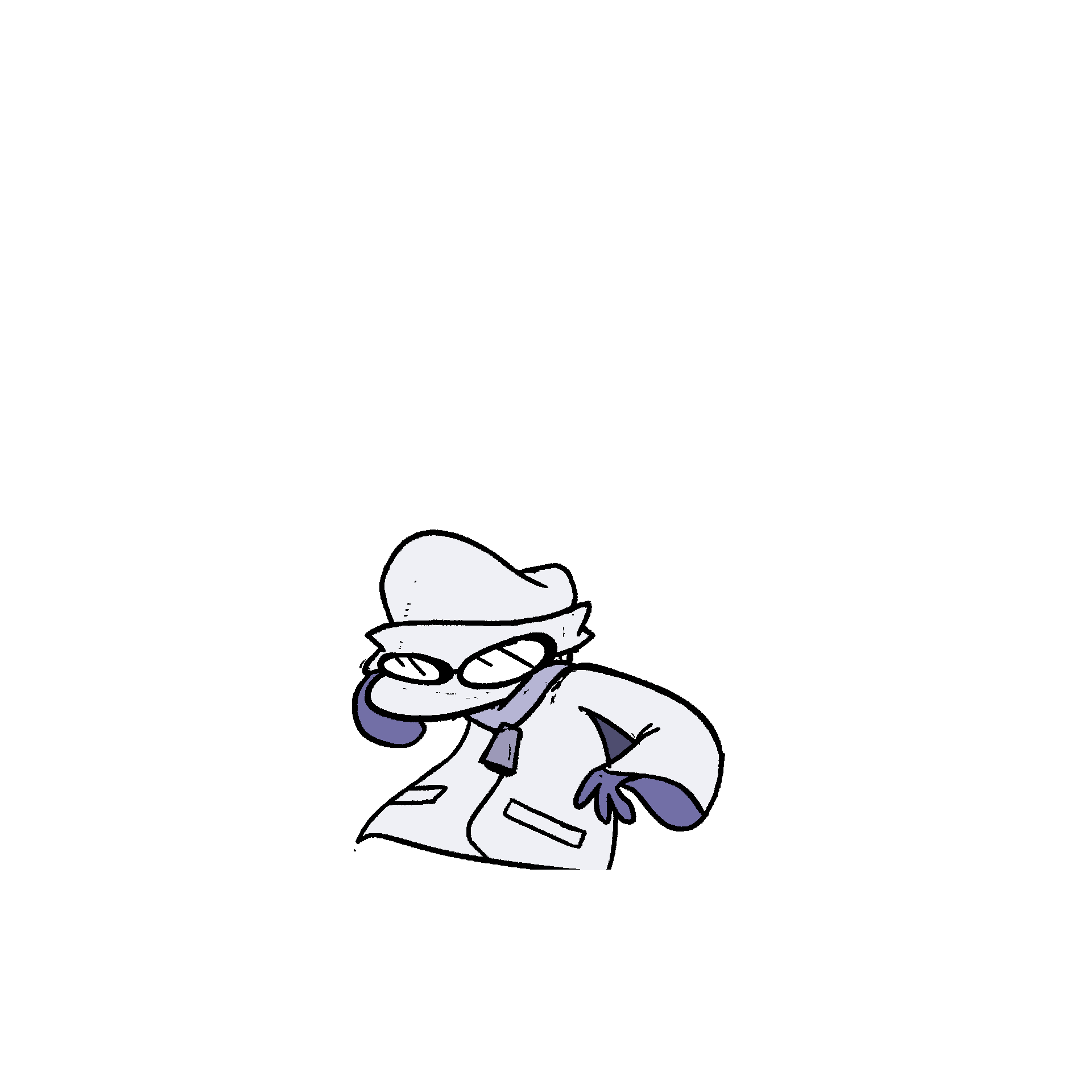} &
  \includegraphics[width=0.2\linewidth]{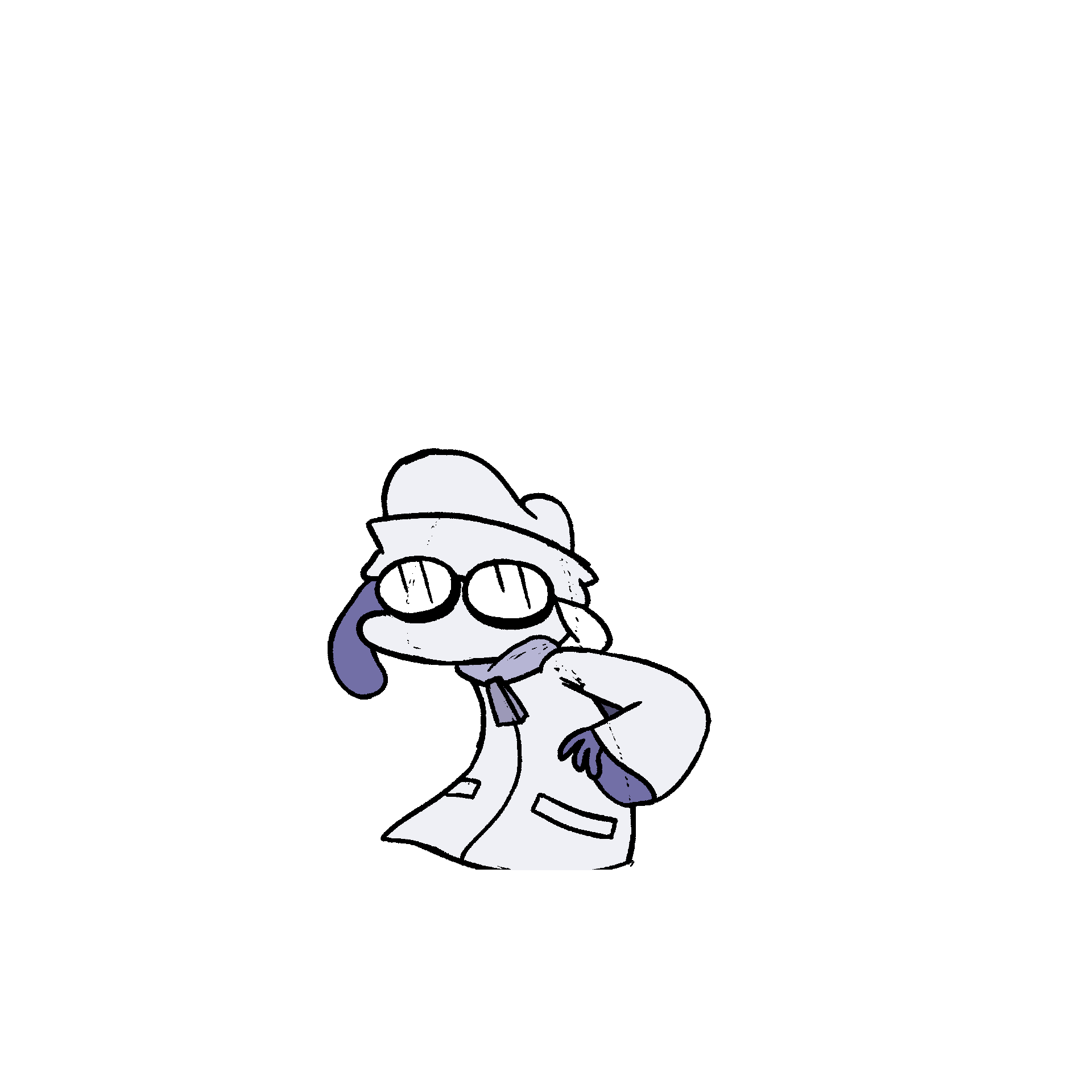} &
  \includegraphics[width=0.2\linewidth]{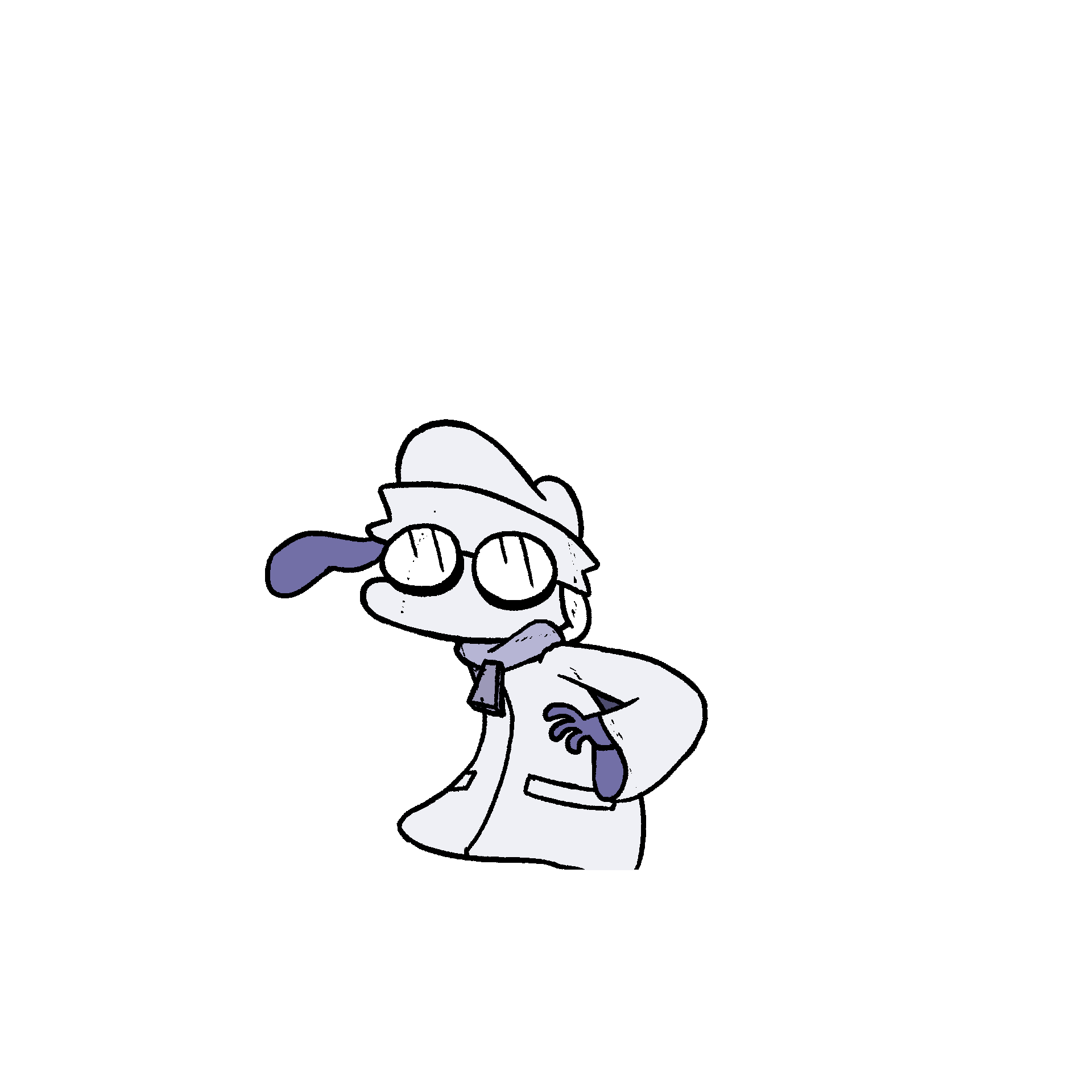} \\
  
  \hline
  
  \raisebox{3\normalbaselineskip}[0pt][0pt]{\rotatebox[origin=c]{90}{DEVC}} &
  \raisebox{0\normalbaselineskip}[0pt][0pt]{\includegraphics[width=0.2\linewidth]{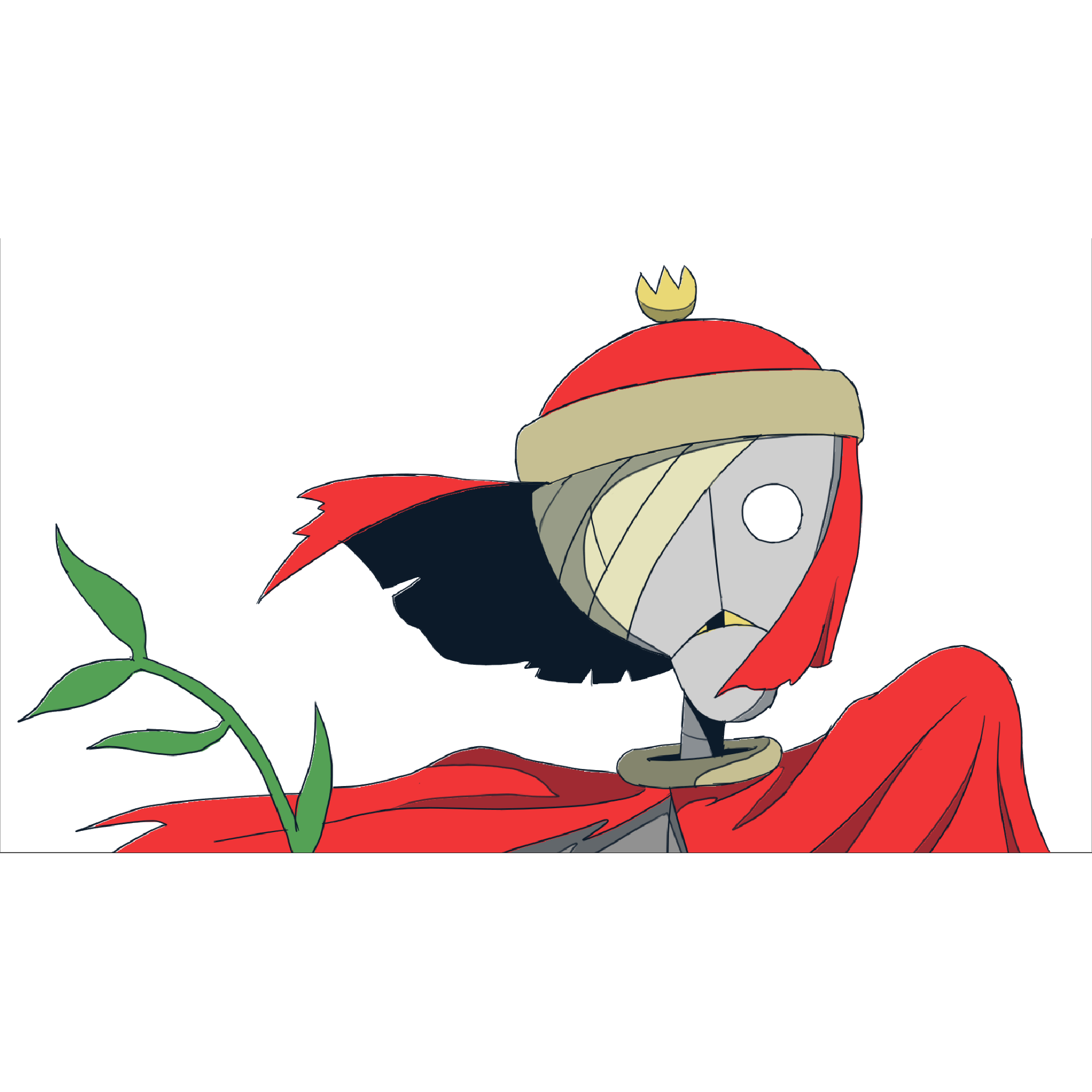}} &
  \includegraphics[width=0.2\linewidth]{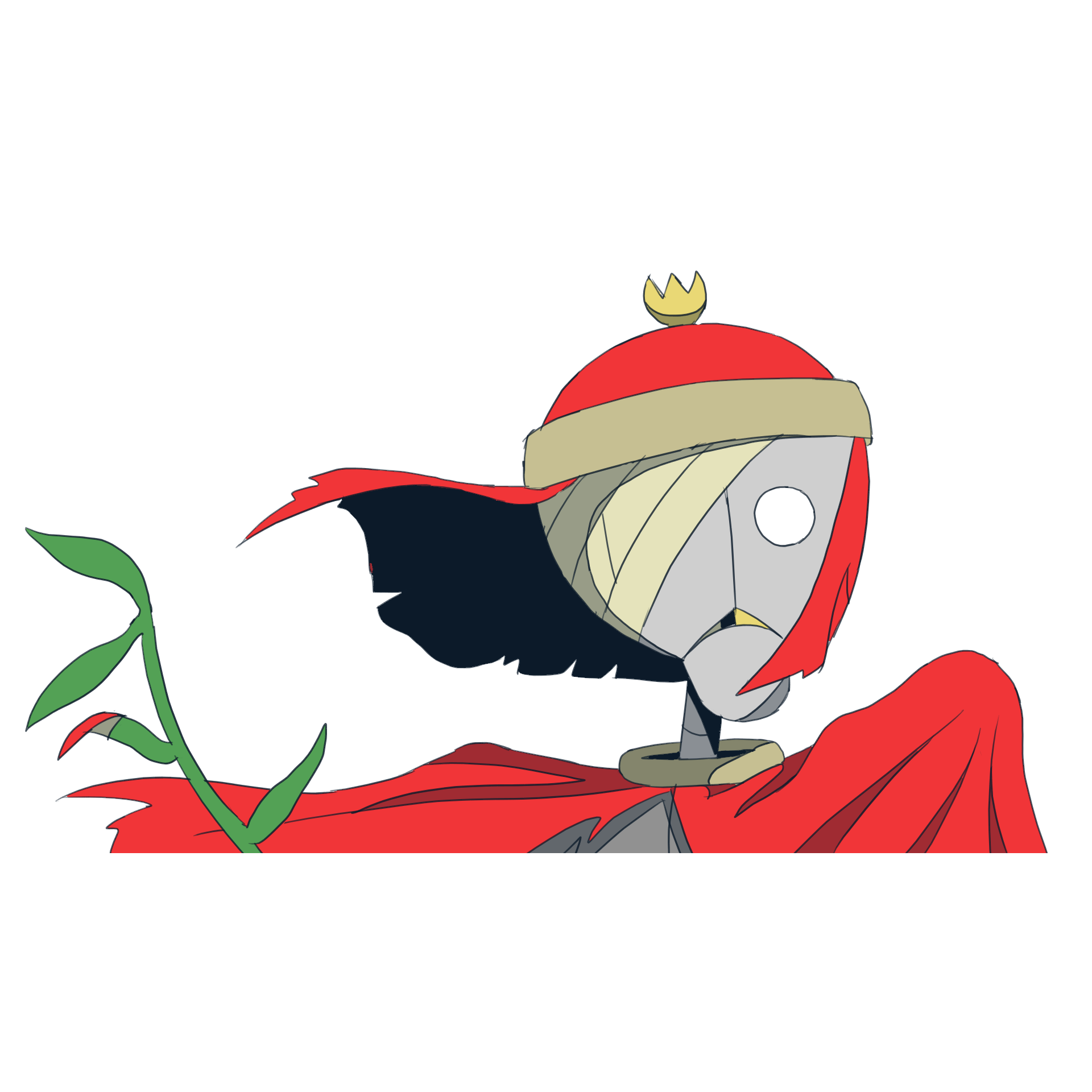} &
  \includegraphics[width=0.2\linewidth]{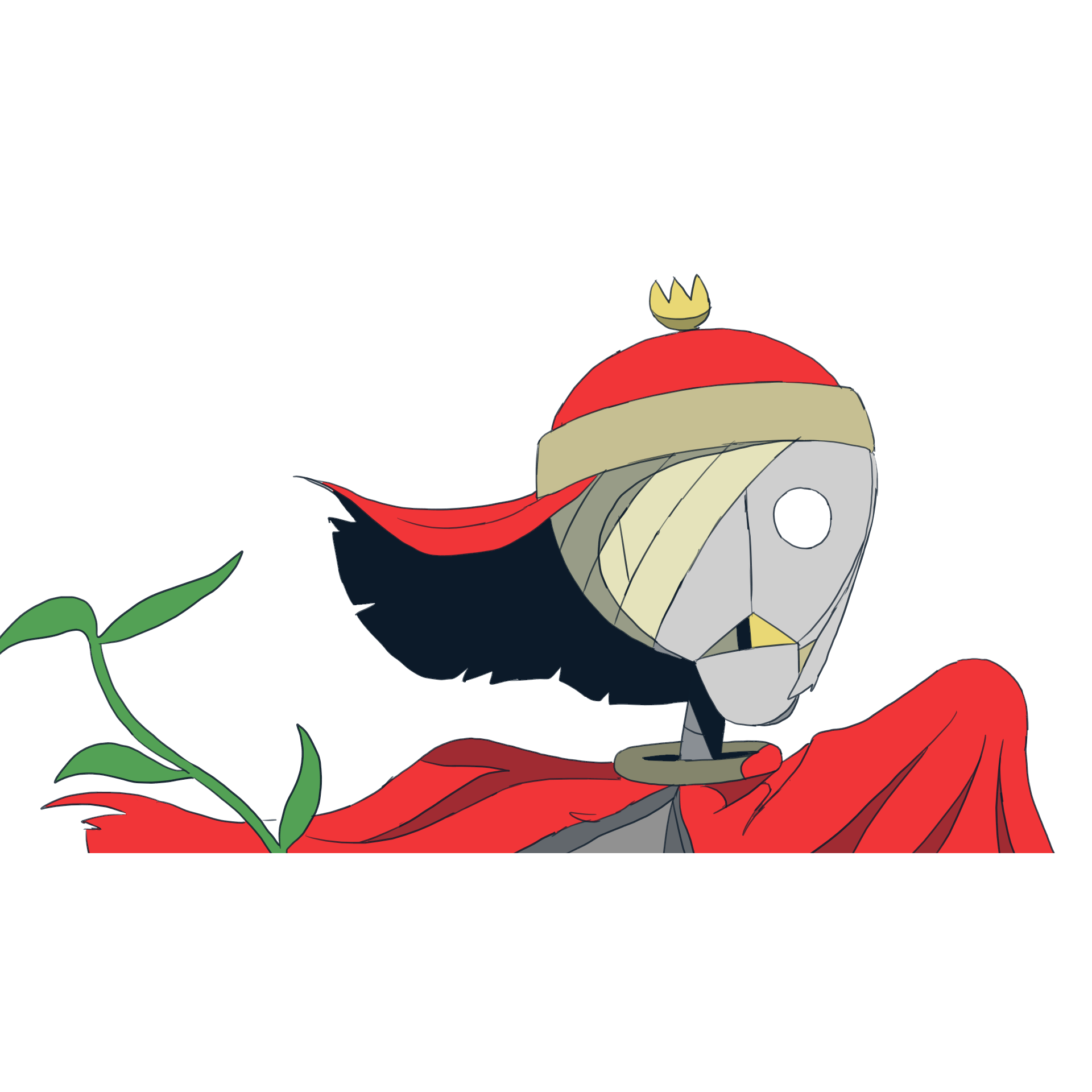} &
  \includegraphics[width=0.2\linewidth]{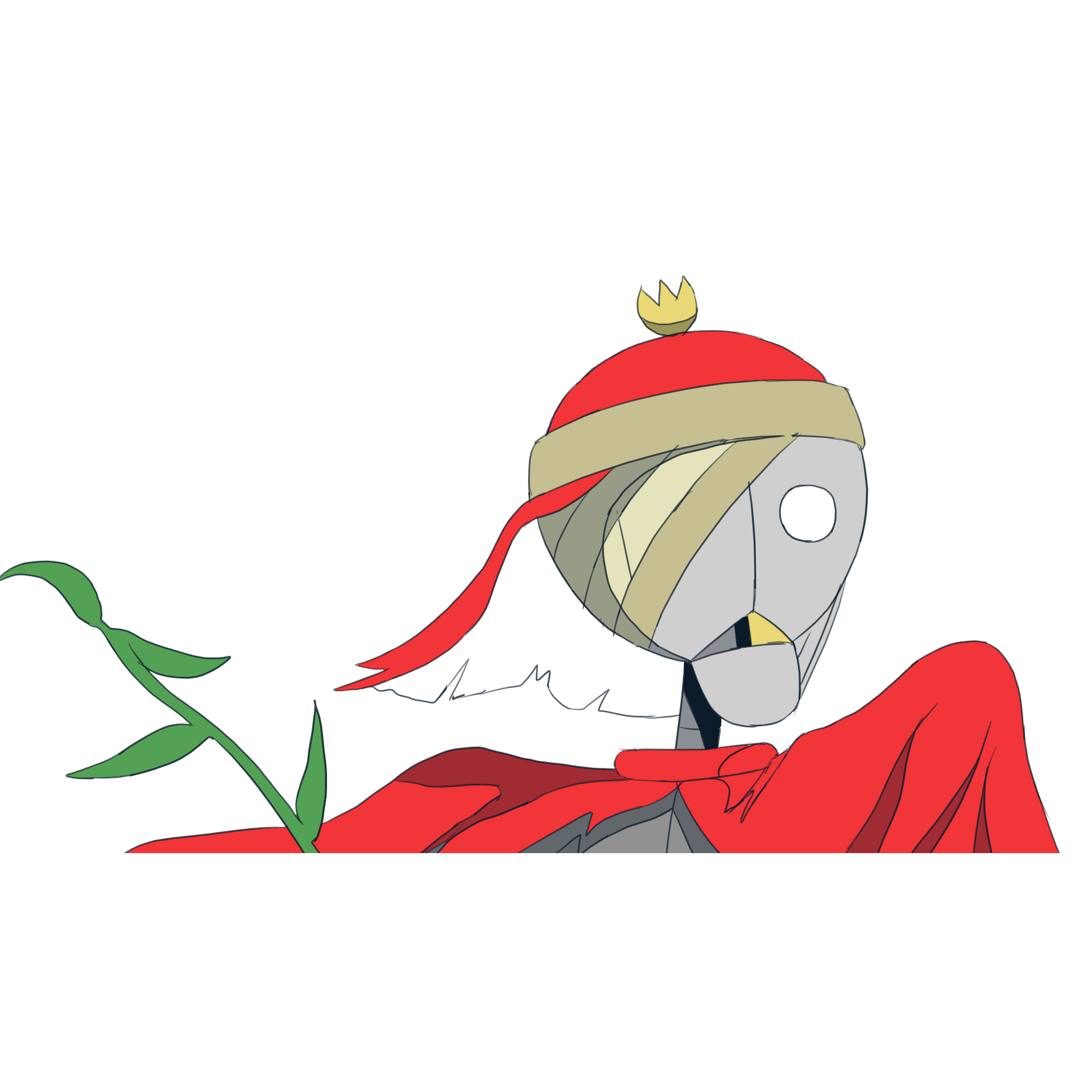} &
  \includegraphics[width=0.2\linewidth]{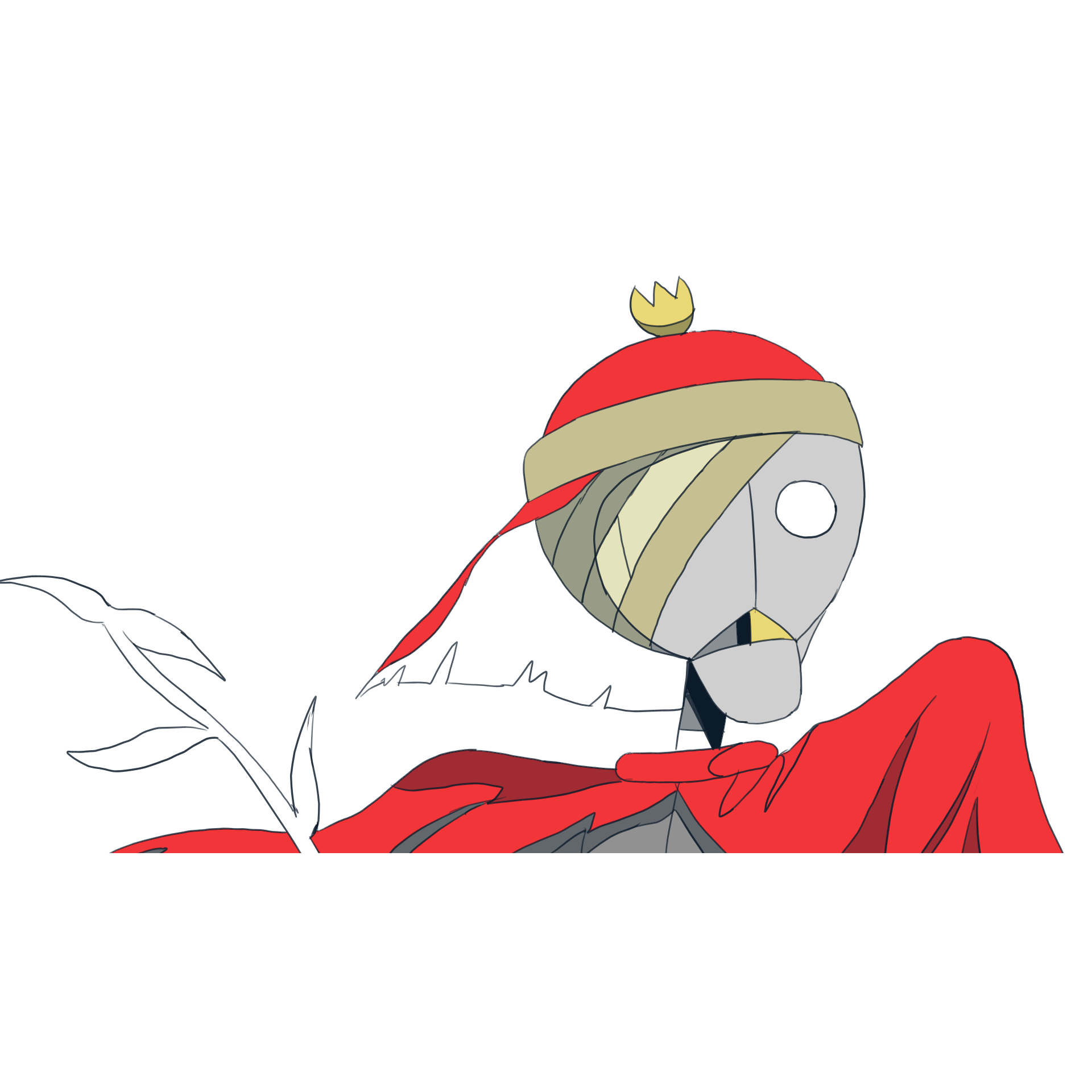} \\
  
  \raisebox{3\normalbaselineskip}[0pt][0pt]{\rotatebox[origin=c]{90}{AnT}} &
  \raisebox{0\normalbaselineskip}[0pt][0pt]{\includegraphics[width=0.2\linewidth]{figures/sh010-20/ref}} &
  \includegraphics[width=0.2\linewidth]{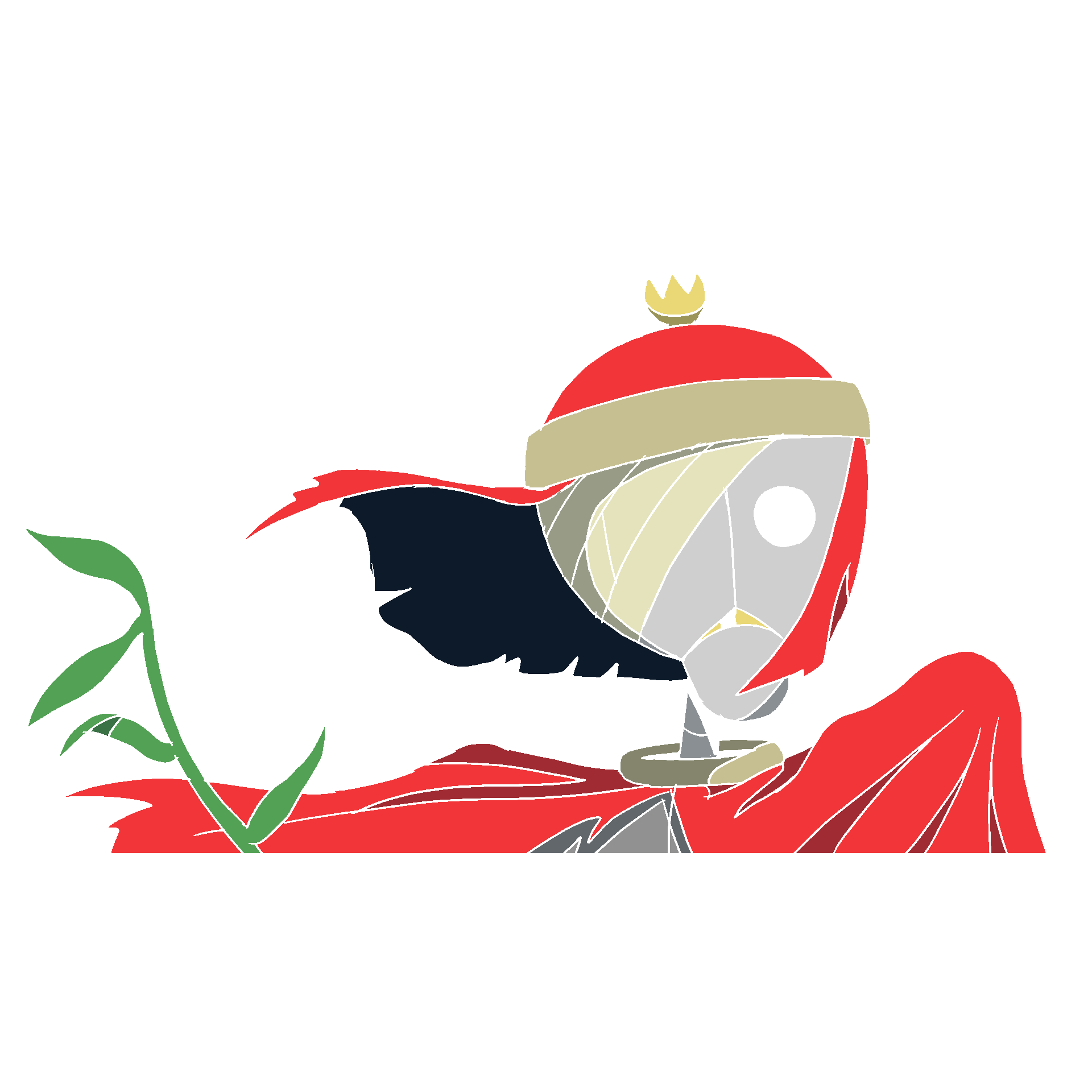} &
  \includegraphics[width=0.2\linewidth]{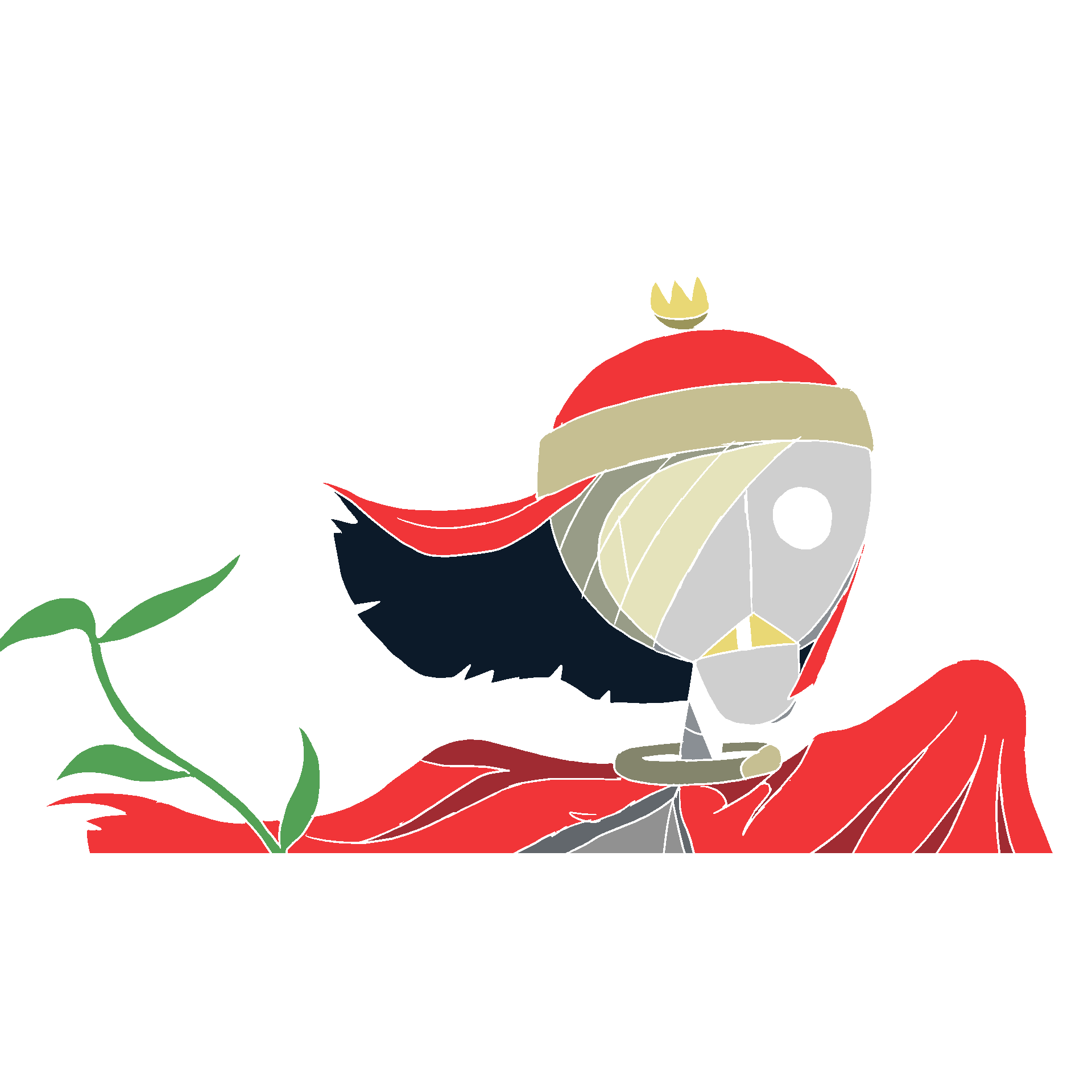} &
  \includegraphics[width=0.2\linewidth]{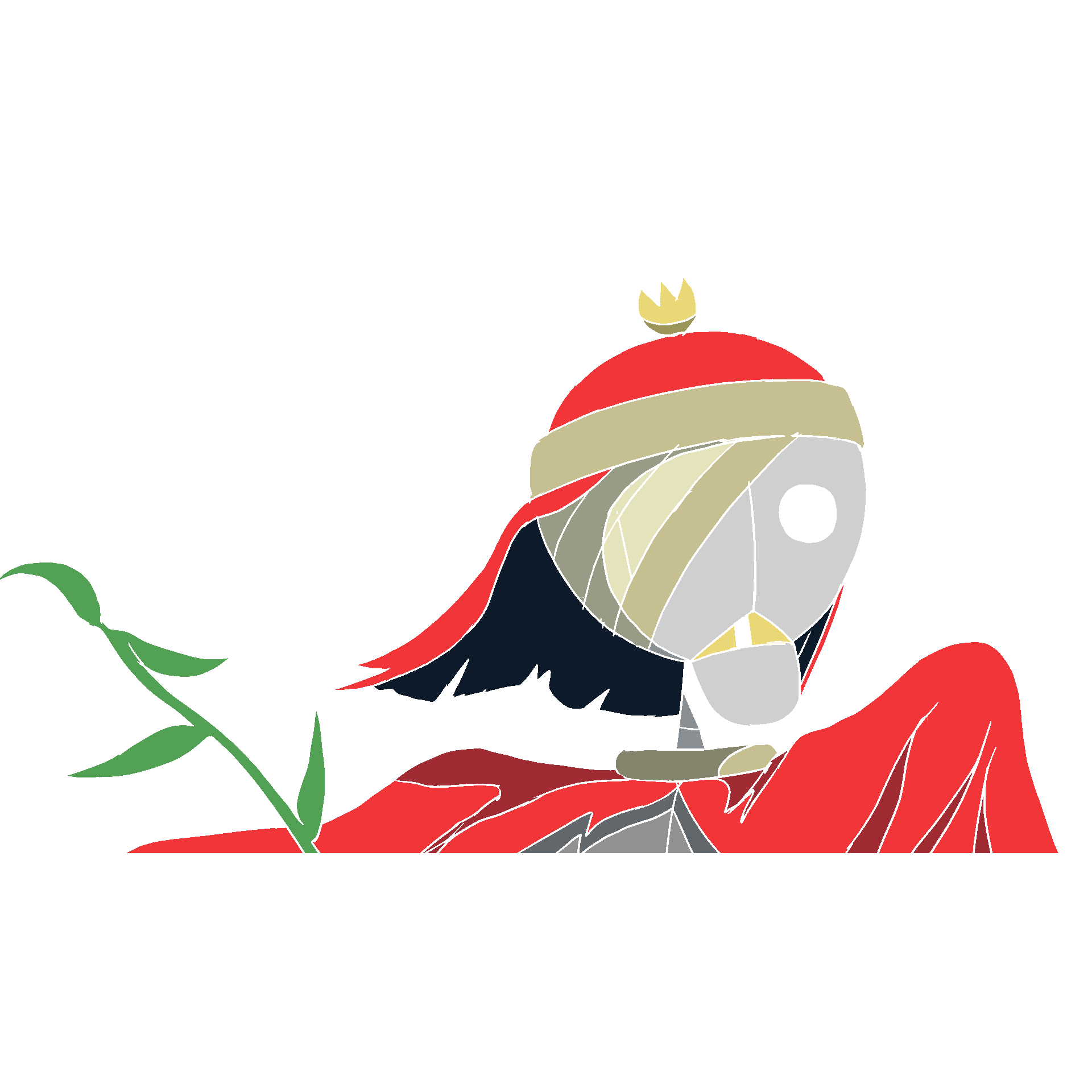} &
  \includegraphics[width=0.2\linewidth]{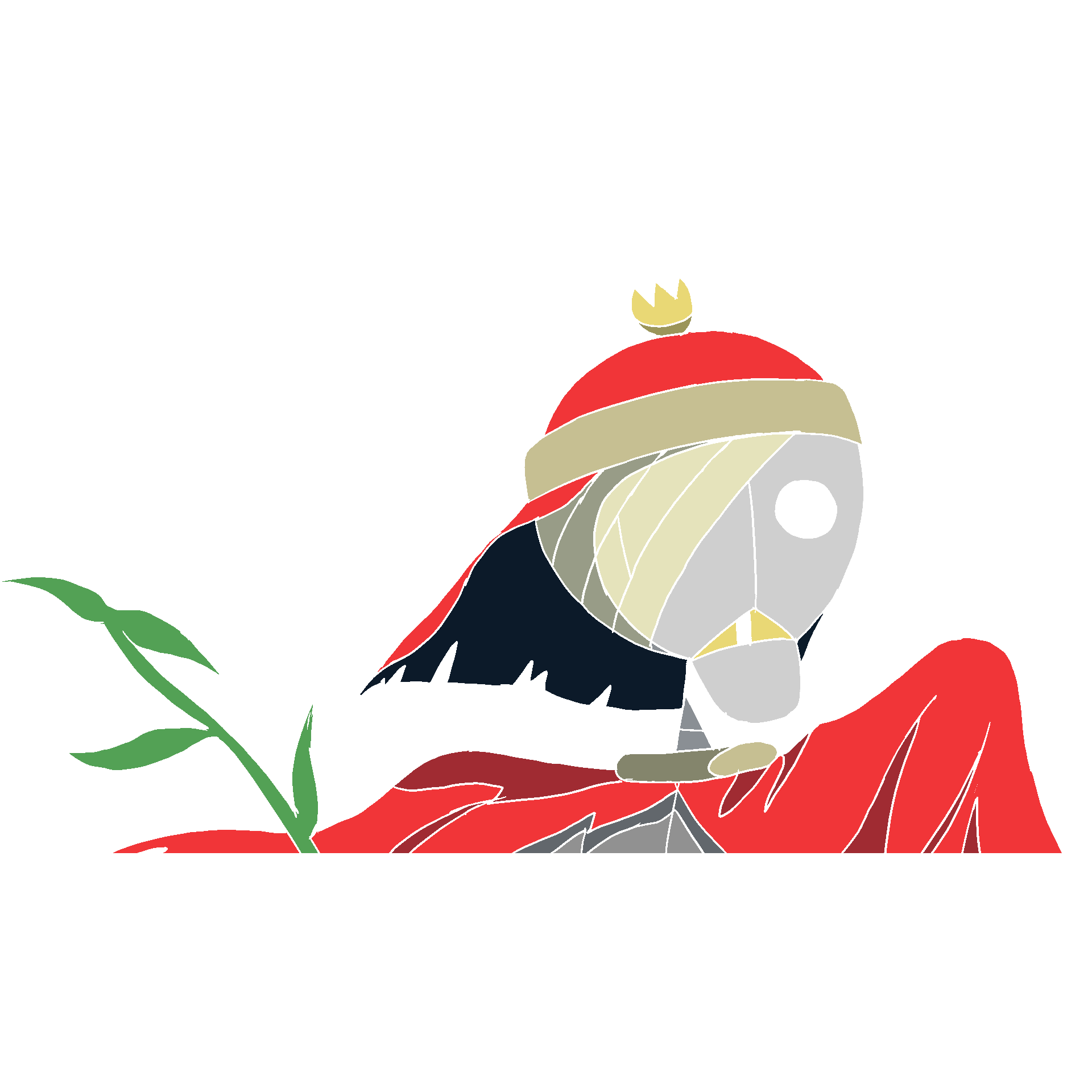} \\
  
\end{tabular}
}
\caption{\textbf{Qualitative results for AnT and DEVC.} Zoom in to view in more detail.}
\label{fig/qualitative}
\end{figure*}

\begin{figure*}[h]
\centering
\includegraphics[width=1\linewidth]{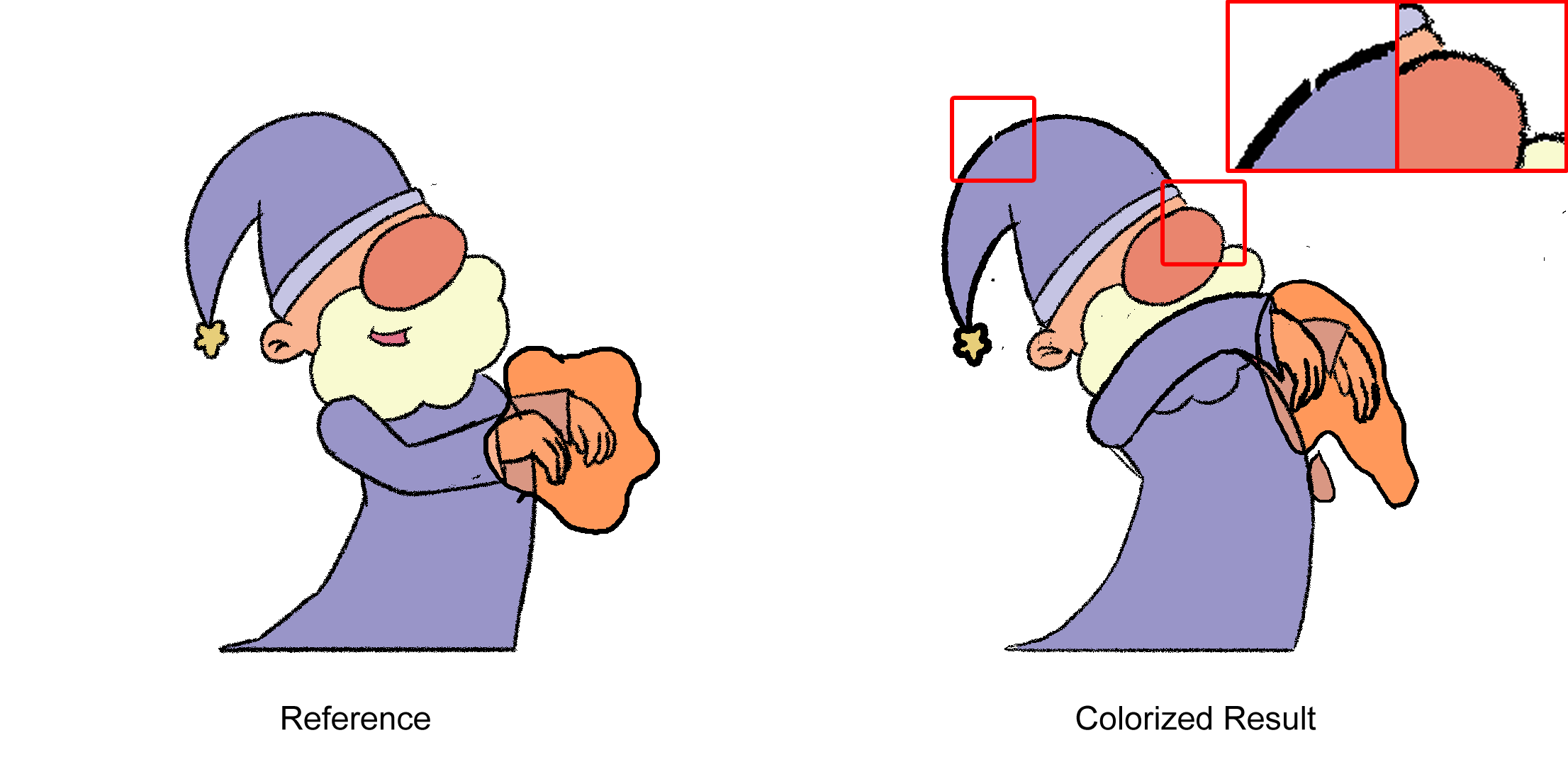}
\caption{ \textbf{Our method can handle line drawings with gaps.} Zoom in to view in more detail.}
\label{fig/gap}
\end{figure*}

\PAR{Additional results:}
In Figure \ref{fig/qualitative} we show qualitative examples of a variate set of sequences colorized with AnT and DEVC. In the same way that previous qualitative examples, these colorization sequences have been created following a recursive propagation of colors, using each colorized image as input for the next generation (as described in figure 8 in main body). AnT presents superior performance especially when dealing with ambiguous segments and occlusions. In Figure \ref{fig/gap} we show results from line drawing with gaps.

\section{Inspecting Attention in AnT}

\begin{figure*}[h]
\centering
\includegraphics[width=1\linewidth]{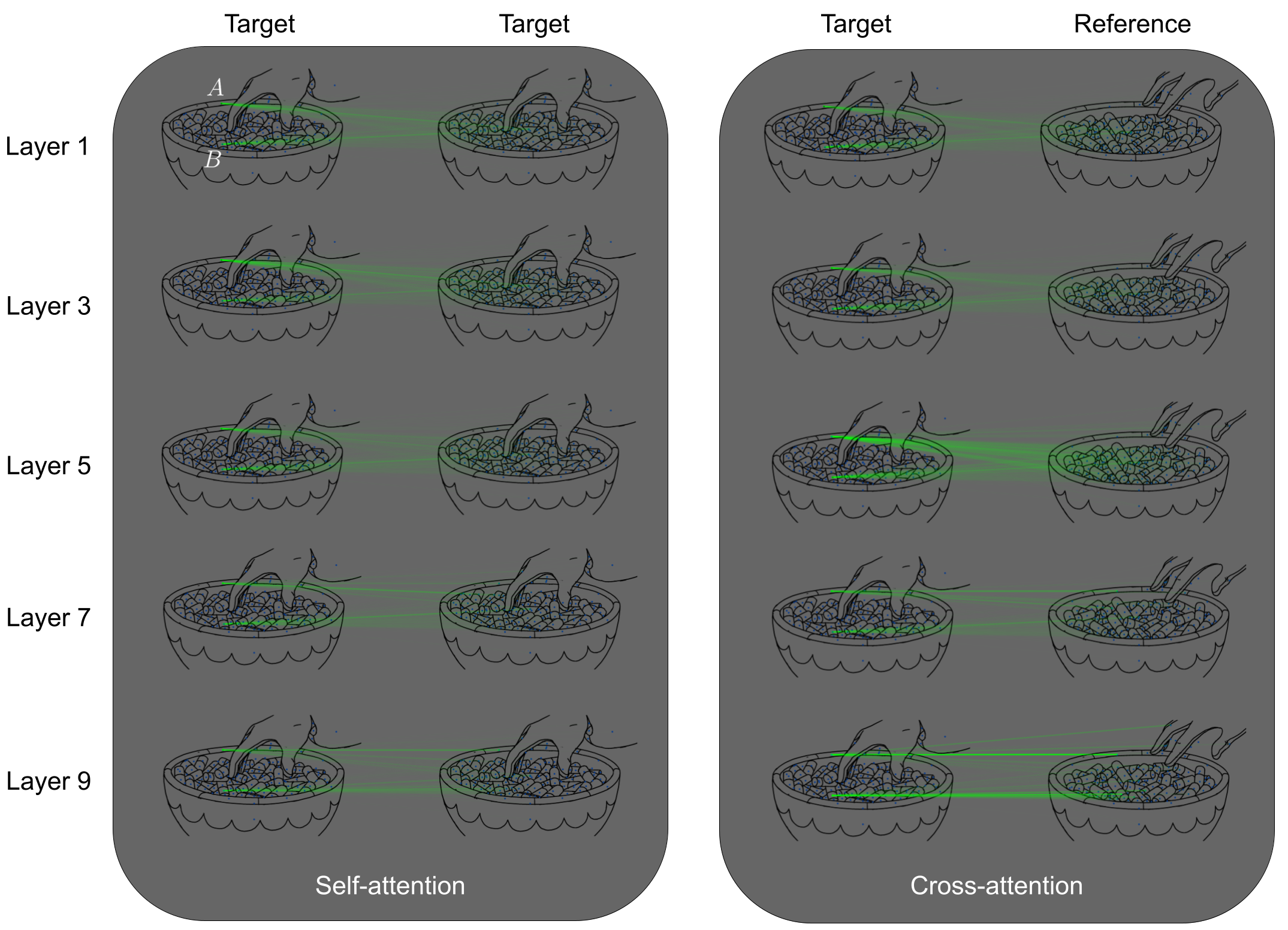}
\caption{ \textbf{Self- and cross- attention layer visualizations for two segments.} The locations of segments $A$ and $B$ are shown in the top left-hand corner.}
\label{fig/attention}
\end{figure*}

In Figure \ref{fig/attention} we present the attention patterns formed in the attention layers of the transformer module at different stages. The visualizations are created for the case where target segment features are updated, i.e. self-attention is computed between segments from the target image and cross-attention aggregates segment information from the reference image to each target segment. The opacity of green lines represents the attention weight between a target segment and each segment from the contrary image. For example, in the first-row of self-attention, the segment $A$ has small attention weights towards multitude of other target segments while in the last row of cross-attention its attention is mainly focused on the correct correspondence from the reference image.

\PAR{ Selected segments:} We have chosen two segments where our model correctly found correspondences in situations where more than just visual information was necessary. In these cases, the spatial and structural information provided by the positional encoder and the transformer was key to disambiguate correct correspondences from wrong matches. We show the robustness of AnT to occlusions with segment $A$ and its ability to find the correct correspondences in ambiguous scenarios with $B$ (which shares visual resemblance with its neighboring segments).

\PAR{Attention patterns:} From our experiments, we can appreciate how attention focuses on gathering information from lots of segments from the contrary images in early layers. We argue that segment representations get benefited from attending a large number of segments all around the image to get a sense of the global structure of the scene and its relative distances with other segments. Towards the later layers, attention gets progressively narrowed towards the most important elements to represent each segment. This is important to disambiguate between similar segments. For example, in the latest row of the cross-attention layer, both segments still gather information from enclosures close to them such as the hand for $A$ or other pills for $B$.

\end{document}